\documentclass[letterpaper]{article} % DO NOT CHANGE THIS
\usepackage[]{aaai2026}  % DO NOT CHANGE THIS
\usepackage{times}  % DO NOT CHANGE THIS
\usepackage{helvet}  % DO NOT CHANGE THIS
\usepackage{courier}  % DO NOT CHANGE THIS
\usepackage[hyphens]{url}  % DO NOT CHANGE THIS
\usepackage{graphicx} % DO NOT CHANGE THIS
\usepackage{natbib}  % DO NOT CHANGE THIS AND DO NOT ADD ANY OPTIONS TO IT
\usepackage{caption} % DO NOT CHANGE THIS AND DO NOT ADD ANY OPTIONS TO IT
\usepackage{algorithm}
\usepackage{algorithmic}

\usepackage{newfloat}
\usepackage{listings}
\DeclareCaptionStyle{ruled}{labelfont=normalfont,labelsep=colon,strut=off} % DO NOT CHANGE THIS
\floatstyle{ruled}
\newfloat{listing}{tb}{lst}{}
\floatname{listing}{Listing}
\usepackage[utf8]{inputenc} % allow utf-8 input
\usepackage[T1]{fontenc}    % use 8-bit T1 fonts
\usepackage{url}            % simple URL typesetting
\usepackage{booktabs}       % professional-quality tables
\usepackage{amsfonts}       % blackboard math symbols
\usepackage{nicefrac}       % compact symbols for 1/2, etc.
\usepackage{microtype}      % microtypography
\usepackage{xcolor}         % colors

\usepackage{amsmath}
\usepackage{amssymb}
\usepackage{mathtools}
\usepackage{amsthm}

\usepackage[capitalize,nameinlink]{cleveref}
\theoremstyle{plain}
\newtheorem{theorem}{Theorem}[section]

\newtheorem*{customthm}{Theorem 3.1}
\newtheorem*{thmreduction}{Theorem 4.1}
\newtheorem*{thmsignal}{Theorem 4.2}

\newtheorem{lemma}[theorem]{Lemma}
\newtheorem{corollary}[theorem]{Corollary}
\theoremstyle{definition}

\theoremstyle{remark}

\usepackage{algorithm}
\usepackage{algorithmic}
\usepackage[algo2e]{algorithm2e}
\usepackage[utf8]{inputenc}
\usepackage{subfigure}
\usepackage{multirow}
\usepackage{pbox}
\usepackage{graphicx}
\usepackage{bm}
\usepackage{siunitx}
\usepackage{booktabs}
\usepackage{lipsum}
\usepackage{caption}
\usepackage{fixltx2e}
\usepackage{cancel}
\usepackage{colortbl}
\usepackage{enumitem}
\usepackage{nicefrac}
\usepackage{microtype}
\usepackage[toc,page,header]{appendix}
\usepackage{minitoc}

\usepackage[textsize=tiny]{todonotes}
\usepackage{marginnote}

\definecolor{bluegray}{rgb}{0.4, 0.6, 0.8}
\definecolor{electriclime}{rgb}{0.8, 1.0, 0.0}
\definecolor{malachite}{rgb}{0.04, 0.85, 0.32}
\definecolor{darkred}{rgb}{0.55, 0.0, 0.0}
\definecolor{darkblue}{rgb}{0.0, 0.0, 0.55}
\definecolor{darkgreen}{rgb}{0.0, 0.2, 0.13}
\definecolor{darkorchid}{rgb}{0.6, 0.2, 0.8}
\Crefname{figure}{Fig.}{Figs.}% {<type>}{<singular>}{<plural>}
\Crefname{Equation}{Eq.}{Eqs.}
\newcommand{\std}[1]{\tiny{\ensuremath{\pm}}\text{\tiny #1}}

\newcommand{\ORG}[1]{\textcolor{orange}{#1}}
\newcommand{\BEST}[1]{\textbf{\textcolor{red}{#1}}}
\newcommand{\SECOND}[1]{\textbf{\textcolor{blue}{#1}}}
\newcommand{\THIRD}[1]{\textbf{\color{malachite}{#1}}}

\usepackage{pifont}% http://ctan.org/pkg/pifont
\newcommand{\cmark}{\ding{51}}%
\newcommand{\xmark}{\ding{55}}%
\definecolor{lightRed}{HTML}{F8CECC}
\definecolor{redBorder}{HTML}{CC0000}
\definecolor{greenBorder}{HTML}{82B366}
\definecolor{lightGreen}{HTML}{D5E8D4}
\definecolor{blueBorder}{HTML}{6C8EBF}
\definecolor{lightBlue}{HTML}{DAE8FC}
\definecolor{blackBorder}{HTML}{000000}

\DeclareRobustCommand\myrednode{\raisebox{-2pt} {\tikz[]{\node[shape=circle,draw=redBorder,fill=lightRed,text=lightRed, inner sep=1pt]{\scriptsize{A}};}}}
\DeclareRobustCommand\mybluenode{\raisebox{-2pt} {\tikz[]{\node[shape=circle,draw=blueBorder,fill=lightBlue,text=lightBlue,inner sep=1pt]{\scriptsize{A}};}}}
\DeclareRobustCommand\mygreennode{\raisebox{-2pt} {\tikz[]{\node[shape=circle,draw=greenBorder,fill=lightGreen,text=lightGreen,inner sep=1pt, ]{\scriptsize{A}};}}}

\usepackage{listings}
\definecolor{codegreen}{rgb}{0,0.6,0}
\definecolor{codegray}{rgb}{0.5,0.5,0.5}
\definecolor{codepurple}{rgb}{0.58,0,0.82}
\lstdefinestyle{mystyle}{
    commentstyle=\color{codegreen},
    keywordstyle=\color{magenta},
    numberstyle=\tiny\color{codegray},
    stringstyle=\color{codepurple},
    basicstyle=\ttfamily\footnotesize,
    breakatwhitespace=false,         
    breaklines=true,                 
    captionpos=b,                    
    keepspaces=true,                 
    numbers=left,                    
    numbersep=5pt,
    showspaces=false,                
    showstringspaces=false,
    showtabs=false,                  
    tabsize=2
}
\definecolor{diffstart}{named}{gray}
\definecolor{diffincl}{named}{green}
\definecolor{diffrem}{named}{orange}
\lstdefinelanguage{diff}{
basicstyle=\ttfamily\small,
morecomment=[f][\color{diffstart}]{@@},
morecomment=[f][\color{diffincl}]{+\ },
morecomment=[f][\color{diffrem}]{-\ },
}
\definecolor{commentcolor}{RGB}{110,154,155}   % define comment color
\newcounter{checksubsection}
\newcounter{checkitem}[checksubsection]

\title{Are Graph Transformers Necessary? Efficient Long-Range Message Passing with Fractal Nodes in MPNNs}
\author{
    Jeongwhan Choi\textsuperscript{\rm 1},
    Seungjun Park\textsuperscript{\rm 1},
    Sumin Park\textsuperscript{\rm 1},
    Sung-Bae Cho\textsuperscript{\rm 2},
    Noseong Park\textsuperscript{\rm 1}
}
\affiliations{
    \textsuperscript{\rm 1}KAIST\\
    \textsuperscript{\rm 2}Yonsei University\\
    jeongwhan.choi@kaist.ac.kr
}

\begin{document}

\maketitle

\begin{abstract}
Graph Neural Networks (GNNs) have emerged as powerful tools for learning on graph-structured data, but often struggle to balance local and global information. While graph Transformers aim to address this by enabling long-range interactions, they often overlook the inherent locality and efficiency of Message Passing Neural Networks (MPNNs). We propose a new concept called `\emph{fractal nodes}', inspired by the fractal structure observed in real-world networks. Our approach is based on the intuition that graph partitioning naturally induces fractal structure, where subgraphs often reflect the connectivity patterns of the full graph. Fractal nodes are designed to coexist with the original nodes and adaptively aggregate subgraph-level feature representations, thereby enforcing feature similarity within each subgraph. We show that fractal nodes alleviate the over-squashing problem by providing direct shortcut connections that enable long-range propagation of subgraph-level representations. Experiment results show that our method improves the expressive power of MPNNs and achieves comparable or better performance to graph Transformers while maintaining the computational efficiency of MPNN by improving the long-range dependencies of MPNN.
\end{abstract}

\begin{links}
    \link{Code}{https://github.com/jeongwhanchoi/MPNN-FN}
    % \link{Datasets}{https://aaai.org/example/datasets}
    % \link{Extended version}{https://aaai.org/example/extended-version}
\end{links}

\section{Introduction}
GNNs have become powerful tools for learning from graph-structured data across several domains~\citep{defferrard2016chebnet,velickovic2018GAT,choi2021ltocf,choi2023gread,jeong2025predicting}. 
At the core of this field are MPNN~\citep{Gilmer2017chemi}, which propagates information by iteratively exchanging messages between neighboring nodes. However, MPNNs face limitations, particularly over-smoothing~\citep{nt2019revisiting} and over-squashing~\citep{alon2021oversquashing}. To address these challenges, Transformer architectures~\citep{vaswani2017attention} have been adapted for graph learning, applying self-attention mechanisms to enable long-range interactions by treating all nodes as tokens~\citep{dwivedi2021gt,wu2021graphtrans,choi2024gfsa}.
While graph Transformers effectively capture global context, they often overlook the inherent locality of MPNNs~\citep{xing2024cobformer}. Although approaches such as GraphGPS~\citep{rampavsek2022gps} attempt to combine MPNN and Transformer node representations to balance local and global information, the computational complexity of Transformers remains a challenge.

\begin{figure*}[t]
    \vspace{-1em}
    \centering
    \subfigure[Renormalized graph]{\includegraphics[width=0.33\textwidth]{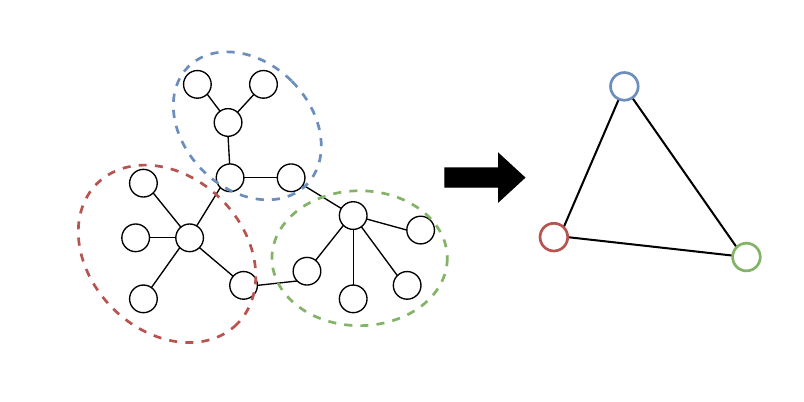}\label{fig:fn-a}}
    % \hfill
    \subfigure[Graph where our fractal nodes are connected]{\includegraphics[width=0.66\textwidth]{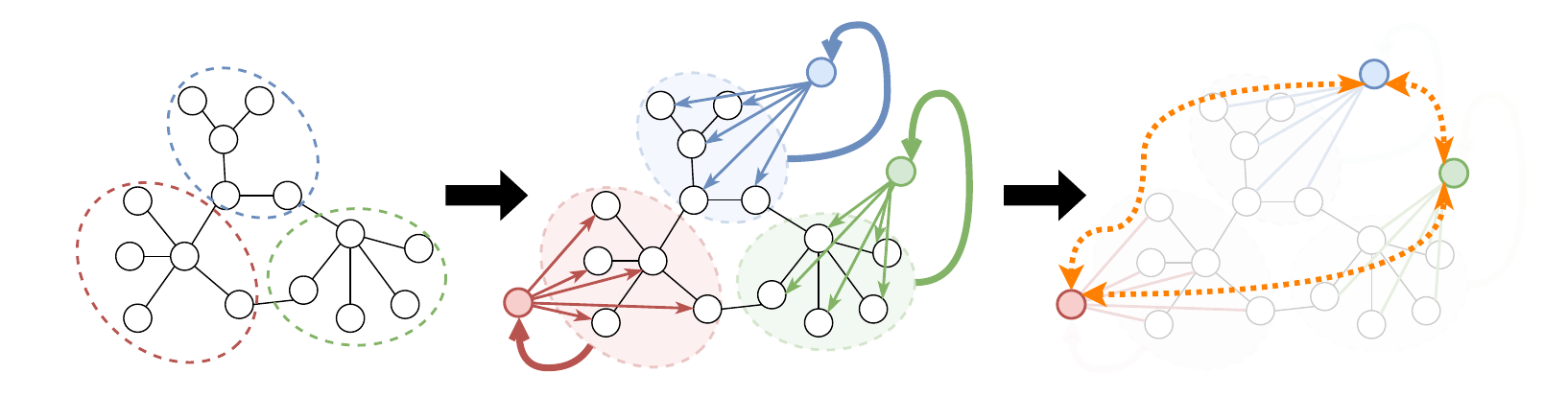}\label{fig:fn-b}}
    % \vspace{-1em}
    \caption{Conceptual comparison of renormalization and our fractal node process. (a) In renormalization, the original graph is replaced by a single node according to each box-covering method, creating a coarsened network. 
    (b) After partitioning the graph, \textit{fractal nodes} (\myrednode, \mybluenode, \mygreennode) are created for each subgraph, which represent the subgraph nodes while having similar features at different scales.
    Then, we propagate the information to the original nodes (our proposed $\mathsf{FN}$). We also support the long-distance interactions (\ORG{orange dashed lines}) between fractal nodes (our proposed $\mathsf{FN}_M$).}
    \label{fig:fn}
\end{figure*}

\paragraph{Motivation.}
The limitations of both MPNN and graph Transformers motivate us to seek a novel approach that balances local and global information while preserving scalability. 
Our inspiration comes from the concept of fractals~\citep{mandelbrot1983fractal}, which describes systems where similar patterns recur across scales. 
Many real-world graphs show this fractal structure, where structural patterns repeat across different parts of the graph~\citep{dill2002selfweb,kim2010fractal,chen2020fractalsocial}.
Fractal structures in graphs are commonly studied through renormalization~\citep{song2005self} (see \Cref{fig:fn-a}), where groups of nodes are coarsened into ``super-nodes'' to analyze global properties emerging from local structures.
In contrast to renormalization, one may consider partitioning the graph into subgraphs and learning representative features for each subgraph --- without changing the original graph topology. This alternative perspective leads us to the following question:

\emph{Can fractal-inspired representations enhance long-range message passing, while preserving the efficiency of MPNNs instead of graph Transformers?}

Our answer is ``yes'', and we introduce our main idea: \emph{fractal nodes} for enforcing feature similarity.

\paragraph{Main idea: graph partitioning causes \emph{fractal structure}, and each \emph{fractal node} enforces feature similarity in each partition.}
We propose a novel concept called `\emph{fractal nodes}', inspired by the fractal structure observed in real-world networks.
We build on the view that graph partitioning naturally causes fractal structure, where subgraphs reflect the connectivity patterns of the full graph.
Rather than replacing groups of nodes with super nodes as in renormalization, we retain the original graph and introduce fractal nodes (\myrednode, \mybluenode, \mygreennode) that coexist with the original nodes (see~\Cref{fig:fn-b}). 
Each fractal node serves as a representative of its subgraph and is connected to all nodes in its subgraph.
These nodes aggregate subgraph-level information and enforce \emph{feature similarity} among the nodes within the same subgraph. To construct these representations, each fractal node adaptively combines both the low-frequency (global) and high-frequency (local) components of the features in its subgraph. This goes beyond simple mean pooling, which only retains the lowest-frequency signal. 
These one-hop shortcut connections between the fractal nodes and their subgraph nodes effectively reduce the information propagation distance. This theoretical reduction in effective resistance between nodes empirically supports the mitigation of over-squashing~\citep{alon2021oversquashing} and improves signal propagation. 
Furthermore, we apply an MLP-Mixer~\citep{tolstikhin2021mlpmixer} at the final layer to flexibly mix the representations of the fractal nodes. This allows long-range interactions between subgraphs without relying on multi-hop message passing, which leads to signal degradation in standard MPNNs.

\paragraph{Contributions.} 
We propose a novel paradigm, \textit{fractal nodes}, which improves message passing in GNNs by enforcing feature similarity within subgraphs, using the fractal structure caused by graph partitioning. Our main contributions are as follows:
\begin{itemize}[leftmargin=10pt]
    \item We introduce \textit{fractal nodes}, a plug-in component that captures subgraph-level information, and integrate them into MPNNs (\Cref{sec:fn}). Our approach is inspired by the fractal nature of networks and discusses the properties of \textit{fractal nodes} (\Cref{sec:properties}).
    \item We analyze the role of fractal nodes in reducing effective resistance and demonstrate that they mitigate the over-squashing problem (\Cref{sec:analysis-os}). We also show improved expressive power compared to standard MPNNs (\Cref{sec:analysis-ex}).
    \item Extensive experiments on various benchmark datasets show that MPNNs augmented with fractal nodes achieve performance comparable to or better than state-of-the-art graph Transformer-based models (\Cref{sec:exp-sota}), while improving the scalability and maintaining efficiency of standard MPNNs (\Cref{sec:runtime}).
\end{itemize}

\section{Preliminaries \& Related Work}
We first introduce the background on MPNNs and their limitations, then discuss related work. Due to space constraints, we provide additional related work in \Cref{app:comp}.

\paragraph{Message passing neural network.}
Given a graph $\mathcal{G}=(\mathcal{V},\mathcal{E})$, we use $\mathcal{V}$ and $\mathcal{E}$ to denote its nodes and edges, respectively.
The nodes are indexed by $v$ and $u$ such that $v,u\in\mathcal{V}$, and an edge connecting the nodes $v$ and $u$ is denoted by $(v,u)\in\mathcal{E}$.
We consider the case where each node $v$ has a hidden vector $h^{(\ell)}_v \in \mathbb{R}^{\text{dim}(h)}$, where $\text{dim}(h)$ is the size of the hidden dimension, and $\ell$ is the number of layers.
MPNNs iteratively update node representations:
\begin{align}
    h_v^{(\ell+1)} = \varphi(h_v^{(\ell)}, \psi^{(\ell)}( \{ h_{u}^{(\ell)} :u \in \mathcal{N}(v) \})),\label{eq:mpnn}
\end{align}
where $\psi^{(\ell)}$ and $\varphi^{(\ell)}$ are the aggregation function and the update function. 

\paragraph{Limitations of MPNNs.}
MPNNs have been investigated for their \textit{expressive power} limitations and \textit{over-squashing} problems. Simple MPNNs are known to be only as powerful as the 1-Weisfeler-Leman graph isomorphism test~\citep{xu2018powerful}.
The over-squashing problem occurs when MPNNs struggle to propagate information along long paths, resulting in loss of information when aggregating from too many neighbors into a fixed-sized node feature vector~\citep{alon2021oversquashing,giovanni2023oversquashing}. 
This over-squashing phenomenon can be precisely characterized through effective resistance~\citep{black2023gtr}. The effective resistance between nodes $u$ and $v$ measures the difficulty of information flow between them:
\begin{align}
    R(u,v)  =  (\mathbf{1}_u - \mathbf{1}_v)^{\mathsf{T}} \mathbf{L}^{+} (\mathbf{1}_u - \mathbf{1}_v),
\end{align}
where $\mathbf{L}^{+}$ is the pseudoinverse of the graph Laplacian and $\mathbf{1}_u$ is the indicator vector for node $u$. Intuitively, higher resistance indicates fewer paths between nodes, creating bottlenecks where information from many neighbors must be compressed into fixed-sized vectors. 
This quantitative framework reveals why local message passing cannot effectively capture both local and global context, leading to the development of graph Transformers, which use self-attention to enable  ``everything is connected to everything''. We use this effective resistance in \Cref{sec:properties} to analyze how fractal nodes alleviate over-squashing while maintaining the efficiency of MPNNs.

\paragraph{Augmented MPNNs.}
To improve information flow and address the limitations of standard MPNNs, various strategies have been proposed~\citep{giovanni2023oversquashing,shi2023exposition}. One approach is incorporating additional global graph features during representation learning~\citep{Gilmer2017chemi,hu2020ogb}. 
The expanded width-aware message passing~\citep{choi2024panda} offers an alternative that mitigates over-squashing without requiring rewiring by selectively expanding the width of high-centrality nodes. Another effective method is rewiring the input graph to enhance connectivity and alleviate structural bottlenecks~\citep{gasteiger2019digl,black2023gtr,karhadkar2023fosr,yu2025piorf}. These adjustments allow for more effective information flow within the network. Another example of graph augmentation is a virtual node. This heuristic, introduced by \citet{Gilmer2017chemi}, has been observed to improve performance on various tasks. Further analyses by \citet{hwang2022vn,cai2023connection_vn,southern2025understanding} have explored the role of virtual nodes in mitigating under-reaching and over-smoothing issues.

\paragraph{Graph Transformers.}
Because of the successes of Transformers, previous works have adapted this architecture for graph learning~\citep{dwivedi2021gt,muller2023attending}.
\citet{dwivedi2021gt} proposed using graph Laplacian eigenvectors as node positional encodings.
GraphGPS~\citep{rampavsek2022gps} provides a general framework combining MPNNs with Transformer components, while
Graphormer\citep{ying2021graphormer} uses attention mechanisms to estimate several types of encoding, such as centrality, spatial, and edge encoding.
\citet{wu2021graphtrans} applies the MPNN directly to all nodes and then applies a Transformer, which is computationally intensive.
\citet{he2023graphViT} generalize ViT~\citep{dosovitskiy2020vit} to graphs and \citet{ma2023GRIT} show that adding inductive biases to graph Transformers removes the need for MPNN modules in GraphGPS.
% Exphormer~\citep{shirzad2023exphormer} improves GraphGPS by using self-attention on expander graphs. 
While graph Transformers effectively capture long-range dependencies, they typically scale quadratically with the number of nodes $\mathcal{O}(|\mathcal{N}|^2)$ compared to the linear scaling of MPNNs $\mathcal{O}(|\mathcal{E}|)$, motivating our work to develop more efficient architectures that preserve global information capture.

\section{Message Passing with Fractal Nodes}\label{sec:fn}
In this section, we propose fractal nodes and explain how they contribute to overcoming the limitations of existing MPNNs. We describe how to create fractal nodes, how to augment the MPNNs, and how to implement interactions between intra and inter-subgraphs.

\paragraph{Notation.}
Let $\{\mathcal{V}_1,\dots,\mathcal{V}_C\}$ be the set of node subsets corresponding to $C$ subgraphs, where $C$ is the number of subgraphs. 
$\mathcal{G}_c = (\mathcal{V}_c, \mathcal{E}_c)$ is the induced subgraph of $\mathcal{G}$.
We define $h^{(\ell)}_{v,c}$ as the hidden vector of node $v$ of the $c$-th subgraph in layer $\ell$, and $f^{(\ell)}_{c}$ as the hidden vector of the fractal node of the $c$-th subgraph in the $\ell$-th layer.

\paragraph{Message passing with \emph{fractal nodes}.} 
We first introduce the message passing process, including fractal nodes. 
The message passing process proceeds as follows:
\begin{align}
    \widetilde{h}_{v,c}^{(\ell+1)} &= \varphi^{(\ell)}(h_{v,c}^{(\ell)}, \psi^{(\ell)}({h_{u,c}^{(\ell)} : u \in \mathcal{N}_v})),\label{eq:up-node}\\
    f_c^{(\ell+1)} &= \varphi^{(\ell)}_{\mathsf{FN}}(f_c^{(\ell)}, \psi_{\mathsf{FN}}^{(\ell)}({\widetilde{h}_{u,c}^{(\ell+1)} : u \in \mathcal{N}_v})),\label{eq:up-fn}\\
    h_{v,c}^{(\ell+1)} &= \widetilde{\varphi}^{(\ell)}(\widetilde{h}_{v,c}^{(\ell+1)}, f_c^{(\ell+1)}),\label{eq:up-final}
\end{align}
where $\mathcal{N}_{v}$ is the set of neighbors of node $v$.
\cref{eq:up-node} performs standard message passing at the node level.
\cref{eq:up-fn} updates the fractal node representations. It aggregates hidden vectors from all nodes in the subgraph, $\mathcal{G}_c$, using the $\widetilde{h}_{u,c}^{(\ell+1)}$, and then updates the fractal nodes.
$\psi^{(\ell)}_{\mathsf{FN}}$ and $\varphi^{(\ell)}_{\mathsf{FN}}$ are aggregate and update functions for fractal nodes, which will be explained in more detail.
The update function $\widetilde{\varphi}^{(\ell)}$ is the step where the message $f_c^{(\ell+1)}$ is propagated to $h^{(\ell)}_{v,c}$.

\paragraph{How to create \emph{fractal nodes}.}
As shown in \Cref{fig:fn-b}, fractal nodes are created from subgraphs. To partition into subgraphs, we consider the METIS~\citep{karypis1998metis} algorithm for its scalability and efficiency. How we use METIS is provided in \Cref{app:metis}.
In our design, each fractal node plays two roles: (i) structurally, it originates from a subgraph that reflects the recurring connectivity patterns of the original graph, and (ii) functionally, it encodes a subgraph-level feature representation that enforces \emph{feature similarity} among the nodes within the same subgraph. While graph partitioning naturally preserves structural characteristics, our method focuses on learning subgraph-level representations by adaptively combining low-frequency (global) and high-frequency (local) components of node features within each subgraph.
To motivate this, we first show that simple mean pooling captures only the direct current (DC) component of the signal.
\begin{theorem}[Mean pooling as a low-pass filter]\label{thr:dc} 
The mean pooling operation applied to the node features is equivalent to extracting the lowest frequency component in the Fourier domain, acting as a low-pass filter that discards all high-frequency information.
\end{theorem}
The mean pooling corresponds to extracting the lowest frequency component --- also known as the DC component --- in the Fourier domain. This DC component captures the global characteristic of the subgraph, but it ignores higher-frequency variations that represent local details. A formal proof of \Cref{thr:dc} is provided in \Cref{app:proof-dc}.

While \Cref{thr:dc} shows that mean pooling only captures the DC component, fractal nodes go beyond this limitation by using $\mathsf{LPF}$ and $\mathsf{HPF}$. We adaptively rescale the high-frequency component, and combine $\mathsf{LPF}$ and $\mathsf{HPF}$ together to form fractal nodes:
\begin{align}
    f_c^{(\ell+1)} = \mathsf{LPF}(\{h_{v,c}^{(\ell+1)}\}_{v \in \mathcal{V}_c}) +  \omega^{(\ell)}_c \cdot \mathsf{HPF}(\{h_{v,c}^{(\ell+1)}\}_{v \in \mathcal{V}_c}),\label{eq:f_c}
\end{align}
where $\omega^{(\ell)}_c$ is a learnable parameter controlling the contribution of high-frequency components. We use a learnable scalar, $\omega^{(\ell)}_c\in\mathbb{R}$, or a learnable vector parameter, $\boldsymbol{\omega}^{(\ell)}_c\in\mathbb{R}^{\textrm{dim}(h)}$.
The $\mathsf{LPF}$ is computed by averaging the node features within the subgraph, so it can capture global information:
\begin{align}
    \mathsf{LPF}(\{h_{v,c}^{(\ell+1)}\}_{v \in \mathcal{V}_c}) := \frac{1}{|\mathcal{V}_c|} \sum_{v\in\mathcal{V}_c} h_{v,c}^{(\ell+1)}.\label{eq:lpf}
\end{align}
\cref{eq:lpf} is analogous to mean pooling and represents the global, low-frequency component of the subgraph. To capture the local details, the $\mathsf{HPF}$ is applied by subtracting the low-pass filtered output from the original node hidden vector. This allows the model to retain the local variations:
\begin{align}
    \mathsf{HPF}(\{h_{v,c}^{(\ell+1)}\}_{v \in \mathcal{V}_c}) := h_{v,c}^{(\ell+1)} - \mathsf{LPF}(\{h_{u,c}^{(\ell+1)}\}_{u \in \mathcal{V}_c}).
\end{align}

\paragraph{Fractal nodes mixing with MLP-Mixer.}
We also allow fractal nodes to exchange messages, as the coarsened network in~\Cref{fig:fn-a} takes advantage of long-distance interactions. To do this, we can apply the MLP-Mixer layer~\citep{tolstikhin2021mlpmixer} to the fractal nodes in the last layer. This means that we do not need to create a coarsened network, and the MLP-Mixer flexibly mixes the representations of fractal nodes:
\begin{align}
    \widetilde{F} = \mathsf{MLPMixer}(F^{(L)}), \;
    F^{(L)} = [f_1^{(L)}, \ldots, f_{C}^{(L)}],\label{eq:mix}
\end{align}
where $F^{(L)}$ is the matrix of all fractal node representations at the final layer $L$.
The MLP-Mixer layer consists of token-mixing and channel-mixing steps:
\begin{align}
U &= F^{(L)} + (W_2\rho(W_1\mathsf{LayerNorm}(F^{(L)}))) \quad \label{eq:mixer1}\\
\widetilde{F}^{(L)} &= U + (W_4\rho(W_3\mathsf{LayerNorm}(U^T)^T),\label{eq:mixer2}
\end{align}
where $\rho$ is a GELU nonlinearity, $\mathsf{LayerNorm}(\cdot)$ is layer normalization, and matrices $W_1, W_2 , W_3, W_4$ are learnable weight matrices.

\paragraph{Instance of our framework.}
We present two variants of our method: $\mathsf{FN}$ and $\mathsf{FN}_M$ (`Fractal Nodes with Mixer'). The key difference is that $\mathsf{FN}$ applies fractal nodes at every layer through the update equations in \cref{eq:up-node,eq:up-fn,eq:up-final}, while $\mathsf{FN}_M$ additionally employs an MLP-Mixer at the final layer (see \cref{eq:mix}) to enable direct communication between fractal nodes across different subgraphs. This allows $\mathsf{FN}_M$ to capture long-range inter-subgraph dependencies more effectively. We provide instantiations with GCN, GINE, and GatedGCN in \Cref{app:fn_detail}.

\paragraph{The output layer.}
Once the final representation $h_G$ is derived, we use a multi-layer perceptron (MLP) as an output layer to predict graph-level outputs: 
\begin{align}
h_G &= \mathsf{MeanPool}({H^{(L)} \text{ for }\mathsf{FN},\; \widetilde{F}^{(L)} \text{ for } \mathsf{FN}_M}), \nonumber\\
y_G &= \mathsf{MLP}(h_G), \nonumber
\end{align}
where $y_G$ is either a scalar for regression tasks or a vector for classification tasks, and $H^{(L)} = [h_1^{(L)}, ..., h_{|V|}^{(L)}]$ is the matrix of node representations at the final layer $L$ for all nodes in the graph.
For details on our method for node classification tasks, please refer to \Cref{app:imp_node}.

\section{Properties of Fractal Nodes}\label{sec:properties}
In this section, we analyze why fractal nodes are effective and what properties they have, discuss the model complexity, and compare them with previous work.

\subsection{Why Fractal Nodes Improve Message Passing}\label{sec:why}
\paragraph{Theoretical analyses.}
We provide theoretical analyses showing that fractal nodes help mitigate over-squashing by reducing the effective resistance between nodes. Full theoretical analyses are provided in \Cref{app:theory}.
\begin{theorem}[Resistance reduction]\label{thm:reduction}
Let $\mathcal{G}$ be the original graph and $\mathcal{G}_f$ be the augmented graph with fractal nodes. For any nodes $u,v \in \mathcal{G}$, the effective resistance in $\mathcal{G}_f$ satisfies:
\begin{equation}
   R_f(u,v) \leq R(u,v),\nonumber
\end{equation}
where $R_f(u,v)$ is the effective resistance in $\mathcal{G}_f$ and $R(u,v)$ is the original effective resistance in $\mathcal{G}$.
\end{theorem}
This reduction in effective resistance improves signal propagation between distant nodes as follows \Cref{thm:signal}.
\begin{theorem}[Signal propagation with fractal nodes]\label{thm:signal}
For an MPNN with fractal nodes, the signal propagation between nodes $u,v$ after $\ell$ layers satisfies:
\begin{align}
    \|h_u^{(\ell)} - h_v^{(\ell)}|| \leq \exp(-\ell/R_f(u,v))||h_u^{(0)} - h_v^{(0)}||,\nonumber
\end{align}
where $R_f(u,v)$ is the effective resistance in $\mathcal{G}_f$.
\end{theorem}
Since $R_f(u,v) \leq R(u,v)$, fractal nodes improve the worst-case signal propagation bound compared to the original graph. \emph{The proofs and detailed analyses can be found in}~\Cref{app:proof-reduction,app:proof-signal}.

\begin{figure}[t!]
\begin{minipage}{.48\columnwidth}
    \centering
    \includegraphics[width=\columnwidth]{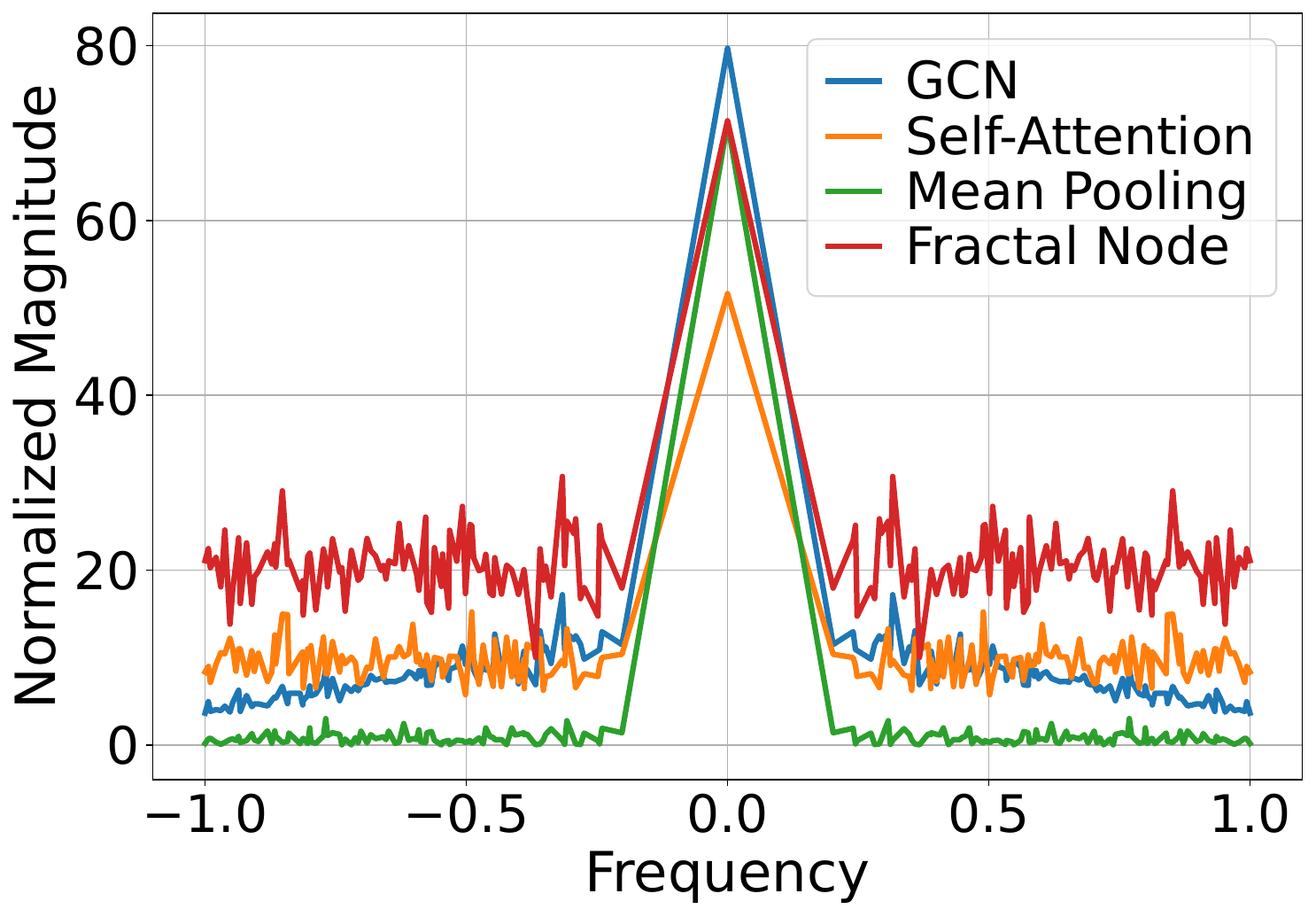}
    \caption{Normalized frequency response on \textsc{Peptides-struct}.}
    \label{fig:filter}
\end{minipage}
\hfill
\begin{minipage}{.50\columnwidth}
    \includegraphics[width=\columnwidth]{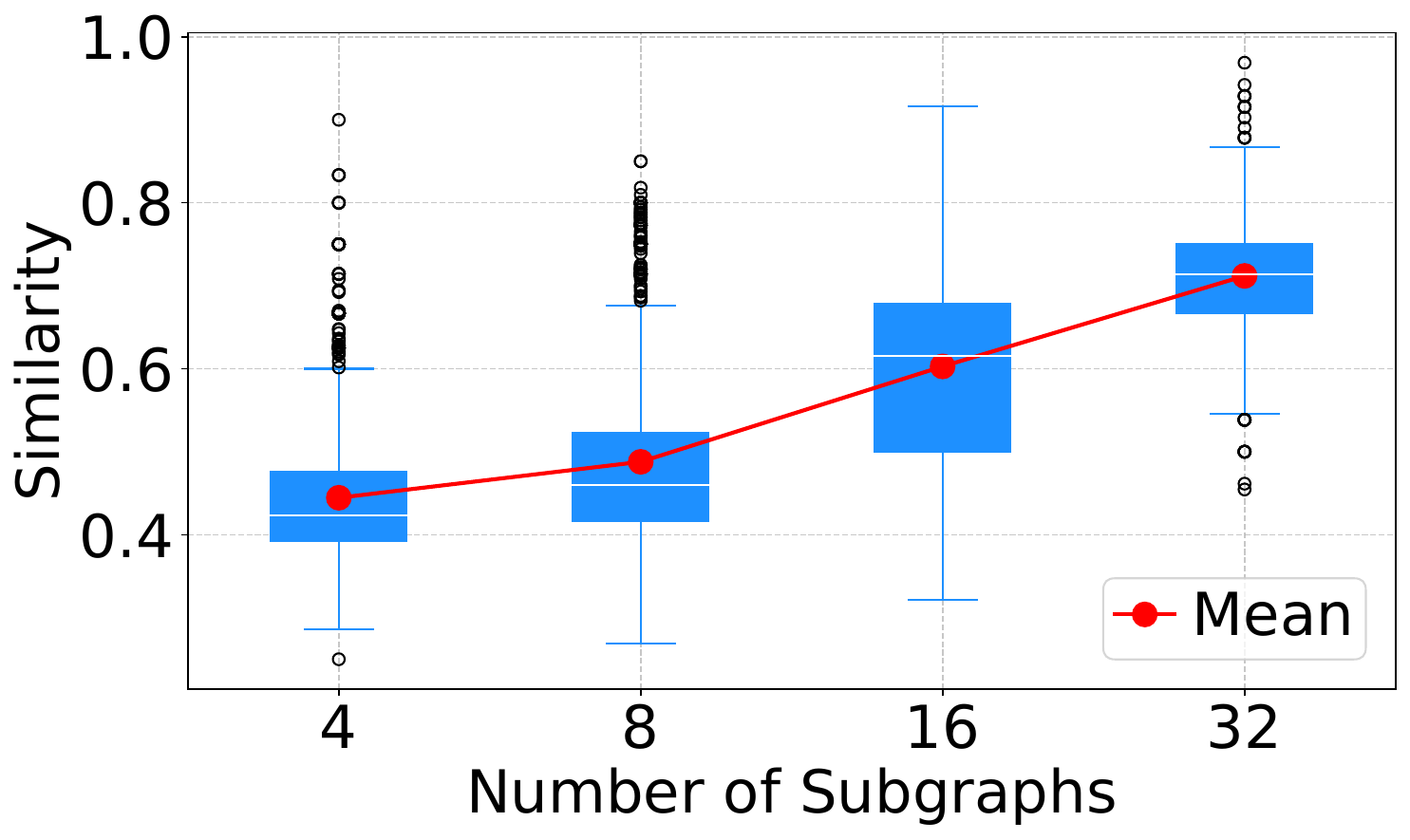}
    \caption{Structural similarity of node centrality distribution in \textsc{Peptides-struct}.}
    % \vspace{1em}
    \label{fig:sim-peptide_struct}
\end{minipage}
\end{figure}
% \paragraph{Spectral perspective: frequency response of fractal nodes.} 
\paragraph{Frequency response analysis.} 
We analyze fractal nodes from a frequency perspective. \Cref{fig:filter} shows normalized frequency responses. Mean pooling shows a minimal response in a high-frequency domain, which suggests an oversimplification of node representations by losing local details. GCN and self-attention have higher responses than mean pooling in the high-frequency domain but are still lower than fractal nodes. The improved high-frequency response of fractal nodes preserves fine local details.

\paragraph{Fractal structure and feature similarity.}
Fractal structure emerges naturally from the graph partitioning process. Since each subgraph is a subset of the original graph, it often reflects the overall connectivity patterns of the full graph. This structural similarity aligns with the concept of fractality, where similar patterns recur across different scales. To quantify this, we compare the betweenness centrality distributions between the original graph and its subgraphs. Betweenness centrality~\citep{freeman1977betweenness} captures both local and global structural importance, particularly in graphs where even low-degree nodes can serve as critical bridges~\citep{kitsak2007betweenness}. As shown in \Cref{fig:sim-peptide_struct}, the structural similarity increases with the number of subgraphs (see \Cref{app:similarity} for details).
On top of this property, our fractal nodes are designed to enforce \emph{feature similarity} within each subgraph. Each fractal node summarizes the features of its corresponding subgraph by adaptively combining low-frequency (global) and high-frequency (local) components using a learnable parameter $\omega^{(\ell)}_c$. While mean pooling only retains global information through DC components, our approach preserves global patterns and local variations in the feature space. 
% This mechanism enforces feature similarity, enhances long-range information flow, and enables our method to perform effective and scalable message passing.

\paragraph{Expressive power of \textit{fractal nodes}.}
The expressive power of fractal nodes can be understood through the lens of existing theoretical results on subgraph-based approaches. 
The methods have been shown to increase expressive power beyond MPNNs.
Encoding local subgraphs is stronger than 1-WL and 2-WL tests~\citep[Theorem 4.3]{zhao2022gnnak}.
In the context of the subgraph WL (SWL) test~\citep{zhang2023subgraphwl}, fractal nodes achieve expressive power comparable to SWL with additional single-point aggregation and potentially approach SWL with additional global aggregation~\citep[Theorem 4.4]{zhang2023subgraphwl}, as the fractal nodes implicitly perform a form of global aggregation within each subgraph. We will empirically verify the expressive power in \Cref{sec:analysis-ex}.

\subsection{Model Complexity}\label{sec:complexity}
Our fractal nodes show improvements in computational efficiency compared to graph Transformers~\citep{dwivedi2021gt,rampavsek2022gps}. The time complexity of our $\mathsf{FN}$ is $\mathcal{O}(L(|\mathcal{V}|+|\mathcal{E}|))$, where $L$ is the number of layers, $|\mathcal{V}|$ is the number of nodes, and $|\mathcal{E}|$ is the number of edges. 
The $\mathsf{FN}_M$ introduces an additional mixing step through the MLP-Mixer, leading to a time complexity of $\mathcal{O}(L(|\mathcal{V}|+|\mathcal{E}|)+Cd^2)$. 
$C$ is the number of subgraphs and $d$ is the hidden dimension. 
Given that $C$ is smaller than $|\mathcal{V}|$, this term does not dominate the overall complexity.
In contrast, graph Transformers incur a time complexity of $\mathcal{O}(L(|\mathcal{V}|^2))$, due to the quadratic cost of computing self-attention over all node pairs.
% , which is expensive for large graphs.
% Similarly, GraphGPS combines MPNNs with self-attention, resulting in comparable quadratic complexity $\mathcal{O}(L(|\mathcal{V}|^2))$. 
% Thus, fractal nodes offer a computational advantage over graph Transformer-based methods.

\section{Experiments}\label{sec:exp}
To evaluate the effectiveness of our fractal nodes, we aim to answer the following key questions:
\textbf{(Q1)} Can fractal nodes mitigate over-squashing in MPNNs?
\textbf{(Q2)} Do fractal nodes improve the expressiveness of MPNNs?
\textbf{(Q3)} How do fractal nodes compare to MPNNs and other graph Transformers regarding performance on benchmark datasets?
\textbf{(Q4)} Does our method lead to a faster run time than graph Transformers?
In this experiment, we aim to verify if fractal nodes provide meaningful benefits. Afterwards, we perform a series of additional studies, ablation studies, and sensitivity analyses. 

\subsection{Fractal Nodes Alleviate Over-squashing (\textbf{Q1})}\label{sec:analysis-os}
\begin{figure}[t!]
\begin{minipage}{.48\columnwidth}
    \centering
    \includegraphics[width=\columnwidth]{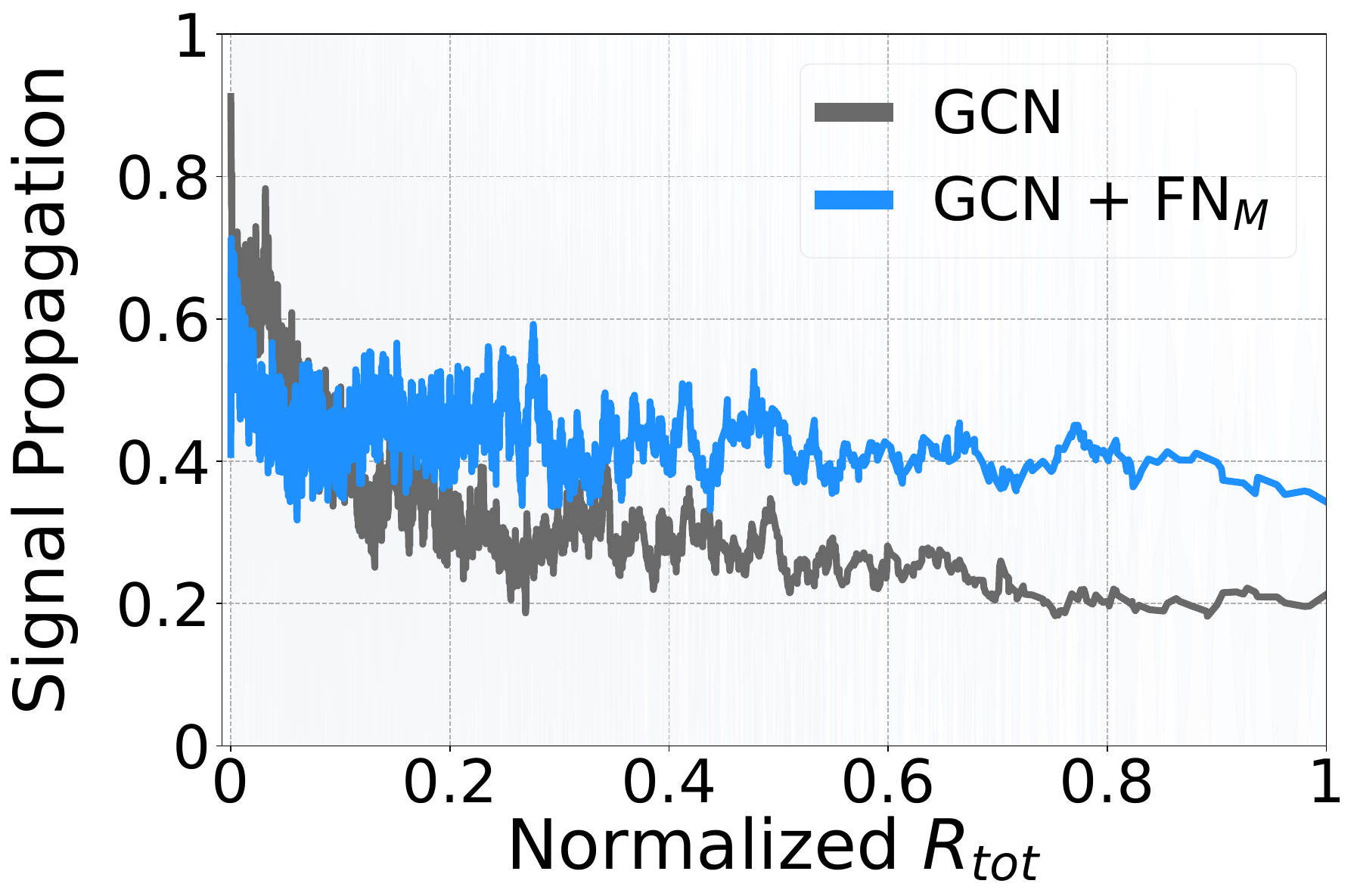}
    \vspace{-1.2em}
    \label{fig:signal}
    % \caption{The amount of signal propagated across the graphs w.r.t. the normalized $R_{tot}$ in \textsc{Peptides-func}. }
    \caption{The amount of signal propagated across the graph in \textsc{Peptides-func}. More results are in \Cref{app:signal_exp}.}
\end{minipage}
\hfill
\begin{minipage}{.48\columnwidth}
    % \vspace{-1em}
    \centering
    \includegraphics[width=\columnwidth]{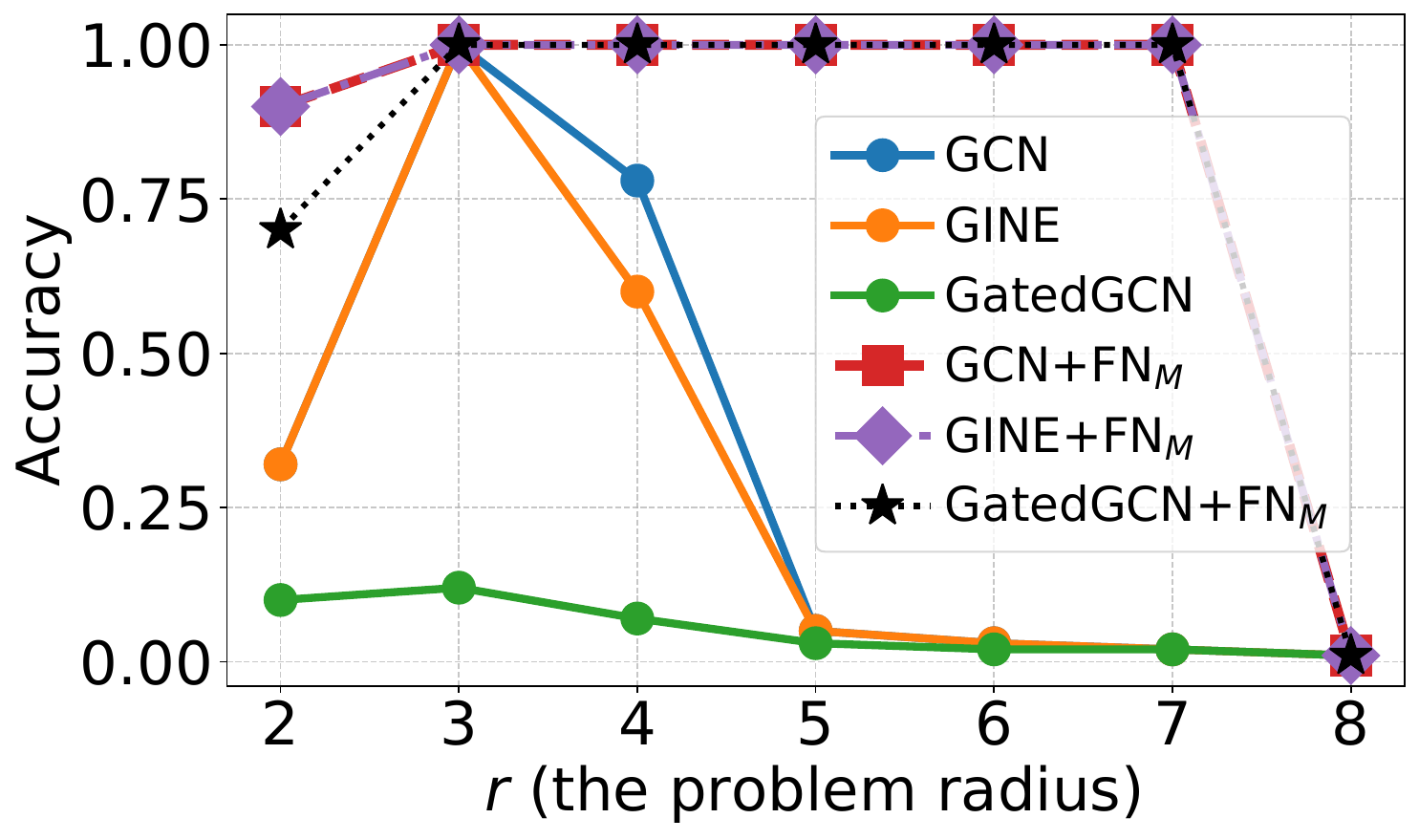}
    % \caption{Test accuracy in the \textsc{TreeNeighbourMatch} problem.}
    \caption{Test accuracy across problem radius (tree depth) in the \textsc{TreeNeighbourMatch} problem.}
    \label{fig:tree}
\end{minipage}
\end{figure}

The signal propagation of MPNNs is inversely proportional to the total effective resistance~\citep{giovanni2023oversquashing}.
As shown in \Cref{fig:signal}, while GCN fails to maintain signal flow under high total effective resistance, indicating severe bottlenecks, GCN+$\mathsf{FN}_M$ shows resilience by maintaining higher signal propagation even under the highest total effective resistance.
This validates our theoretical predictions (\Cref{thm:reduction,thm:signal}) that fractal nodes reduce effective resistance by serving as single-hop shortcuts and enabling long-range interactions through the MLP-Mixer.
We further evaluate \textsc{TreeNeighboursMatch}~\citep{alon2021oversquashing}, where the task requires propagating information from leaf nodes to a target node in binary trees of depth $r$. As shown in \Cref{fig:tree}, standard MPNNs fail for $r>4$, while our methods generalize up to $r=7$, empirically confirming that fractal nodes mitigate over-squashing by enabling long-range interaction.

\begin{table}[t!]
    \centering
    \small
    \setlength{\tabcolsep}{2pt}
    \centering
    \caption{Results on 3 synthetic datasets}
    \label{tab:express}
    \begin{tabular}{l ccc}\toprule
        Method & \textsc{CSL} (Acc $\uparrow$)& \textsc{SR25} (Acc $\uparrow$)& \textsc{EXP}(Acc $\uparrow$)\\\midrule
        GCN      & 10.00 & 6.67 & 52.17\\
        \rowcolor{gray! 10}
        GCN + $\mathsf{FN}_M$ & 39.67 & 100.0 & 86.40\\
        \cmidrule(lr){1-4}
        GINE     & 10.00 & 6.67 & 51.35\\
        \rowcolor{gray! 10}
        GINE + $\mathsf{FN}_M$ & 47.33 & 100.0 & 95.58\\
        \cmidrule(lr){1-4}
        GatedGCN & 10.00 & 6.67 & 51.25\\
        \rowcolor{gray! 10}
        GatedGCN + $\mathsf{FN}_M$ & 49.67 & 100.0 & 96.50 \\\bottomrule
    \end{tabular}
\end{table}

\subsection{Expressive Power of Fractal Nodes (\textbf{Q2})}\label{sec:analysis-ex}

We evaluate the expressive power of fractal nodes on 3 simulated datasets: CSL~\citep{murphy2019csl}, EXP~\citep{Ralph2021exp}, and SR25~\citep{balcilar2021SR25}.
Each dataset contains graphs indistinguishable by the 1 to 3-WL test, and details are provided in \Cref{app:dataset}.
\Cref{tab:express} shows that our model achieves superior accuracy on all 3 datasets while MPNNs fail (see detailed result in \Cref{app:exp_power}).
Our results are empirical, but align with our discussion in \Cref{sec:why}.

\begin{table*}[t]
    \footnotesize
    \setlength{\tabcolsep}{1.5pt}
    \centering
    % \caption{Test performance on 2 peptide datasets from LRGB~\citep{dwivedi2022LRGB} and 4 other benchmark datasets~\citep{hu2020ogb,dwivedi2023benchmarking}. 
    \caption{Test performance on 6 benchmark datasets.
    The top three models are colored by \BEST{first}, \SECOND{second}, \THIRD{third}. Cohen's d effect sizes: $^{\dagger}$ small, $^{\ddagger}$ medium, $^{*}$ large (comparing our best model vs. second-best baseline). $\mathsf{FN}$ and $\mathsf{FN}_M$ consistently show large effects compared to their backbone MPNNs; thus, these annotations are omitted.
    }
    \label{tab:result}
    \begin{tabular}{l cccccc}\toprule
        \multirow{2}{*}{Method} & \textsc{Peptides-func}   & \textsc{Peptides-struct} & \textsc{MNIST} & \textsc{CIFAR10} & \textsc{MolHIV} & \textsc{MolTox21}\\ \cmidrule(lr){2-2}\cmidrule(lr){3-3}\cmidrule(lr){4-4}\cmidrule(lr){5-5}\cmidrule(lr){6-6}\cmidrule(lr){7-7}
        & AP $\uparrow$ & MAE $\downarrow$ & Accuracy $\uparrow$ & Accuracy $\uparrow$ & ROCAUC $\uparrow$ &  ROCAUC $\uparrow$\\ \midrule
        GCN~\citep{kipf2017GCN}    & 0.6328\std{0.0023} & 0.2758\std{0.0012} &  0.9269\std{0.0023} & 0.5423\std{0.0056} & 0.7529\std{0.0098} & 0.7525\std{0.0031}\\
        GINE~\citep{xu2018GIN}    & 0.6405\std{0.0086} & 0.2780\std{0.0021} &  0.9705\std{0.0023} & 0.6131\std{0.0035} & 0.7885\std{0.0034} & 0.7730\std{0.0064}\\
        GatedGCN~\citep{bresson2017GatedGCN}    & 0.6300\std{0.0029} & 0.2778\std{0.0017} &  0.9776\std{0.0017} & 0.6628\std{0.0017} & 0.7874\std{0.0119} & 0.7641\std{0.0057}\\ 
        \cmidrule(lr){1-7}
        GT~\citep{dwivedi2021gt} &  - & - & 0.9083\std{0.0016} & 0.5975\std{0.0029} & 0.7350\std{0.0040} & 0.7500\std{0.0060}\\
        GraphiT~\citep{mialon2021graphit} &  - & -  & - & - & 0.7460\std{0.0100} & 0.7180\std{0.0130}\\
        Graphormer~\citep{ying2021graphormer} &  - & - & - & - & 0.7930\std{0.0040} & 0.7730\std{0.0800}\\
        Transformer + LapPE~\citep{Kreuzer20021SAN} & 0.6326\std{0.0126} & 0.2529\std{0.0016} & 0.9083\std{0.0016} & 0.5975\std{0.0029} & - & 0.7323\std{0.0057}\\ 
        SAN + LapPE~\citep{Kreuzer20021SAN} & 0.6384\std{0.0121} & 0.2683\std{0.0043} & - & - & 0.7775\std{0.0061} & 0.7130\std{0.0080} \\ 
        % SAN + RWSE & 0.6439\std{0.0075} & 0.2545\std{0.0012}\\ 
        EGT~\citep{hussain2022egt} & - & - & 0.9817\std{0.0009} & 0.6870\std{0.0041} & - & -\\ 
        GraphGPS~\citep{rampavsek2022gps} & 0.6534\std{0.0091} & 0.2509\std{0.0014} & 0.9805\std{0.0013} & 0.7230\std{0.0036} & 0.7880\std{0.0101} & 0.7570\std{0.0040}\\ 
        GRIT~\citep{ma2023GRIT} & \SECOND{0.6988\std{0.0082}} & 0.2460\std{0.0012}  & 0.9811\std{0.0011} & \BEST{0.7647\std{0.0089}} & - & - \\ 
        Graph-ViT/MLP-Mixer~\citep{he2023graphViT}  & 0.6970\std{0.0080} & \THIRD{0.2449\std{0.0016}} & \SECOND{0.9846\std{0.0009}} & 0.7158\std{0.0009} & 0.7997\std{0.0102} & \THIRD{0.7910\std{0.0040}}\\
        Exphormer~\citep{shirzad2023exphormer}       & 0.6527\std{0.0043} & 0.2481\std{0.0007} & \THIRD{0.9841\std{0.0035}} & \THIRD{0.7469\std{0.0013}} & - & - \\ 
        GECO~\citep{sancak2024GECO} & \THIRD{0.6975\std{0.0025}} &  0.2464\std{0.0009} & - & - & 0.7980\std{0.0200} & - \\ 
        \cmidrule(lr){1-7}
        CRaWl~\citep{tonshoff2023crawl} & 0.6963\std{0.0079} &  0.2506\std{0.0022} & 0.9794\std{0.0050} & 0.6901\std{0.0026} & 0.7707\std{0.1490} & - \\
        PNA~\citep{corso2020pna}    & - & - &  0.9794\std{0.0012} & 0.7035\std{0.0063} & 0.7905\std{0.0132} & -\\
        GNN-AK+~\citep{zhao2022gnnak}    & 0.6480\std{0.0075} & 0.2736\std{0.0012} & - & 0.7219\std{0.0013} & 0.7961\std{0.0119} & -\\ 
        SUN~\citep{Frasca2022SUN}        & 0.6730\std{0.0115} & 0.2498\std{0.0008} & - & - & 0.8003\std{0.0055} &- \\
        CIN~\citep{bodnar2021CIN}        & - & - & - & - & \THIRD{0.8094\std{0.0057}} & - \\ 
        \cmidrule(lr){1-7}
        % GCN + virtual node~\citep{hu2020ogb} & - & - & - & - & 0.7599\std{0.0119} & 0.7551\std{0.0100} \\ 
        % GIN + virtual node~\citep{hu2020ogb} & - & - & - & - & 0.7707\std{0.0149} & 0.7621\std{0.0062} \\ 
        % % \cmidrule(lr){1-7}
        % GCN + VN~\citep{rosenbluth2024distinguished} & 0.6732\std{0.0066} & 0.2505\std{0.0022} & - & - &  -  & - \\ 
        % GatedGCN + VN~\citep{rosenbluth2024distinguished} & 0.6823\std{0.0069} & 0.2475\std{0.0018} & - & - & - & - \\  
        % \cmidrule(lr){1-7}
        \rowcolor{gray! 10}
        GCN + $\mathsf{FN}$ & 0.6802\std{0.0043} & 0.2530\std{0.0004} &  0.9393\std{0.0084} & 0.6006\std{0.0070} & 0.7564\std{0.0059} & 0.7670\std{0.0073} \\ 
        \rowcolor{gray! 10}
        GINE + $\mathsf{FN}$ & 0.6815\std{0.0059} & 0.2515\std{0.0020} & 0.9790\std{0.0012} & 0.6584\std{0.0069} & 0.7890\std{0.0104} & 0.7751\std{0.0029}\\ 
        \rowcolor{gray! 10}
        GatedGCN + $\mathsf{FN}$ & 0.6778\std{0.0056} & 0.2536\std{0.0019} & 0.9826\std{0.0012} & 0.7125\std{0.0035} & 0.7967\std{0.0098} & 0.7759\std{0.0054}\\ 
        \rowcolor{gray! 10}
        \cmidrule(lr){1-7}
        \rowcolor{gray! 10}
        GCN + $\mathsf{FN}_M$ & 0.6787\std{0.0048} & 0.2464\std{0.0014} & 0.9455\std{0.0004} & 0.6413\std{0.0068} & 0.7866\std{0.0034} & 0.7882\std{0.0041}\\
        \rowcolor{gray! 10}
        GINE + $\mathsf{FN}_M$ & \BEST{0.7018\std{0.0074}}$^{\dagger}$  & \BEST{0.2446\std{0.0018}} & 0.9786\std{0.0004} & 0.6672\std{0.0068} & \BEST{0.8127\std{0.0076}}$^{\ddagger}$ &  \BEST{0.7926\std{0.0021}}\\ 
        \rowcolor{gray! 10}
        GatedGCN + $\mathsf{FN}_M$ & 0.6950\std{0.0047} & \SECOND{0.2453\std{0.0014}} & \BEST{0.9848\std{0.0005}}$^{\dagger}$  & \SECOND{0.7526\std{0.0033}} & \SECOND{0.8097\std{0.0047}} & \SECOND{0.7922\std{0.0054}}\\
 \bottomrule
    \end{tabular}
    \vspace{-1em}
\end{table*}

\subsection{Experiments on Graph Benchmarks  (\textbf{Q3})}\label{sec:exp-sota}
\paragraph{Experimental setting and baselines.}
We evaluate our method on two tasks: graph-level prediction and large-scale node classification.
For graph-level tasks, we use 6 benchmark datasets: 2 peptide datasets from LRGB~\citep{dwivedi2022LRGB}, 2 graph-level super-pixel image datasets from Benchmarking GNNs~\citep{dwivedi2023benchmarking}, and 2 molecular datasets from OGB~\citep{hu2020ogb}.
For large-scale node classification, we use 2 large datasets from OGB~\citep{hu2021ogb}: ogbn-arxiv and ogbn-products.
We compare our fractal nodes to MPNNs, graph Transformer-based models, and other state-of-the-art models. 
Detailed experimental settings for graph-level tasks and large-scale node classification are provided in~\Cref{app:exp_detail,app:ogbn}, respectively.
% Detailed experimental settings for graph-level tasks are provided in~\Cref{app:exp_detail}, while the setup and baseline comparisons for the large-scale node classification experiments are described separately in~\Cref{app:ogbn}.

\begin{table}[t!]
    % \vspace{-2em}
    \footnotesize
    \centering
    \setlength{\tabcolsep}{1.2pt}
    \caption{Results on large-scale graphs}
    \label{tab:ogbn-result}
    \begin{tabular}{l cc}
        \toprule
        Method  & ogbn-arxiv  & ogbn-product \\
        \cmidrule(lr){1-3}
        \# nodes & 169,343 & 2,449,029\\
        \# edges & 1,166,243 & 61,859,140\\
        \midrule
        % MLP & 55.50\std{0.33} &  71.59\std{0.71} \\ 
        % LINKX~\citep{lim2022LINKX} & 66.18\std{0.33} &  71.59\std{0.71} \\ \cmidrule(lr){1-3}
        GraphGPS~\citep{rampavsek2022gps} & 70.97\std{0.41} &  OOM\\
        NAGphormer~\citep{chen2023nagphormer} & 70.13\std{0.55} &  73.55\std{0.21} \\
        Exphormer~\citep{shirzad2023exphormer} & 72.44\std{0.28} &  OOM\\
        NodeFormer~\citep{wu2022nodeformer} & 69.86\std{0.25} &  72.93\std{0.13} \\
        DiffFormer~\citep{wu2023difformer} & 72.41\std{0.40} &  74.16\std{0.31} \\
        PolyNormer~\citep{deng2024polynormer} & 71.82\std{0.23} &  \THIRD{82.97\std{0.28}} \\
        SGFormer~\citep{wu2023sgformer} & 72.63\std{0.13} &  74.16\std{0.31}  \\ \cmidrule(lr){1-3}
        HC-GNN~\citep{zhong2023hcgnn} & 72.79\std{0.25} &  -  \\ 
        ANS-GT~\citep{cai2021ansgt} & 72.34\std{0.50} &  80.64\std{0.29}  \\ 
        HSGT~\citep{zhu2023hsgt}   & 72.58\std{0.31} &  81.15\std{0.13}  \\ 
        \cmidrule(lr){1-3}
        GCN & 71.74\std{0.29} &  75.64\std{0.21} \\
        \rowcolor{gray! 10}
        GCN + $\mathsf{FN}$ & \BEST{73.03\std{0.37}}$^{\ddagger}$ & 81.29\std{0.21} \\
        \rowcolor{gray! 10}
        GCN + $\mathsf{FN}_M$ & \SECOND{72.93\std{0.35}} & 81.33\std{0.33} \\ \cmidrule(lr){1-3}
        GraphSAGE & 71.49\std{0.27} &  78.29\std{0.16} \\
        \rowcolor{gray! 10}
        GraphSAGE + $\mathsf{FN}$ & \THIRD{72.70\std{0.11}}  & \SECOND{83.07\std{0.35}} \\
        \rowcolor{gray! 10}
        GraphSAGE + $\mathsf{FN}_M$ & 72.54\std{0.30}  & \BEST{83.11\std{0.07}}$^{\ddagger}$\\
        \bottomrule
    \end{tabular}
    \vspace{-2em}
\end{table}

\begin{table*}[t!]
    \footnotesize
    % \setlength{\tabcolsep}{5pt}
    % \scriptsize
    % \setlength{\tabcolsep}{1.2pt}
    \setlength{\tabcolsep}{2pt}
    \centering
    \caption{Comparison to virtual node methods.}
    \label{tab:vn}
    \begin{tabular}{l cccc}\toprule
        \multirow{2}{*}{Method} & \textsc{Peptides-func}   & \textsc{Peptides-struct} & \textsc{MolHIV} & \textsc{MolTox21}\\ \cmidrule(lr){2-2}\cmidrule(lr){3-3}\cmidrule(lr){4-4}\cmidrule(lr){5-5}
        & AP $\uparrow$ & MAE $\downarrow$ & ROCAUC $\uparrow$ &  ROCAUC $\uparrow$\\ \midrule
        GCN + virtual node~\citep{hu2020ogb} & 0.6455\std{0.0020} & 0.2745\std{0.0013} & 0.7599\std{0.0119} & 0.7551\std{0.0100} \\ 
        GIN + virtual node~\citep{hu2020ogb} & - & - & 0.7707\std{0.0149} & 0.7621\std{0.0062} \\ 
        \cmidrule(lr){1-5}
        GCN + VN~\citep{rosenbluth2024distinguished} & 0.6732\std{0.0066} & 0.2505\std{0.0022} &  -  & - \\ 
        GatedGCN + VN~\citep{rosenbluth2024distinguished} & \THIRD{0.6823\std{0.0069}} & 0.2475\std{0.0018} & - & - \\ 
        \cmidrule(lr){1-5}
        GatedGCN + $\text{VN}_G$~\citep{southern2025understanding} & 0.6822\std{0.0052} & 0.2458\std{0.0006} & \THIRD{0.7910\std{0.0086}} & 0.7655\std{0.0060} \\ 
        \cmidrule(lr){1-5}
        \rowcolor{gray! 10}
        GCN + $\mathsf{FN}_M$ & 0.6787\std{0.0048} & \THIRD{0.2464\std{0.0014}} & 0.7866\std{0.0034} & \THIRD{0.7882\std{0.0041}}\\
        \rowcolor{gray! 10}
        GINE + $\mathsf{FN}_M$ & \BEST{0.7018\std{0.0074}}$^{*}$ & \BEST{0.2446\std{0.0018}}$^{\dagger}$ & \BEST{0.8127\std{0.0076}}$^{*}$ &  \BEST{0.7926\std{0.0021}}$^{*}$\\ 
        \rowcolor{gray! 10}
        GatedGCN + $\mathsf{FN}_M$ & \SECOND{0.6950\std{0.0047}} & \SECOND{0.2453\std{0.0014}} & \SECOND{0.8097\std{0.0047}} & \SECOND{0.7922\std{0.0054}}\\
 \bottomrule
    \end{tabular}
    \vspace{-1em}
\end{table*}
\paragraph{Results on graph-level tasks.}
Our proposed fractal nodes ($\mathsf{FN}$ and $\mathsf{FN}_M$) consistently enhance the performance of baseline MPNNs on all benchmark datasets, often surpassing graph Transformer models. In \Cref{tab:result}, on \textsc{Peptides-func}, GINE+$\mathsf{FN}_M$ achieves an average precision (AP) of 0.7018, outperforming both Exphormer and GraphGPS.
Our comparable performance with Graph-ViT and Exphormer on \textsc{MNIST} shows that fractal nodes can effectively capture local and global information without a self-attention layer.
\paragraph{Results on large-scale graphs.}
The effectiveness of our method is particularly evident in large-scale graph experiments in~\Cref{tab:ogbn-result}. On ogbn-arxiv, GCN+$\mathsf{FN}$ improves accuracy from 71.74\% to 73.03\%, while on ogbn-product, GraphSAGE+$\mathsf{FN}_M$ demonstrates a substantial improvement.
Exphormer and GraphGPS fail to scale to ogbn-products due to their complexity in attention computation. In contrast, our method maintains computational efficiency while achieving superior performance, which we discuss further in~\Cref{sec:runtime}.
This highlights the effectiveness of fractal nodes in capturing local and graph information and their practical applicability to large-scale graphs.

\subsection{Runtime Comparison (Q4)}\label{sec:runtime}
As discussed in~\Cref{sec:complexity}, our fractal nodes benefit from capturing long-range dependencies without increasing computational complexity.
As shown in~\Cref{tab:profile-large}, fractal nodes introduce trivial computational overhead -- GCN+$\mathsf{FN}$ maintains identical training time and memory usage compared to the GCN. Our method uses common MPNN operations without introducing complex additional computations. 
In contrast, graph Transformers (e.g., GraphGPS and Exphormer) require more computational resources due to their attention mechanisms. 
Given these results shown in~\Cref{sec:exp-sota,sec:runtime}, we believe our method achieves a balance between accuracy and computational efficiency.

In~\Cref{app:scale}, we also conduct experiments on synthetic Erd\H{o}s-R\'enyi graphs with nodes ranging from 1,000 to 100,000 to evaluate efficiency and scalability, and we can verify linear space complexity while our GPU memory usage scaled linearly with the graph size. In \Cref{tab:runtime-partition} of \Cref{app:partition}, we show that using METIS with $\mathcal{O}(|\mathcal{E}|)$ complexity enables efficient fractal node creation.

\begin{table}[t!]
    \small
    \centering
    \setlength{\tabcolsep}{4pt}
    \caption{Training time per epoch and memory usage on ogbn-arxiv}
    \label{tab:profile-large}
    \begin{tabular}{l cc}
        \toprule
        Method & Runtime (s) & Mem. (GB)\\
        \midrule
        GCN & 1.27 & 16.49 \\
        % GraphSAGE & 0.55 & 7.74 \\
        \cmidrule(lr){1-3}
        GraphGPS & 1.32 & 38.91 \\
        Exphormer & 0.74 & 34.04  \\
        NodeFormer & 1.20 & 16.30 \\
        DiffFormer & 0.77 & 24.51 \\
        PolyNormer & 0.31 & 16.09 \\
        \cmidrule(lr){1-3}
        \rowcolor{gray! 10}
        GCN + $\mathsf{FN}$ & 1.27 & 16.49 \\
        \rowcolor{gray! 10}
        GCN + $\mathsf{FN}_M$ & 1.27 & 16.49 \\
        \rowcolor{gray! 10}
        GraphSAGE + $\mathsf{FN}$ & 0.57 & 7.74 \\
        \rowcolor{gray! 10}
        GraphSAGE + $\mathsf{FN}_M$ & 0.58 & 7.76 \\
        \bottomrule
    % \vspace{-2.2em}
\end{tabular}
\end{table}

\begin{table}[t!]
    \footnotesize
    \centering
    \setlength{\tabcolsep}{1.2pt}
    % \caption{Comparison to rewiring methods}
    \caption{Results of GCN with rewiring methods.}
    \label{tab:rewiring}
    \begin{tabular}{l cc}\toprule
        % \multirow{2}{*}{Method} & \textsc{Peptides-func} & \textsc{Peptides-struct} \\ \cmidrule(lr){2-2}\cmidrule(lr){3-3}
        % Method & \textsc{Pep.-func} & \textsc{Pep.-struct} \\ 
        \multirow{2}{*}{Method} & \textsc{Pep.-func} & \textsc{Pep.-struct} \\ \cmidrule(lr){2-2}\cmidrule(lr){3-3}
        & AP $\uparrow$ & MAE $\downarrow$\\
        \midrule
        GCN  & 0.5930\std{0.0023} & 0.3496\std{0.0013} \\ 
         + FoSR (Karhadkar et al. 2023)    & 0.5947\std{0.0035} & 0.3473\std{0.0007} \\ 
        % \; + FoSR~\citep{karhadkar2023fosr}    & 0.5947\std{0.0035} & 0.3473\std{0.0007} \\ 
         + GTR~\citep{black2023gtr}      & 0.5075\std{0.0029} & 0.3618\std{0.0010} \\ 
         + SDRF~\citep{topping2022riccurvature}    & 0.5947\std{0.0126} & 0.3478\std{0.0013} \\ 
         + BORF~\citep{nguyen2023borf}    & 0.5994\std{0.0037} & 0.3514\std{0.0009} \\ 
         + PANDA~\citep{choi2024panda} &  \THIRD{0.6028\std{0.0031}} & \THIRD{0.3272\std{0.0001}} \\ 
         + LASER~\citep{barbero2023laser} &  \SECOND{0.6440\std{0.0010}} & \SECOND{0.3043\std{0.0019}} \\ 
        \cmidrule(lr){1-3}
        \rowcolor{gray! 10}
         + $\mathsf{FN}$ & \BEST{0.6445\std{0.0057}} & \BEST{0.2535\std{0.0012}}$^{*}$ \\ 
        \bottomrule
    \end{tabular}
\end{table}

\subsection{Ablation, Sensitivity, and Additional Studies}
We report ablation studies for $\omega^{(\ell)}_c$ and $\mathsf{HPF}$ in \Cref{app:omega,app:hpf}. We report results when $\omega^{(\ell)}_c$ is zero, that is, without $\mathsf{HPF}$, and when we use either a scalar parameter or a learnable vector parameter.
We also report sensitivity studies on $C$, i.e., the number of fractal nodes, and additional analyses on a variant of message passing between fractal nodes and the distribution of subgraph size ratios in~\Cref{app:sens-c,app:all-fn,app:ratio}. 
An analysis of the use of partitioning algorithms other than METIS, such as random, Louvain~\citep{blondel2008louvain}, Girvan-Newman~\citep{girvan2002community} partitioning, is reported in~\Cref{app:partition}.

% \paragraph{MPNNs with fractal nodes vs. other augmented MPNNs.}
% We compare our fractal nodes to augmented MPNN methods in \Cref{tab:rewiring,tab:vn}
% In \Cref{tab:rewiring}, we show that our $\mathsf{FN}$ outperforms LASER and PANDA 6 baselines on both datasets. We choose the hidden dimension to respect the 500k parameter budget. In our fractal node, we opt out of the positional encodings for a fair comparison.
% If there is only one fractal node and no subgraph is created, our method can be reduced to the virtual node method. To compare with existing virtual node methods, we compare with the virtual node method. 
% As shown in~\Cref{tab:vn}, our methods show superior results to several virtual node methods.

\paragraph{Comparison with augmented MPNNs.}
We compare fractal nodes against two categories of augmented MPNNs: graph rewiring and virtual node approaches. \Cref{tab:rewiring} compares our method with 6 methods on 2 \textsc{Peptides} datasets. 
We replicate the experimental settings of \citet{dwivedi2022LRGB} and use the results from \citet{barbero2023laser}. All methods use the same 500k parameter budget without positional encodings for fair comparison.
Our \textsf{FN} achieves competitive performance, outperforming all baselines, including LASER. 
When using a single fractal node ($C=1$), our method conceptually reduces to a virtual node that aggregates global information. However, as shown in \Cref{tab:vn}, our $\mathsf{FN}_M$ consistently outperforms all virtual node methods. Particularly, GINE+$\mathsf{FN}_M$ achieves the best results on all metrics. This shows that our frequency-based aggregation provides more effective representations than virtual nodes.

% To provide a comprehensive comparison with existing virtual node methods, we compare with the 3 virtual node methods by \citet{hu2020ogb} (denoted as `virtual node'), \citet{rosenbluth2024distinguished} (denoted as  `VN') and \citet{southern2025understanding} (denoted as  `$\text{VN}_G$').
% As shown in \Cref{tab:vn}, both $\mathsf{FN}$ and $\mathsf{FN}_M$ outperform the GCN and GIN models augmented with virtual nodes from \citet{hu2020ogb} on \textsc{MolHIV} and \textsc{MolTox21}. On the Peptides datasets, our methods show competitive results with the VN method of \citet{rosenbluth2024distinguished}.

\section{Concluding Remark}\label{sec:conclusion}
We introduced \emph{fractal nodes} that preserve the computational efficiency of MPNNs while allowing long-range interactions that graph Transformers typically capture. Our core idea is that graph partitioning naturally induces \emph{fractal structure}, and fractal nodes enforce \emph{feature similarity} within each subgraph. This design alleviates over-squashing and improves expressive power through subgraph-level representations. Experiments show that fractal nodes consistently improve MPNN performance and achieve competitive results with graph Transformers.
\paragraph{Limitations and future directions.}
While fractal nodes are effective, they rely on graph partitioning, which may not capture optimal clustering for all graph types. 
While we show robustness across different partitioning methods in \Cref{app:partition}, learnable partitioning could improve performance.
The fixed number of fractal nodes also limits adaptability to varying graph structures. 
Despite these limitations, our extensive experiments demonstrate that fractal nodes offer a practical and effective alternative to graph Transformers for most real-world applications.

\section*{Acknowledgements}
Noseong Park is the corresponding author.
This work was partly supported by the Institute for Information \& Communications Technology Planning \& Evaluation (IITP) grants funded by the Korean government (MSIT) (No. RS-2024-00457882, AI Research Hub Project; No. RS-2025-25442149, LG AI STAR Talent Development Program for Leading Large-Scale Generative AI Models in the Physical AI Domain), Samsung Electronics Co., Ltd. (No. G01240136, KAIST Semiconductor Research Fund (2nd)), and an ETRI grant funded by the Korean government (No. 24ZB1100, Core Technology Research for Self-Improving Integrated Artificial Intelligence System).

\bibliography{reference}
\clearpage

\appendix
\addcontentsline{toc}{section}{Appendix} % Add the appendix text to the document TOC
\part{\LARGE{\textit{Appendix}}} % Start the appendix part
% \parttoc % Insert the appendix TOC
% \clearpage

% \section{Broader Impact Statement}~\label{app:impact}
% In terms of the broader impact of this research on society, we do not see the very negative impacts that might be expected. However, adverse or malicious applications of the proposed algorithms in various domains, including drug discovery and healthcare, may lead to undesirable effects.

% \section{Reproducibility Statement}~\label{app:reproduce}
% To ensure reproducibility and completeness, we have included appendices in this paper.
% \Cref{app:proof-dc} provides a proof of \Cref{thr:dc}.
% \Cref{app:proof-reduction,app:proof-signal} also provide proofs of \Cref{thm:reduction,thm:signal} to ensure the completeness of the theoretical analyses in \Cref{sec:why}.
% We provide details of our experiments presented in the paper in \Cref{app:exp_detail}.
% Only a part of the source code that reproduces the experiments is available at \url{https://sites.google.com/view/fractalnode/}.
% We plan to make all the code available after acceptance.

\section{Proof of \Cref{thr:dc}}\label{app:proof-dc}
\begin{customthm}[Mean pooling as a low-pass filter]
The mean pooling operation applied to the node features is equivalent to extracting the lowest frequency component in the Fourier domain, acting as a low-pass filter that discards all high-frequency information.
\end{customthm}

\begin{proof}
    The mean pooling operation aggregated the features of all nodes in the subgraph or graph by computing the average, 
    \begin{align}
    f^{mean}_c=\frac{1}{n}\sum_{v\in\mathcal{C}_c}h_v.
    \end{align}
    To understand this operation in the frequency domain, we use the discrete Fourier transform (DFT), which transforms the node feature matrix $H_c$ into its frequency domain. 
    The DFT of a signal $h_v$ is represented as: 
    \begin{align}
    \mathcal{F}(h_v) = \textsf{DFT}\cdot h_v,
    \end{align}
    where $\textsf{DFT}\in\mathbb{C}^{n\times n}$ is the Fourier matrix.
    The rows of the Fourier matrix are given by the Fourier basis vectors, which are complex exponential functions. These basis vectors represent different frequencies, and each row in the DFT corresponds to a specific frequency component.
    The first row of the Fourier matrix $\textsf{DFT}$ corresponds to the DC component, which is the lowest frequency component of the signal. This row is a vector of ones:
    \begin{align}
    \textsf{DFT}[1, :] = \frac{1}{\sqrt{n}} \cdot [1, 1, \dots, 1].    
    \end{align}
    This row corresponds to the mean or average of the signal. Therefore, when we project the input signal onto this basis vector, we are effectively extracting the global, smooth structure of the signal.
    
    The DC component of the DFT is then expressed as:
    \begin{align}
        DC[x] = \textsf{DFT}^{-1} \text{diag}(1,0,\ldots,0) \textsf{DFT} x = \frac{1}{n}11^{\intercal}x.
    \end{align}
    This operation corresponds to projecting the input signal $x$ onto the vector of ones, effectively averaging all elements of $x$, which is exactly the result of mean pooling:
    \begin{align}
        f^{DC}_c = \frac{1}{n}11^{\intercal}H_c = \frac{1}{n}\sum_{v\in\mathcal{C}_c}h_v.
    \end{align}
\end{proof}
Therefore, mean pooling captures the DC component of the signal, which is the lowest frequency component. This corresponds to extracting the global, smooth node features of the subgraph, but it does not retain higher-frequency variations, which represent the local details.

Thus, mean pooling is equivalent to applying a low-pass filter that only retains the DC component of the signal.

\section{Theoretical Analyses}\label{app:theory}
In this section, we provide theoretical analyses of fractal nodes to show how they mitigate over-squashing. Our analysis builds on effective resistance theory to characterize information flow in networks with fractal nodes.

\paragraph{Preliminaries on effective resistance.}
For a connected, non-bipartite graph, the pseudoinverse of the normalized Laplacian can be expressed as~\citep{black2023gtr}:
\begin{align}
\hat{\mathbf{L}}^+ = \sum_{j=0}^{\infty} \hat{\mathbf{A}}^j,
\end{align}
Furthermore, the effective resistance between nodes $u$ and $v$ can be written as:
\begin{align}
R(u,v) = \sum_{i=0}^{\infty} \bigg( \frac{1}{d_u} (\hat{\mathbf{A}}^i)_{uu} + \frac{1}{d_v} (\hat{\mathbf{A}}^i)_{vv} - \frac{2}{\sqrt{d_u d_v}} (\hat{\mathbf{A}}^i)_{uv} \bigg),
\end{align}
where $(\hat{\mathbf{A}}^i)_{u,v}$ represents the number of paths of length $i$ between nodes $u$ and $v$ \citep{black2023gtr}. This equation intuitively shows that shorter and disjoint paths connecting two nodes lead to lower effective resistance.

\subsection{Effective Resistance with Fractal Nodes}
\begin{lemma}[Fractal Node Effective Resistance]
Let $\mathcal{G}_f$ be a connected graph with $\mathcal{C}$ subgraphs and their associated fractal nodes. The effective resistance between any two nodes $u$, $v$ with fractal nodes can be expressed as:
\begin{equation}
    R_f(u,v) = (1_u - 1_v)^T \mathbf{L}_f^+ (1_u - 1_v),\nonumber
\end{equation}
where $\mathbf{L}_f$ is the augmented Laplacian incorporating fractal node connections:
\begin{equation}
\mathbf{L}_f = \begin{bmatrix}
\mathbf{L} + \sum_{i=1}^{\mathcal{C}} \mathbf{F}_i\mathbf{F}_i^T & -[\mathbf{F}_1, \mathbf{F}_2, ..., \mathbf{F}_{\mathcal{C}}] \\
-[\mathbf{F}_1, \mathbf{F}_2, ..., \mathbf{F}_{\mathcal{C}}]^T & \mathbf{I}_{\mathcal{C}}
\end{bmatrix},\nonumber
\end{equation}
where $\mathbf{L}$ is the original Laplacian matrix, $\mathbf{F}_i$ is the incidence vector for fractal node $i$ indicating its connections to the original nodes. 
\end{lemma}

Similar to the path-based interpretation in \cite{black2023gtr}, we can express $R_f(u,v)$ in terms of paths:
\begin{align}
R_f(u,v) = \sum_{i=0}^{\infty} \bigg( \frac{1}{d_u} (\hat{\mathbf{A}}_f^i)_{uu} + \frac{1}{d_v} (\hat{\mathbf{A}}_{f}^i)_{vv} - \frac{2}{\sqrt{d_u d_v}} (\hat{\mathbf{A}}_f^i)_{uv} \bigg)
\end{align}
where $\hat{\mathbf{A}}_f$ is the normalized adjacency matrix including fractal node connections.

\subsection{Proof of \Cref{thm:reduction}}\label{app:proof-reduction}
\begin{thmreduction}[Resistance reduction]
Let $\mathcal{G}$ be the original graph and $\mathcal{G}_f$ be the augmented graph with fractal nodes. For any nodes $u,v \in \mathcal{G}$, the effective resistance in $\mathcal{G}_f$ satisfies:
\begin{equation}
   R_f(u,v) \leq R(u,v),\nonumber
\end{equation}
where $R_f(u,v)$ is the effective resistance in $\mathcal{G}_f$ and $R(u,v)$ is the original effective resistance in $\mathcal{G}$.
\end{thmreduction}

\begin{proof}
Let $\mathcal{G}=(\mathcal{V},\mathcal{E})$ be the original graph and $\mathcal{G}_f=(\mathcal{V} \cup \mathcal{F}, \mathcal{E} \cup \mathcal{E}_f)$ be the augmented graph with fractal nodes, where $\mathcal{F}$ is the set of fractal nodes and $\mathcal{E}_f$ is the set of edges connecting nodes to fractal nodes.

Following~\citet{black2023gtr}, we express the effective resistance in terms of path decomposition:
\begin{align}
R_f(u,v) = \sum_{i=0}^{\infty} \bigg( \frac{1}{d_u} (\hat{\mathbf{A}}_f^i)_{uu} + \frac{1}{d_v} (\hat{\mathbf{A}}_{f}^i)_{vv} - \frac{2}{\sqrt{d_u d_v}} (\hat{\mathbf{A}}_f^i)_{uv} \bigg),
\end{align}
where $\hat{\mathbf{A}}_f$ is the normalized adjacency matrix of $\mathcal{G}_f$.

Let $\mathcal{P}_{uv}$ be the set of all paths connecting $u$ and $v$ in $\mathcal{G}_f$. The effective resistance can be expressed as:
\begin{equation}
R_f(u,v) = \min_{p \in \mathcal{P}_{uv}} \sum_{(x,y) \in p} r_{xy},
\end{equation}
where $r_{xy}$ is the resistance of edge $(x,y)$.

By Rayleigh's monotonicity principle~\citep{black2023gtr}, since $\mathcal{G}_f$ contains all edges of $\mathcal{G}$ plus additional edges through fractal nodes, adding these edges can only decrease the effective resistance between any pair of nodes. Therefore:
\begin{equation}
R_f(u,v) \leq R(u,v).
\end{equation}
\end{proof}

\subsection{Proof of \Cref{thm:signal}}\label{app:proof-signal}
\begin{thmsignal}[Signal propagation with fractal nodes]
For a MPNN with fractal nodes, the signal propagation between nodes $u,v$ after $\ell$ layers satisfies:
\begin{align}
    \|h_u^{(\ell)} - h_v^{(\ell)}\| \leq \exp(-\ell/R_f(u,v))\|h_u^{(0)} - h_v^{(0)}\|,\nonumber
\end{align}
where $R_f(u,v)$ is the effective resistance in the augmented graph with fractal nodes.
\end{thmsignal}
\begin{proof}
First, the message passing process in MPNN (i.e., GCN) with fractal nodes can be expressed as:
\begin{equation}
h_v^{(\ell+1)} = \sigma\bigg(W h_v^{(\ell)} + \sum_{u \in \mathcal{N}(v)} \frac{1}{\sqrt{d_v d_u}}W h_u^{(\ell)} + W_f h_f^{(\ell)}\bigg),
\end{equation}

where $h_f^{(\ell)}$ is the fractal node representation. To analyze the signal propagation, we consider the continuous-time analog by removing the nonlinearity $\sigma$:
\begin{equation}
\frac{d}{dt}h_v(t) = -\mathbf{L}_f h_v(t),
\end{equation}

The solution to this differential equation is:
\begin{equation}
h_v(t) = \exp(-t\mathbf{L}_f)h_v(0),
\end{equation}

The signal difference between two nodes $u,v$ is bounded as follows:
\begin{align}
||h_u(t) - h_v(t)|| &= \|(\exp(-t\mathbf{L}_f))(h_u(0) - h_v(0))\| \\
&\leq ||exp(-t\mathbf{L}_f)|| \cdot ||h_u(0) - h_v(0)|| \\
&\leq \exp(-t/R_f(u,v))||h_u(0) - h_v(0)||
\end{align}

The last inequality comes from the spectral bound related to the effective resistance $R_f(u,v)$ in the graph augmented with fractal nodes.
Mapping back to the discrete layer steps by setting $t = \ell$, we obtain our desired bound:
\begin{equation}
||h_u^{(\ell)} - h_v^{(\ell)}|| \leq \exp(-\ell/R_f(u,v))||h_u^{(0)} - h_v^{(0)}||,
\end{equation}

This provides the worst-case signal propagation bound in the graph with fractal nodes. By the previously proven \Cref{thm:reduction}, we know that $R_f(u,v) \leq R(u,v)$, thus fractal nodes provide better signal propagation guarantees than the original graph.
\end{proof}

\begin{corollary}[Improved signal propagation]\label{cor:signal}
Since $R_f(u,v) \leq R(u,v)$ by the Resistance Reduction theorem, fractal nodes improve the worst-case signal propagation bound compared to the original graph:
\begin{equation}
\exp(-\ell/R_f(u,v)) \leq \exp(-\ell/R(u,v)).\nonumber
\end{equation}
\end{corollary}

\subsection{Total Resistance Analysis}
\begin{theorem}[Total Resistance with Fractal Nodes]
Let $\mathcal{G}_f$ be the graph augmented with $C$ fractal nodes. The total effective resistance satisfies:
\begin{equation}
    R_{tot}^f = n\cdot \text{tr}(\mathbf{L}_f^+) = n\cdot\sum_{i=2}^{n+C} \frac{1}{\sigma_i},\nonumber
\end{equation}
where $\mathbf{L}_f$ is the augmented Laplacian and $\sigma_i$ are its non-zero eigenvalues.
\end{theorem}

\begin{proof}
% The total resistance can be expressed through the trace of the pseudoinverse of the Laplacian matrix $\mathbf{L}_f$. By construction, $\mathbf{L}_f$ has dimension $(n+C) \times (n+C)$ and its eigendecomposition yields $n+C$ eigenvalues. The pseudoinverse $\mathbf{L}_f^+$ has the same eigenvectors as $\mathbf{L}_f$ with reciprocal non-zero eigenvalues, giving us the stated formula. The factor $n$ appears because we sum over all pairs of the $n$ original nodes.
The augmented Laplacian $\mathbf{L}_f$ has dimension $(n+C) \times (n+C)$. The first eigenvalue is 0 (for the connected graph), so the sum starts from $i=2$. The factor $n$ appears because we sum effective resistance over all pairs of the $n$ original nodes only.
\end{proof}

\begin{corollary}[Impact of Fractal Node Count]\label{cor:c}
For a graph $\mathcal{G}$ augmented with $C$ fractal nodes, adding more fractal nodes generally decreases the total resistance. Specifically, the additional $C$ eigenvalues in the spectrum of $\mathbf{L}_f$ contribute additional terms $\frac{1}{\sigma_i}$ to the sum, reducing $R_{tot}^f$. This leads to improved signal propagation bounds: $\|h_u^{(\ell)} - h_v^{(\ell)}\| \leq \exp(-\ell/R_f(u,v))$.
\end{corollary}

\section{Additional Related Work}\label{app:comp}
\paragraph{Subgraphs in graph learning.}
Several works introduce hierarchical clustering and coarsening for learning on graphs~\citep{dong2023megraph}.
\citet{chiang2019clustergcn} use graph clustering to identify well-connected subgraphs on large graphs. HC-GNN~\citep{zhong2023hcgnn} shows competitive performance in node classification on large-scale graphs, using hierarchical community structures for message passing. 
In graph Transformers, several hierarchical models~\citep{zhao2022gnnak,gao2022patchgt,zhu2023hsgt,he2023graphViT} attempt to manage computational complexity, though they still face challenges with scalability as all nodes remain within the computational burden of the Transformer architecture. However, our approach, incorporating fractal nodes to MPNN, can reduce this computational cost while preserving structural information.

\paragraph{Comparison to graph coarsening methods.} 
% Our method differs from existing graph coarsening approaches.
Coarformer~\citep{kuang2022coarformer} tries to use coarsened and original graphs as separate views, where the coarsened graph is input to the Transformer. At the same time, ANS-GT~\citep{cai2021ansgt} feeds a sequence of node representations to the graph Transformer by combining original, global, and coarsened node representations formed via adaptive sampling. 
Our method, on the other hand, incorporates fractal nodes representing subgraph information into the MPNN and enables fractal nodes to exchange messages with the original nodes and exchange information between fractal nodes via MLP-Mixer. 

\paragraph{Comparison to virtual node.}
If we do not split into subgraphs, there will be only one fractal node. This can be compared to a virtual node~\citep{Gilmer2017chemi,hwang2022vn,cai2023connection_vn}, which is known to have the information of a global node. 
While both approaches facilitate global or subgraph-level information exchange, the key difference lies in how they process information. Virtual nodes aggregate global information from the entire graph, whereas fractal nodes operate at a subgraph level. 
A virtual node has its own update and aggregation functions that process messages from all graph nodes. In contrast, regular nodes incorporate both their local neighborhood messages and the virtual node's message. In contrast, our fractal nodes adaptively decompose and process both low and high frequency components of subgraph features. 
This allows fractal nodes to capture richer information at the subgraph level compared to virtual node implementations that typically aggregate global information.

\paragraph{Fractality and self-similarity in networks.}
The concept of fractals, introduced by \citet{mandelbrot1983fractal}, describes systems that exhibit self-similarity, where similar patterns recur across different scales.
This concept has been widely applied in network science, where many real-world networks have been shown to exhibit fractal structures and scale-free properties~\citep{song2005self,kim2007fractality,fronczak2024scaling}. For instance, social networks, web networks, and even protein interaction networks have been found to have fractal properties~\citep{chen2020fractalsocial}. 
A common approach to analyzing fractality in networks involves coarsening techniques such as renormalization or box-covering techniques~\citep{kim2007fractality}, which group nodes into coarser units and study the coarsened structures. However, these methods are not directly applicable in graph learning settings, where node coordinates or geometric information are often unavailable.
These observations suggest the potential to incorporate fractality into graph learning through scalable and learnable mechanisms. Instead of relying on explicit structural coarsening methods, box-covering, and renormalization, we explore how self-similar patterns can be captured and utilized directly within the message passing process.

\section{Implementation Detail}\label{app:fn_detail}
\subsection{Metis Partitioning for Fractal Node Creation}\label{app:metis}
To create fractal nodes, we employ METIS \citep{karypis1998metis}, a graph clustering algorithm known for its excellent balance between accuracy and computational efficiency. METIS partitions a graph into a pre-defined number of clusters, maximizing within-cluster connections while minimizing between-cluster links. This approach effectively captures the community structure of the graph.

However, using non-overlapping partitions could result in the loss of important edge information, particularly at the boundaries between partitions. To address this issue and retain all original edges, we introduce overlapping subgraph. After the initial METIS partitioning, we expand each partition to include nodes from neighboring partitions.

Formally, we first apply METIS to partition a graph $\mathcal{G}$ into $C$ non-overlapping subgraphs: $\{ \mathcal{V}_1, \ldots, \mathcal{V}_C \}$ such that $\mathcal{V} = \{\mathcal{V}_1 \cup \ldots \cup \mathcal{V}_C \} $ and $\mathcal{V}_i \cap \mathcal{V}_j = \varnothing , \forall i \neq j$, where $C$ is the number of fractal nodes or subgraphs. Then, we expand these subgraphs to include $k$-hop neighborhoods:
\begin{align}
\mathcal{V}_i \leftarrow \mathcal{V}_i \cup \{ \mathcal{N}_k(j) | j \in \mathcal{V}_i \},
\end{align}
where $\mathcal{N}_k(j)$ defines the $k$-hop neighbourhood of node $j$. This expansion ensures that each subgraph retains information about its immediate surroundings.
The choice of $k$ allows us to control the degree of overlap between subgraphs. A larger $k$ value increases the overlap, potentially capturing more global information but at the cost of increased computational complexity.
This overlapping subgraph approach allows our fractal nodes to capture both local structural details and broader subgraph-level information, enhancing the model's ability to learn multi-scale representations of the graph structure.

\subsection{Instance of Our Framework}\label{app:instance}
We describe update equations for how our fractal node is applied to MPNN.

The update equation for GCN + $\mathsf{FN}$ is the following:
\begin{align}
\begin{split}
\widetilde{h}_{v,c}^{(\ell+1)} &= \sigma\Big(h_{v,c}^{(\ell)} + \sum_{u \in N(v)} \frac{1}{\sqrt{d_v d_u}} h_{u,c}^{(\ell)} {W}^{(\ell)}\Big),\\
f_c^{(\ell+1)} &= \mathsf{LPF}(\widetilde{h}_{v,c}^{(\ell+1)}) + \omega^{(\ell)}_c \cdot \mathsf{HPF}(\widetilde{h}_{v,c}^{(\ell+1)}),\\
h_{v,c}^{(\ell+1)} &= \widetilde{h}_{v,c}^{(\ell+1)} + f_c^{(\ell+1)},
\end{split}\label{eq:gcn-fn}
\end{align}
where $\sigma$ a ReLU activation function, and $d_v$ and $d_u$ are their node degrees. 

The update equation for GatedGCN + $\mathsf{FN}$ is the following:
\begin{align}
\begin{split}
\widetilde{h}_{v,c}^{(\ell+1)} &= \sigma\Big(\Omega^{(\ell)}h_{v,c}^{(\ell)} \\&+ \sum_{u \in N(v)} \mathsf{gate}^{(\ell)}(h_{v,c}^{(\ell)}, h_{u,c}^{(\ell)}) \odot h_{u,c}^{(\ell)} W_1^{(\ell)}\Big),\\
f_c^{(\ell+1)} &= \mathsf{LPF}(\widetilde{h}_{v,c}^{(\ell+1)}) + \omega^{(\ell)} \cdot \mathsf{HPF}(\widetilde{h}_{v,c}^{(\ell+1)}),\\
h_{v,c}^{(\ell+1)} &= \widetilde{h}_{v,c}^{(\ell+1)} + f_c^{(\ell+1)},\\
\mathsf{gate}^{(\ell)}(h_{v,c}^{(\ell)}, h_{u,c}^{(\ell)}) &= \mathsf{sigmoid}({W}2^{(\ell)} h_{v,c}^{(\ell)} + W_3^{(\ell)}) h_{u,c}^{(\ell)},
\end{split}
\end{align}
where $\sigma$ is a ReLU activation function, $W_0^{(\ell)}$, $W_1^{(\ell)}$, $W_2^{(\ell)}$, $W_3^{(\ell)}$ are learnable weight matrices, $\mathsf{gate}^{(\ell)}$ is a gating mechanism that controls the information flow between nodes.

The update equation for GINE + $\mathsf{FN}$ is the following:
\begin{align}
\begin{split}
\widetilde{h}_{v,c}^{(\ell+1)} &= \mathsf{MLP}^{(\ell)}\Big((1 + \epsilon^{(\ell)}) \cdot h_{v,c}^{(\ell)} + \sum_{u \in N(v)} \sigma(h_{u,c}^{(\ell)} + e_{uv}^{(\ell)})\Big),\\
f_c^{(\ell+1)} &= \mathsf{LPF}(\widetilde{h}_{v,c}^{(\ell+1)}) + \omega^{(\ell)}_c \cdot \mathsf{HPF}(\widetilde{h}{v,c}^{(\ell+1)}),\\
h_{v,c}^{(\ell+1)} &= \widetilde{h}_{v,c}^{(\ell+1)} + f_c^{(\ell+1)},
\end{split}\label{eq:gine-fn}
\end{align}
where $\epsilon^{(\ell)}$ is a learnable scalar parameter, and $e_{uv}^{(\ell)}$ is a edge hidden vector between node $u$ and $v$.

Note that the positional encoding scheme and readout function schemes can also be applied to MPNNs with fractal nodes. 

\subsection{Positional Encoding}\label{app:pe}
When we integrate our fractal node to MPNN, we incorporate two distinct positional encodings (PE): an absolute PE for individual nodes and a relative PE for fractal nodes.

For node-level encoding, we consider dataset-specific approaches. We utilize random-walk structural encoding (RWSE) for molecular graphs and Laplacian eigenvector encodings for super-pixel image-based tasks. To enhance robustness, we randomly flip the sign of Laplacian eigenvectors during training.

Let $M \in \{0,1\}^{C \times |\mathcal{V}|}$ be a binary matrix where each row corresponds to a fractal node and each column to an original graph node. $M_{ij} = 1$ if node $j$ belongs to fractal node $i$, and 0 otherwise. Then, the coarsened adjacency matrix is computed as $A^C = M M^\top$.
This operation effectively counts the number of connections between fractal nodes, where $A^C_{ij}$ represents the number of edges between fractal nodes $i$ and $j$ in the original graph.
We then derive a positional encoding $p_v \in \mathbb{R}^{d_p}$ for each fractal node from this coarsened adjacency matrix. This encoding is incorporated into the fractal node representation through a linear transformation:
\begin{align}
f_v^{(L)} = Tp_v + O f_v^{(L)} + b \in \mathbb{R}^d,
\end{align}
where $T \in \mathbb{R}^{d \times d_p}$ and $O \in \mathbb{R}^{d \times d}$ are learnable transformation matrices, and $b \in \mathbb{R}^d$ is a learnable bias vector.

By incorporating relative positional information between fractal nodes, we enable the $\mathsf{FN}_M$ variant to better use the hierarchical structure of the graph.

\section{Fractal Structure and Node Centrality}\label{app:similarity}
In this section, we describe how we calculate the fractal structure of a network by comparing the node centrality distributions between the original graph and its subgraphs using betweenness centrality. Specifically, we use the Kolmogorov-Smirnov (KS) test to measure the similarity between these distributions.

\begin{figure*}[t!]
    \centering
    \subfigure[$k=0$]{\includegraphics[width=0.3\linewidth]{img/cent/cent_hop0.pdf}}
    \subfigure[$k=1$]{\includegraphics[width=0.3\linewidth]{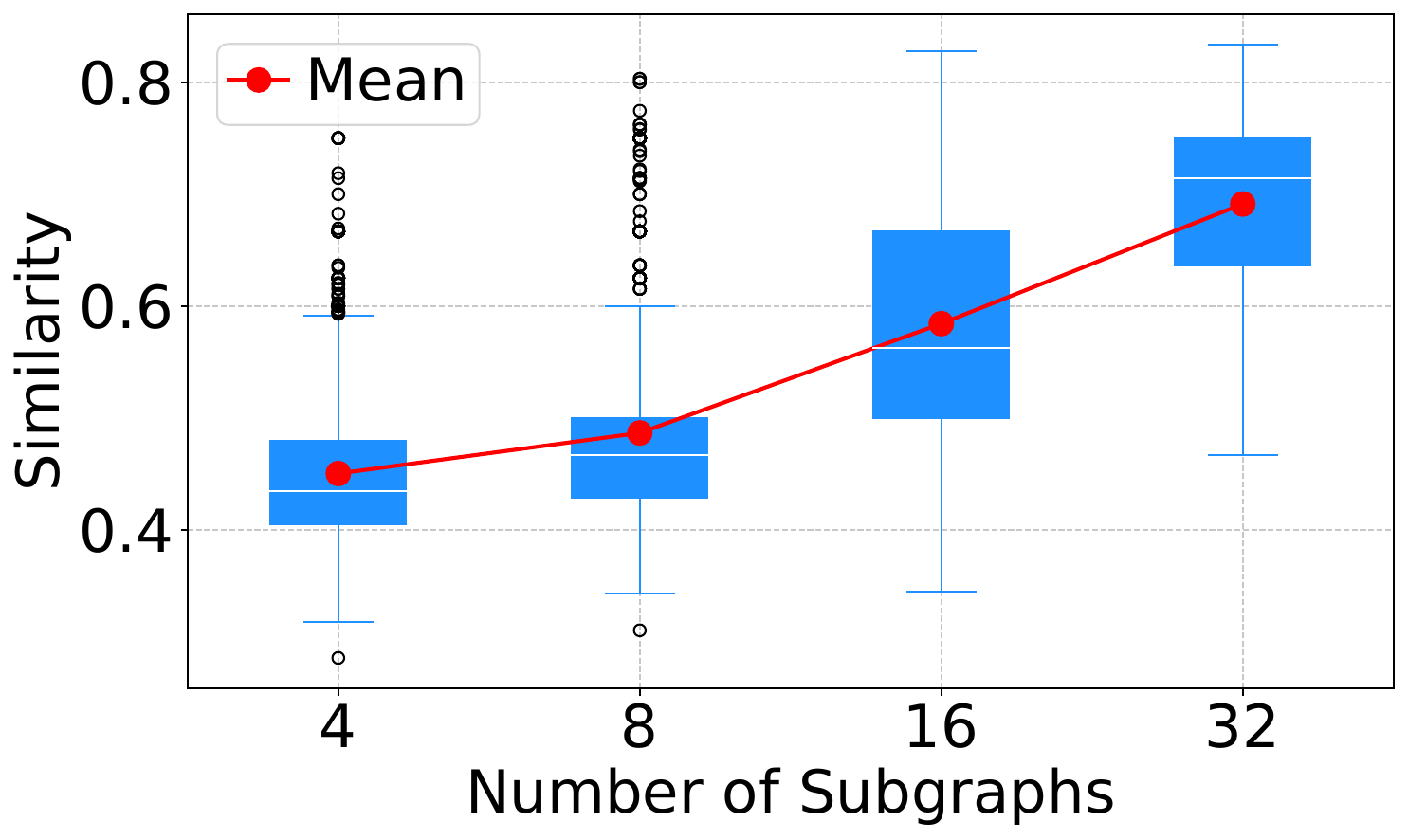}}
    \subfigure[$k=2$]{\includegraphics[width=0.3\linewidth]{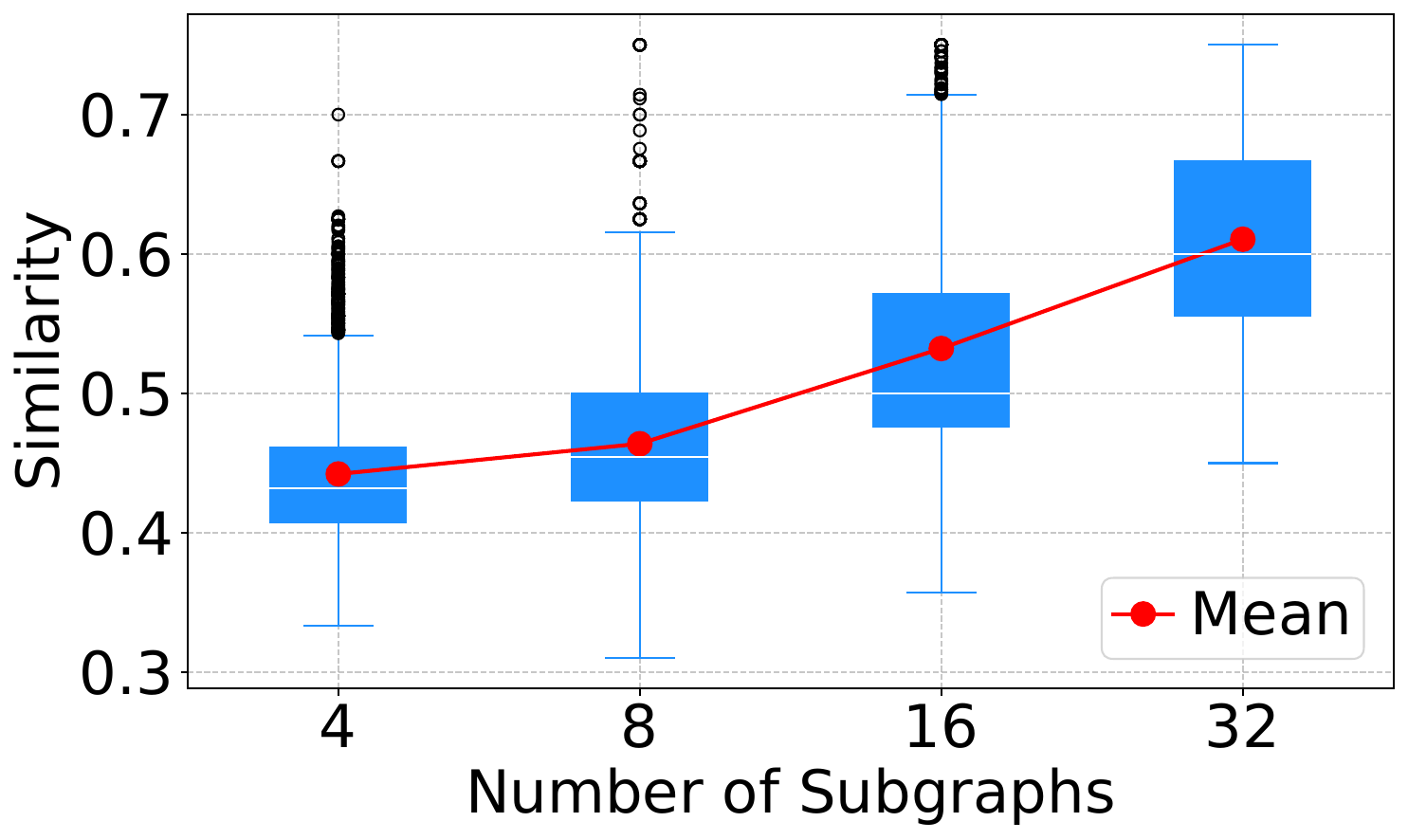}}
    \caption{Similarity of node centrality distribution in \textsc{Peptides-struct}.}
    \label{fig:sim_hops}
\end{figure*}

\paragraph{Fractal structure definition.} We define the fractality of a graph as the degree to which the properties of the subgraphs resemble those of the original graph when the graph is partitioned consistently. In this work, we focus on how the betweenness centrality distribution of the original graph compares to those of its subgraphs.

Let $\Psi(x)$ represent the node centrality distribution function for the original graph, and let $\Psi_0(x), \dots, \Psi_{32}(x)$ represent the centrality distributions for each of the subgraphs obtained by partitioning the original graph into 32 subgraphs. We aim to quantify the similarity between $\Psi(x)$ and the subgraph distributions using the KS test.

\paragraph{Kolmogorov-Smirnov test.} The KS test is a non-parametric test that compares the empirical cumulative distribution function (CDF) $\Psi_n(x)$ of the sample (subgraph centrality) with the CDF $\Psi(x)$ of the reference distribution (original graph centrality). The KS test statistic $D$ is defined as:
\begin{align}
    D = \sup_x|\Psi_n(x)-\Psi(x)|,
\end{align}
where $D$ represents the maximum distance between the two CDFs. A smaller $D$ value indicates higher similarity between the two distributions.

\paragraph{Similarity metric.} We define the similarity between the original graph and a subgraph as $1-D$, where $D$ is the KS test statistic. Therefore, a higher $1-D$ value implies greater similarity. For each graph, we compute the similarity for all $C$ subgraphs, yielding $C$ similarity values.

\paragraph{Fractality calculation.} In our structural fractality evaluation, we identify the subgraph whose centrality distribution is most similar to that of the original graph. This is because not all subgraphs need to exhibit structural similarity for the graph to be considered fractal-like; the presence of one or more highly similar subgraphs is indicative of fractality. Thus, we take the maximum of the $C$ similarity values ($1-D$) as a structural similarity score for the graph:
\begin{align}
    \text{Similarity Score} = \max_i(1-D_i),
\end{align}
where $D_i$ is the KS test statistic for the $i$-th subgraph.
This approach allows us to compute the structural similarity score for a single graph based on betweenness centrality. The comparison according to the number of subgraphs is shown in \Cref{fig:sim_hops}.

\begin{figure*}[h!]
    % \vspace{-1em}
    \centering
    \subfigure[GCN]{\includegraphics[width=0.33\textwidth]{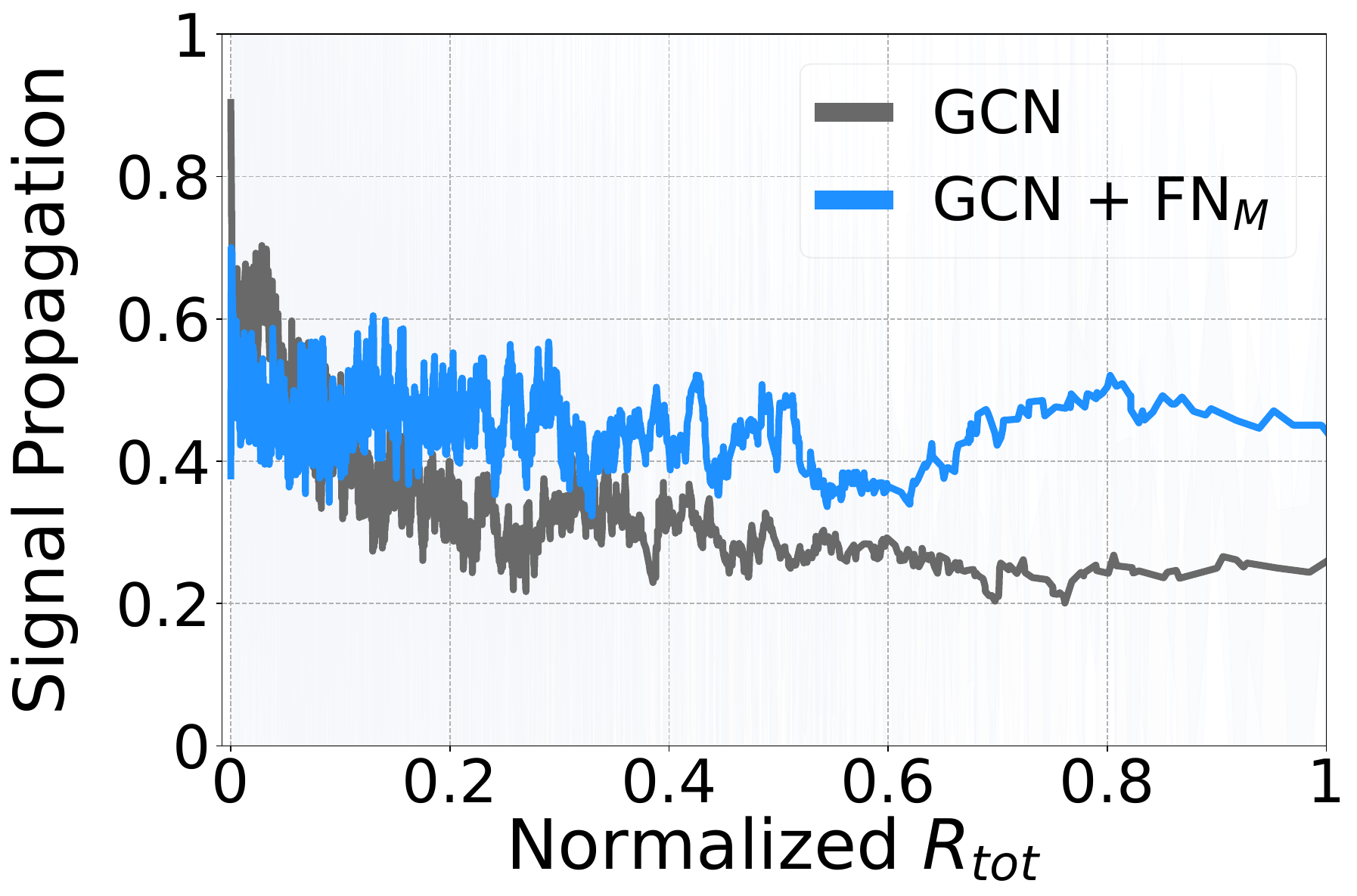}}
    \subfigure[GINE]{\includegraphics[width=0.33\textwidth]{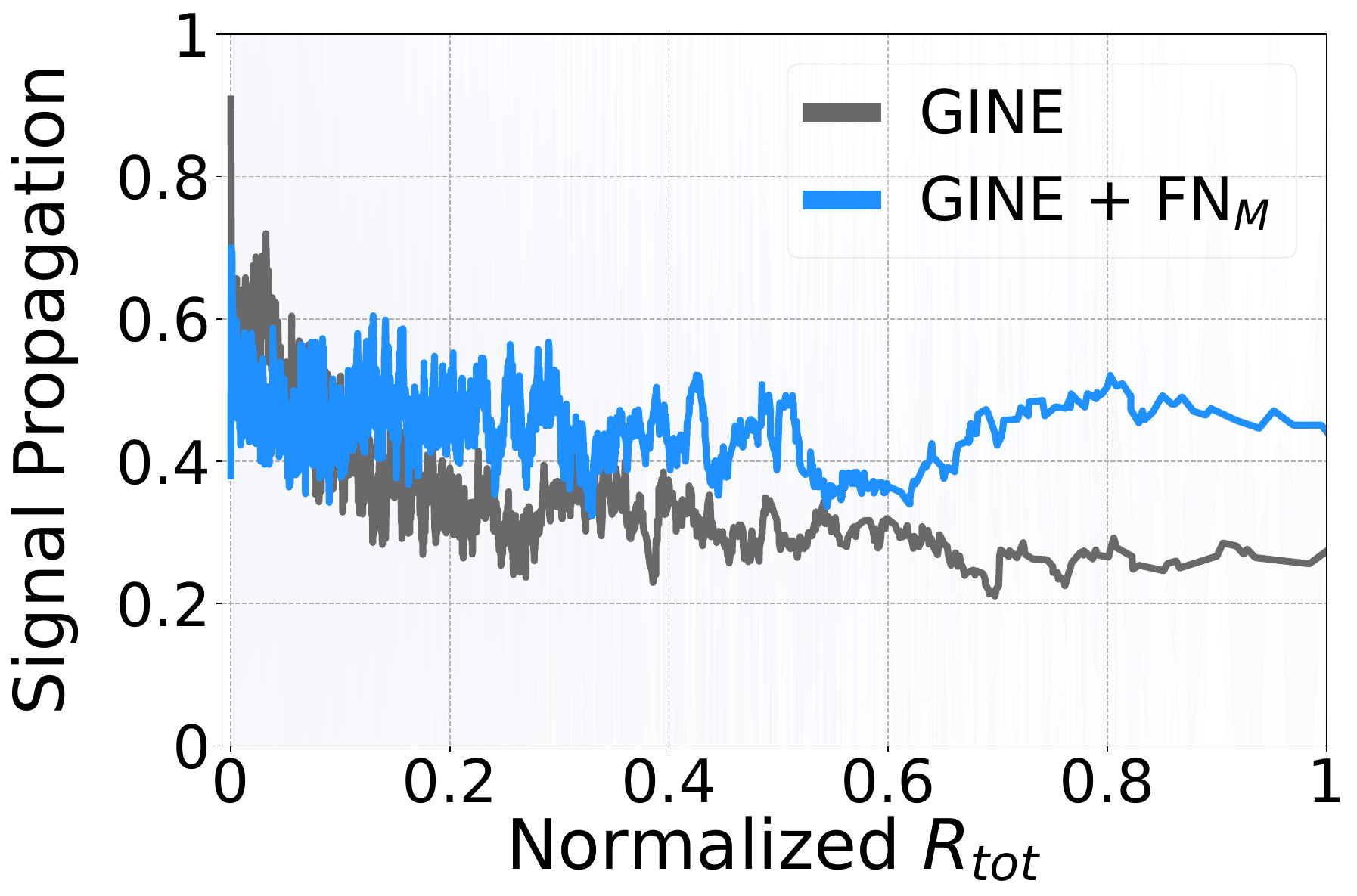}}
    \subfigure[GatedGCN]{\includegraphics[width=0.33\textwidth]{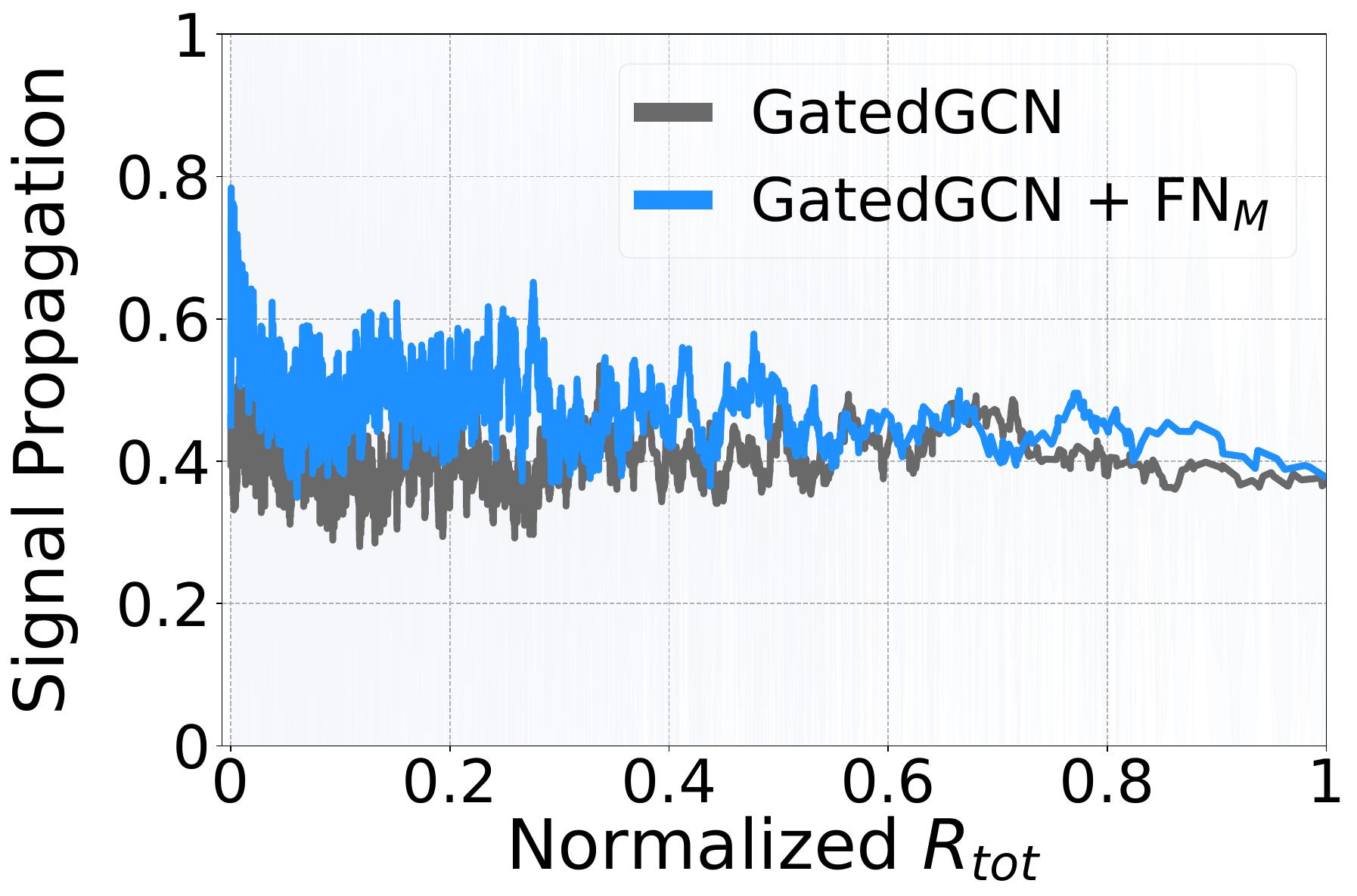}}
    \caption{The amount of signal propagated across the graphs w.r.t. the normalized $R_{tot}$ in \textsc{Peptides-struct}.}
    \label{fig:signal-peptide_struct}
\end{figure*}

\begin{figure*}[h!]
    % \vspace{-1em}
    \centering
    \subfigure[GCN]{\includegraphics[width=0.33\textwidth]{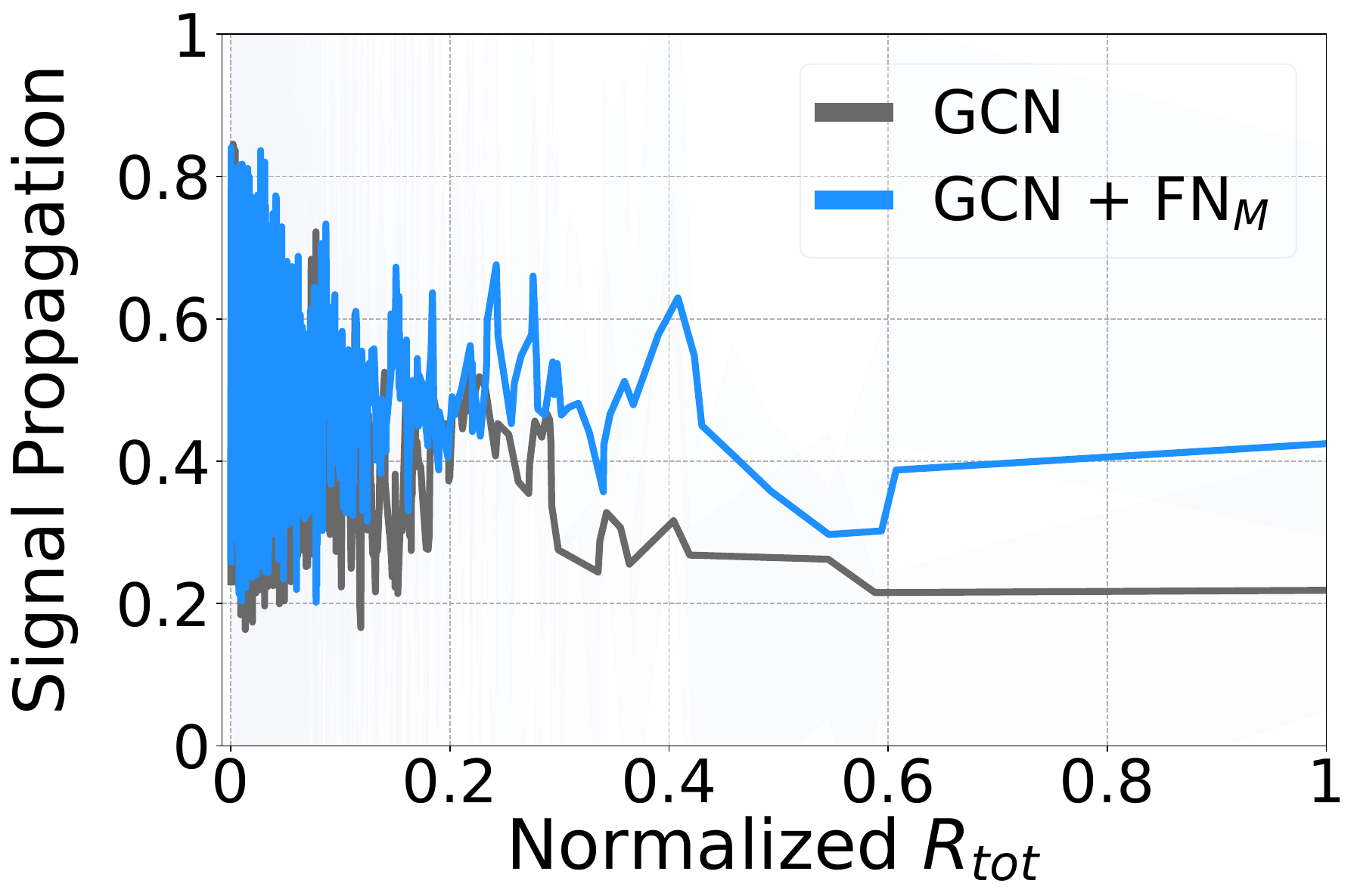}}
    \subfigure[GINE]{\includegraphics[width=0.33\textwidth]{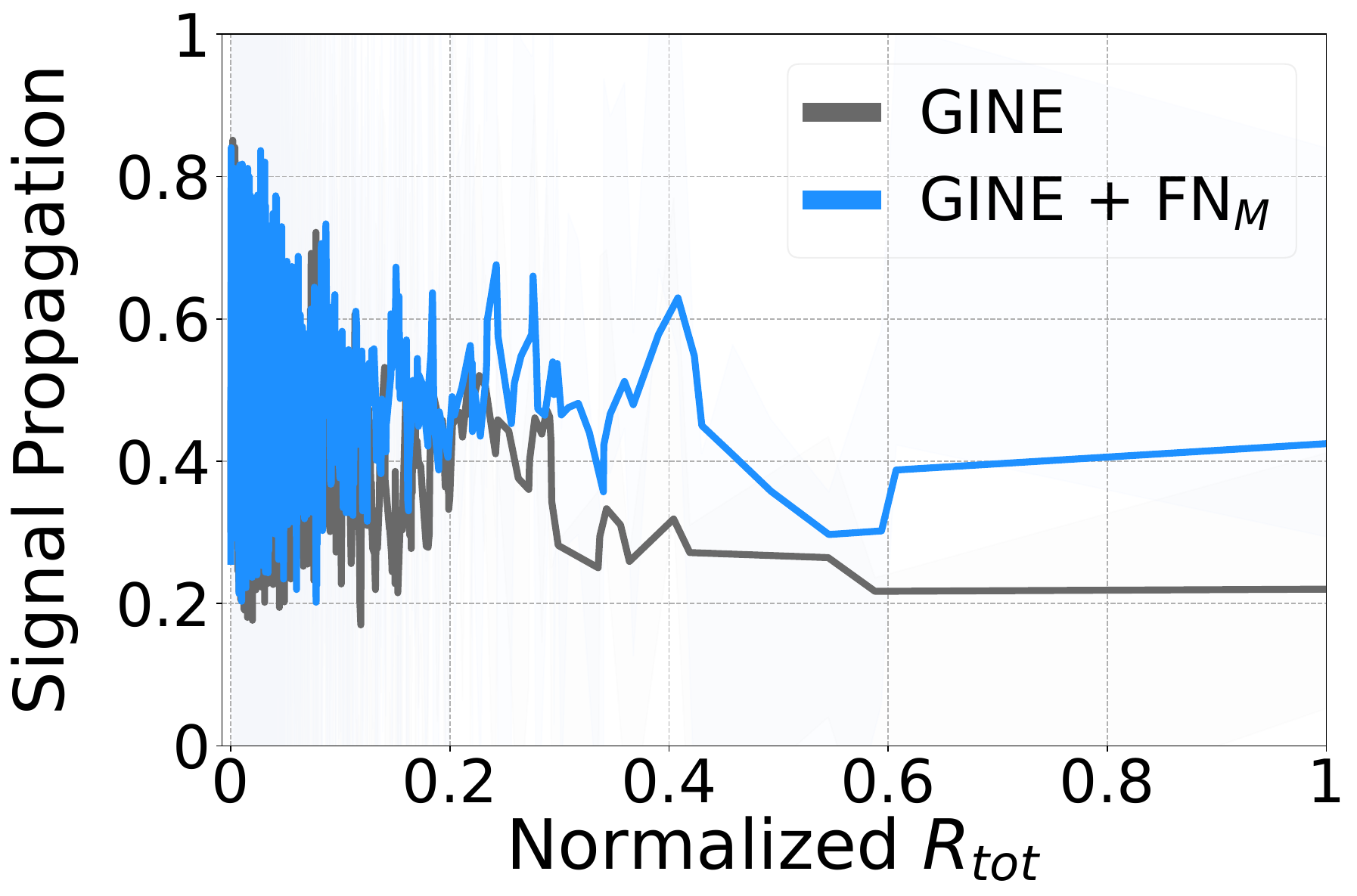}}
    \subfigure[GatedGCN]{\includegraphics[width=0.33\textwidth]{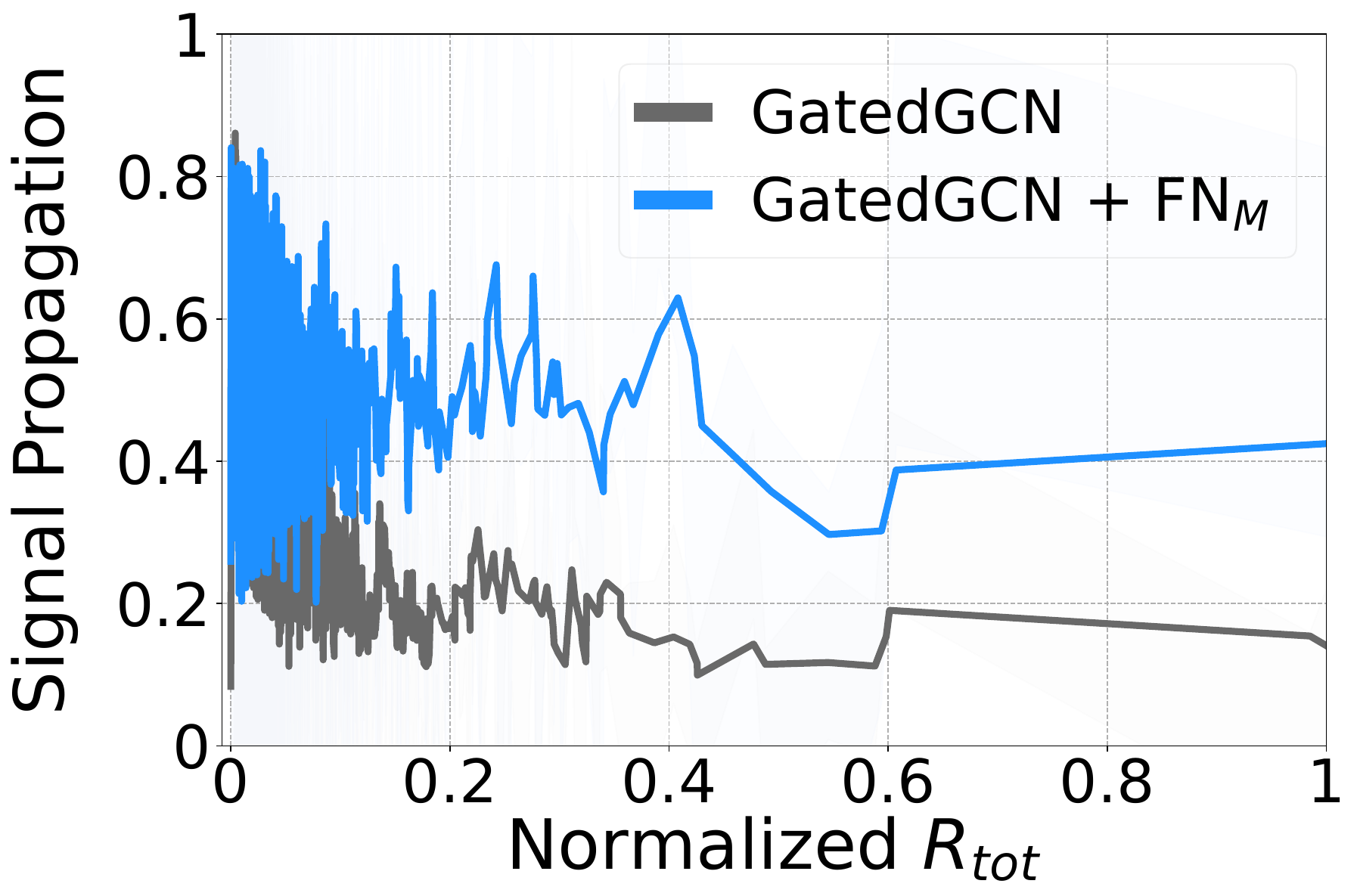}}
    \caption{The amount of signal propagated across the graphs w.r.t. the normalized $R_{tot}$ in \textsc{Molhiv}.}
    \label{fig:signal-molhiv}
\end{figure*}

\begin{figure*}[h!]
    % \vspace{-1em}
    \centering
    \subfigure[GCN]{\includegraphics[width=0.33\textwidth]{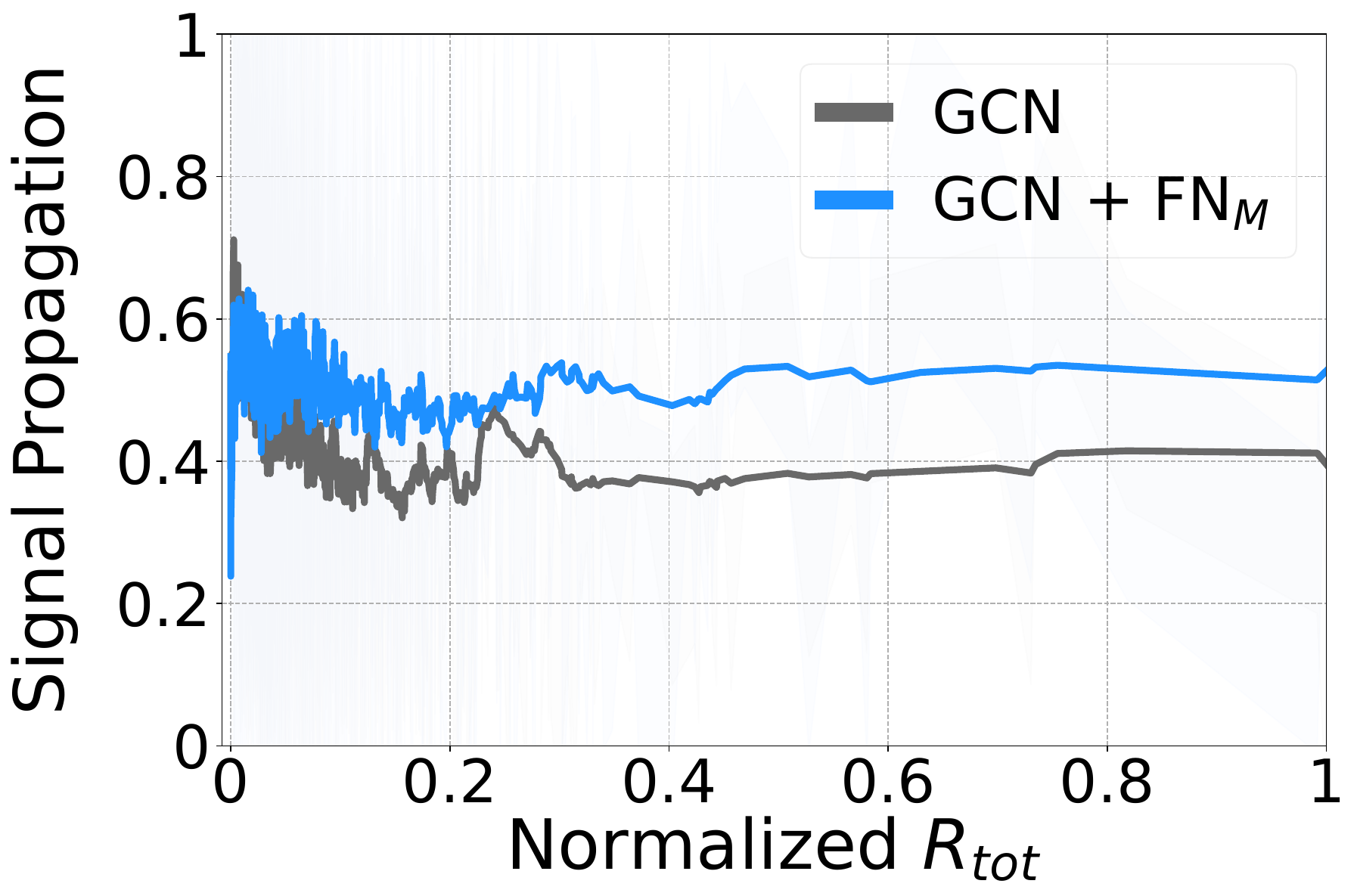}}
    \subfigure[GINE]{\includegraphics[width=0.33\textwidth]{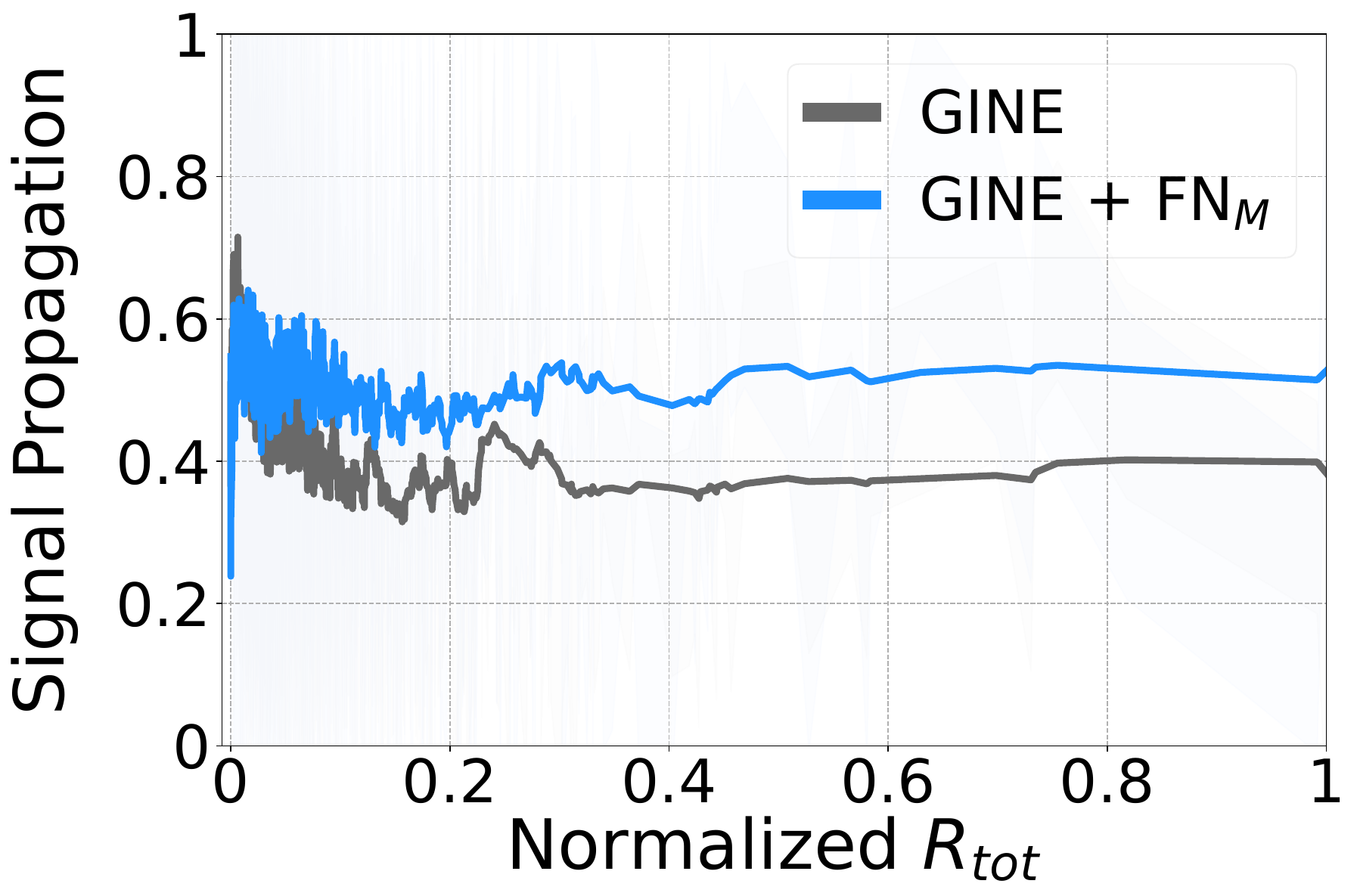}}
    \subfigure[GatedGCN]{\includegraphics[width=0.33\textwidth]{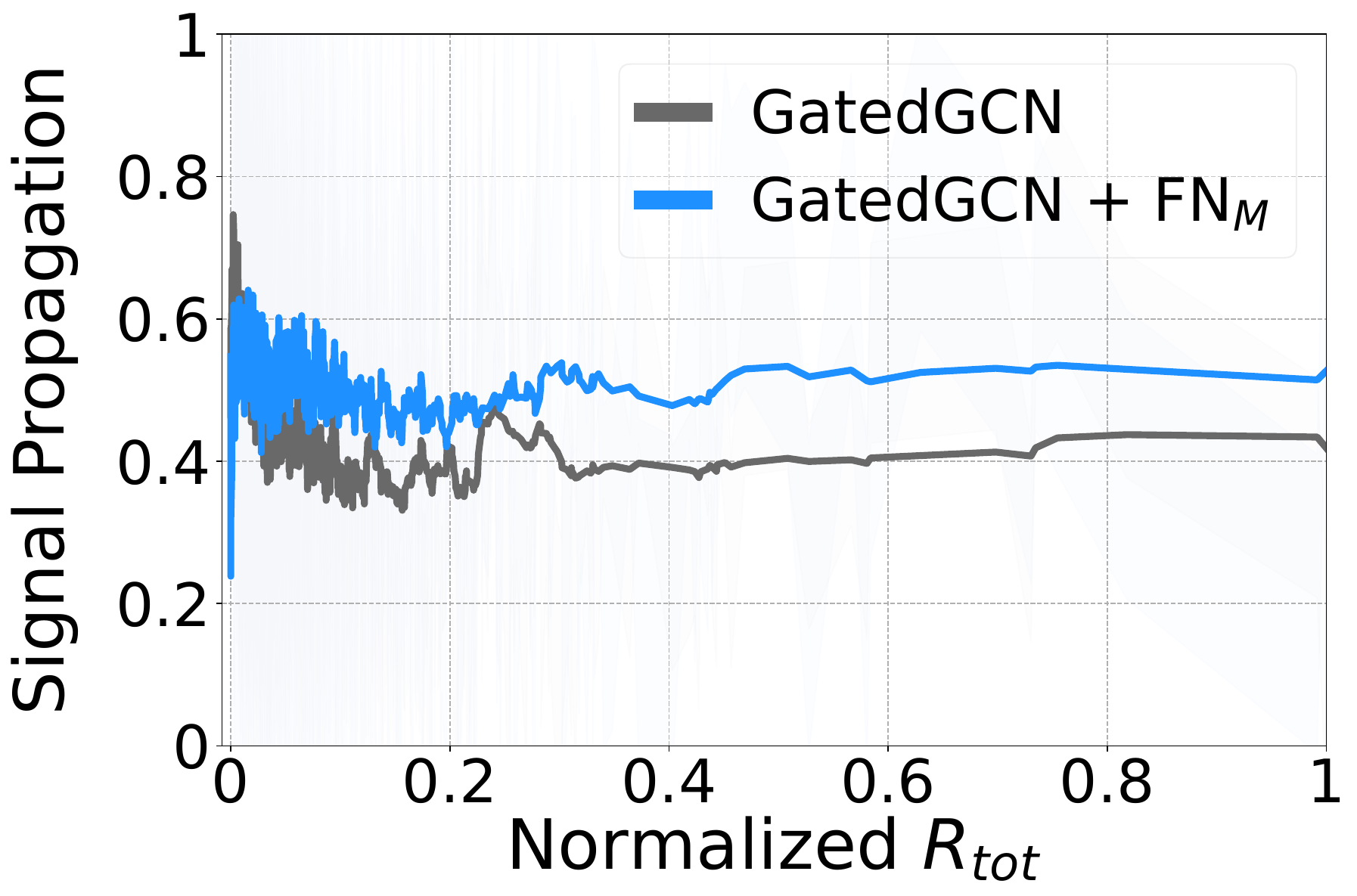}}
    \caption{The amount of signal propagated across the graphs w.r.t. the normalized $R_{tot}$ in \textsc{Moltox21}.}
    \label{fig:signal-moltox}
\end{figure*}
\section{Details on Signal Propagation Analyses}\label{app:signal_exp}
Here, we outline the experimental details for measuring signal propagation with respect to the normalized total effective resistance of the graphs. First, we randomly select a source node and assign a $p$-dimensional feature vector to it, while all other nodes are initialized with zero vectors.
Then, the amount of signal that has been propagated over the graph by the randomly initialized model with $\ell$ layers is given by
\begin{align}
    h_{\odot}^{(\ell)} = \frac{1}{p \max_{u\neq v} k_{\mathcal{G}}(u, v)}  \sum_{t=1}^{p} \sum_{u\neq v} \frac{h_{u}^{(\ell),t}}{\|h_{u}^{(\ell),t}\|} k_{\mathcal{G}}(u, v),
\end{align} 
where $h_{u}^{(\ell), t}$ is the $t$-th feature of $p$-dimensional feature vector of node $u$ at layer $\ell$ and $k_{\mathcal{G}}(u, v)$ is the distance between two nodes $u$ and $v$, computed as a shortest path.
Every unitary signal $h_{u}^{(\ell), t} / \| h_{u}^{(\ell),t}\|$ propagated across the graph $G$ from the source node $v$ is weighted by the normalized propagation distance $k_{\mathcal{G}}(u, v) / \max_{u\neq v}k_{G}(u, v)$ for all nodes $u \neq v$ and then averaged over entire output channels.
To estimate the total effective resistance of the graph, 10 nodes are randomly sampled from each graph, and the total effective resistance of the graph is estimated for each source node. The final $h_{\odot}^{(\ell)}$ and total resistance of the graph are obtained by averaging across the 10 sampled nodes. The experiment is repeated for every graph in the dataset, and the signal propagation measured for each graph is plotted against the normalized total effective resistance of the corresponding graph.

In \Cref{,fig:signal-peptide_struct,fig:signal-molhiv,fig:signal-moltox}, we report the results of this analysis.

\section{Detailed Discussion on \Cref{sec:analysis-ex}}\label{app:exp_power}
% \subsection{Expressive Power of Fractal Nodes (\textbf{Q2.})}\label{sec:analysis-ex}
To thoroughly analyze the role of positional encodings (PEs) and fractal nodes in model expressivity, we conducted extensive ablation studies analyzing different combinations of structural components. \Cref{tab:express-full} shows results across three synthetic datasets (CSL, SR25, EXP) designed to test model expressiveness.

Our ablation study reveals several important insights about the interplay between positional encodings and our method. Without PEs, base MPNNs (GCN, GINE, GatedGCN) consistently show limited expressiveness across all datasets, achieving only 10.00\% on CSL, 6.67\% on SR25, and approximately 51-52\% on EXP. Adding PEs substantially improves base model performance, as evidenced by GCN's significant improvement from 10.00\% to 76.17\% on CSL and from 52.17\% to 100\% on EXP.

Particularly, even without any positional encodings, our fractal node variants demonstrate significantly enhanced expressivity. GINE+$\mathsf{FN}_M$ achieves 47.33\% on CSL and 95.58\% on EXP without any PE, while GatedGCN+$\mathsf{FN}_M$ reaches 49.67\% on CSL. All $\mathsf{FN}_M$ variants achieve 100\% on SR25 regardless of PE configuration, and this indicates that our method provides inherent structural awareness independent of positional encodings.

\begin{table}[h]
    \small
    \setlength{\tabcolsep}{1.2pt}
    \centering
    \caption{Synthetic results (Accuracy $\uparrow$). The gray shaded rows are the results without using PE, and are the fairest to compare against.}
    \label{tab:express-full}
    \begin{tabular}{l cc ccc}\toprule
            \multirow{2}{*}{Method} & \multicolumn{2}{c}{Use of PE} & \multicolumn{3}{c}{Dataset}\\
            \cmidrule(lr){2-3}\cmidrule(lr){4-6}
            & Original $\mathcal{G}$ & Coarsened $\mathcal{G}$ & \textsc{CSL} & \textsc{SR25} & \textsc{EXP}\\\midrule
        \rowcolor{gray! 20}
        GCN      & \xmark & N/A & 10.00 & 6.67 & 52.17\\
        GCN      & \cmark & N/A & 76.17 & 100.0 & 100.0\\
        \cmidrule(lr){1-6}
        \rowcolor{gray! 20}
        GINE     & \xmark & N/A & 10.00 & 6.67 & 51.35\\
        GINE     & \cmark & N/A & 100.0 & 100.0 & 100.0\\
        \cmidrule(lr){1-6}
        \rowcolor{gray! 20}
        GatedGCN & \xmark & N/A & 10.00 & 6.67 & 51.25\\
        GatedGCN & \cmark & N/A & 100.0 & 100.0 & 100.0\\
        \midrule
        \rowcolor{gray! 20}
        GCN + $\mathsf{FN}_M$      & \xmark & \xmark & 39.67 & 100.0 &  86.40\\
        GCN + $\mathsf{FN}_M$      & \xmark & \cmark & 76.17 & 100.0 & 100.0 \\
        GCN + $\mathsf{FN}_M$      & \cmark & \xmark & 100.0 & 100.0 &  100.0\\
        GCN + $\mathsf{FN}_M$      & \cmark & \cmark & 100.0 & 100.0 & 100.0 \\
        \cmidrule(lr){1-6}
        \rowcolor{gray! 20}
        GINE + $\mathsf{FN}_M$     & \xmark & \xmark & 47.33 & 100.0 & 95.58\\
        GINE + $\mathsf{FN}_M$     & \xmark & \cmark & 84.83 & 100.0 & 100.0\\
        GINE + $\mathsf{FN}_M$     & \cmark & \xmark & 100.0 & 100.0 & 100.0 \\
        GINE + $\mathsf{FN}_M$     & \cmark & \cmark & 100.0 & 100.0 & 100.0 \\
        \cmidrule(lr){1-6}
        \rowcolor{gray! 20}
        GatedGCN + $\mathsf{FN}_M$ & \xmark & \xmark & 49.67 & 100.0 & 96.50 \\
        GatedGCN + $\mathsf{FN}_M$ & \xmark & \cmark & 81.83 & 100.0 & 100.0\\
        GatedGCN + $\mathsf{FN}_M$ & \cmark & \xmark & 100.0 & 100.0 & 100.0 \\
        GatedGCN + $\mathsf{FN}_M$ & \cmark & \cmark & 100.0 & 100.0 & 100.0\\
        \bottomrule
    \end{tabular}
\end{table}

%%%%%%%%%%%%%%%%%%%%%%%%%%%%%%%%%%%%%%%%%%%%%%%%%%%%%%%%%%%%%%%%%%%%%%%%%%%%%%%

\section{Experimental Details on Graph-level Tasks}\label{app:exp_detail}
In this section, we provide further details about our experiments.

\subsection{Dataset Description}\label{app:dataset}
We provide the descriptions and statistics of all datasets used in our experiments.
\paragraph{\textsc{Peptides-func} \& \textsc{Peptides-struct}.} (\texttt{CC BY-NC 4.0 License})~\citep{dwivedi2022LRGB}: These datasets comprise 16K atomic peptide graphs from SAT-Pdb, with residues as nodes. They differ in graph-level tasks: \textsc{Petides-func} is a multi-label classification task with 10 nonexclusive functional classes, while \textsc{Peptides-struct} involves regression on 11 3D structural properties. Dataset splitting utilizes meta-class holdout based on original peptide labels.

\paragraph{\textsc{MNIST} \& {CIFAR10}.} (CC BY-SA 3.0 and MIT License): These datasets adapt popular image classification tasks to graph classification. Images are converted to graphs using super-pixels, representing homogeneous intensity regions. Both are 10-class classification tasks following standard splits: 55K/5K/10K for MNIST and 45K/5K/10K for CIFAR10 (train/validation/test).

\paragraph{\textsc{Molhiv} \& \textsc{Moltox21}.} (\texttt{MIT License}) \citep{hu2020ogb}: These molecular property prediction datasets use common node and edge features representing chemophysical properties, pre-processed with RDKit~\citep{rdkit}. Molecules are represented as graphs with atoms as nodes and chemical bonds as edges. Node features are 9-dimensional, including atomic number, chirality, and other properties. Predefined scaffold partitions are used: \textsc{Moltox21} 6K/0.78K/0.78K and \textsc{MolHiv} 32K/4K/4K for training/validation/test.

\paragraph{CSL.} \textsc{CSL}~\citep{murphy2019csl} is a synthetic dataset testing GNN expressivity, containing 150 4-regular graphs in 10 isomorphism classes. These graphs, indistinguishable by 1-WL tests, form cycles with skip-links. The task is to classify them into their respective isomorphism classes.

\paragraph{EXP.} \textsc{EXP}~\citep{Ralph2021exp} consists of 600 graph pairs that 1\&2-WL tests fail to distinguish, aiming to classify these into two categories.

\paragraph{SR25.} SR25~\citep{balcilar2021SR25} consists of 15 strongly regular graphs (3-WL indistinguishable) with 25 nodes each, forming a 15-way classification problem.

\paragraph{\textsc{TreeNeighbourMatch}.} Proposed by \citet{alon2021oversquashing}, this synthetic dataset highlights over-squashing in MPNNs. It uses binary trees of depth $r$ (problem radius), requiring information propagation from leaves to a target node for label prediction, thus demonstrating over-squashing issues.

\subsection{Hardware Specifications and Libraries}
We have implemented our method using \textsc{PyTorch-Geometric}, and built on the source code of \citet{rampavsek2022gps}\footnote{\url{https://github.com/rampasek/GraphGPS}} and \citet{he2023graphViT}\footnote{\url{https://github.com/XiaoxinHe/Graph-ViT-MLPMixer}}.
All experiments were performed using the following software and hardware environments: \textsc{Ubuntu} 18.04 LTS, \textsc{Python} 3.7.13, \textsc{PyTorch} 1.12.1, \textsc{PyTorch Geometric} 2.5.2, , \textsc{torch-scatter} 2.1.0, \textsc{torch-sparse} 0.6.16, \textsc{Numpy} 1.24.3, \textsc{metis} 0.2a5, \textsc{CUDA} 11.3, \textsc{NVIDIA} Driver 465.19, i9 CPU, \textsc{NVIDIA RTX 3090/A6000}.

\subsection{Setup \& Hyperparameters}\label{app:best}
We use the same learning rates and weight decay to GCN, GINE, and GatedGCN, and the hyperparameters we considered are shown in \Cref{tab:param-fn,tab:best-param-fn,tab:best-param-fnm}. The experimental results of MPNN are the same as the results using positional encoding, and we use the setup of \citet{he2023graphViT}.

In \Cref{tab:param-fn,tab:best-param-fn,tab:best-param-fnm}, we report the hyperparameters used in our experiments. $L_M$ denotes the number of layers of MLP-Mixer.

% built on the source code of \citet{rampavsek2022gps}\footnote{\url{https://github.com/rampasek/GraphGPS}} and \citet{he2023graphViT}\footnote{\url{https://github.com/XiaoxinHe/Graph-ViT-MLPMixer}}.

\begin{table}[h]
    \small
    \centering
    \setlength{\tabcolsep}{3em}
    \caption{Hyperparameter search space of fractal nodes for benchmark datasets}
    \label{tab:param-fn}
    \begin{tabular}{cc}\toprule
        Hyperparameters & Search Space  \\\midrule
        $\omega_{c}^{(\ell)}$ & \{SC, VC\}  \\
        $C$ & \{4, 8, 16, 32\} \\
        $\mathsf{HPF}$ & \{True, False\} \\
        $k$-hop & \{0, 1, 2\} \\
        $L$ & \{2, 3, 4, 5, 6, 7, 8\} \\
        $L_M$ & \{1, 2, 4\} \\
        \bottomrule
    \end{tabular}
\end{table}

\begin{table*}[h]
    \footnotesize
    \setlength{\tabcolsep}{2pt}
    \centering
    \caption{Best hyperparameter of \textsf{FN} for \textsc{Peptides-func}, \textsc{Peptides-struct}, \textsc{MNIST}, \textsc{CIFAR10}, \textsc{MolHIV}, and \textsc{MolTox21}.}
    \label{tab:best-param-fn}
    \begin{tabular}{ll cccccc}\toprule
        Hyperparameter & Method & \textsc{Peptides-func}   & \textsc{Peptides-struct} & \textsc{MNIST} & \textsc{CIFAR10} & \textsc{MolHIV} & \textsc{MolTox21}\\ \midrule
        \multirow{3}{*}{$\omega_{c}^{(\ell)}$} 
         & GCN      & VC  & SC & VC & VC & VC & SC \\
         & GINE     & SC  & VC & VC & VC & SC & SC  \\
         & GatedGCN & SC  & VC & VC & VC & VC & SC \\
        \midrule
        \multirow{3}{*}{$C$}
         & GCN      & 32 & 32 & 32 & 32 & 32 & 32\\
         & GINE     & 32 & 32 & 32 & 32 & 32 & 32\\
         & GatedGCN & 32 & 32 & 32 & 32 & 32 & 32\\
        \midrule
        \multirow{3}{*}{$\mathsf{HPF}$}
         & GCN      & True & True & True & True & True & True\\
         & GINE     & True & True & True & True & True & True \\
         & GatedGCN & True & True & True & True & True & True\\
        \midrule
        \multirow{3}{*}{$k$-hop}
         & GCN      & 1 & 1 & 1 & 1 & 1 & 1\\
         & GINE     & 1 & 1 & 1 & 1 & 1 & 1\\
         & GatedGCN & 1 & 1 & 1 & 1 & 1 & 1\\
        \midrule
        \multirow{3}{*}{$L$} 
         & GCN      & 4 & 4 & 4 & 7 & 2 & 4\\
         & GINE     & 4 & 4 & 4 & 7 & 2 & 4\\
         & GatedGCN & 4 & 4 & 4 & 7 & 2 & 4\\
        \bottomrule
    \end{tabular}
\end{table*}
\begin{table*}[h]
    \footnotesize
    \setlength{\tabcolsep}{2pt}
    \centering
    \caption{Best hyperparameter of $\mathsf{FN}_M$ for \textsc{Peptides-func}, \textsc{Peptides-struct}, \textsc{MNIST}, \textsc{CIFAR10}, \textsc{MolHIV}, and \textsc{MolTox21}.}
    \label{tab:best-param-fnm}
    \begin{tabular}{ll cccccc}\toprule
        Hyperparameter & Method & \textsc{Peptides-func}   & \textsc{Peptides-struct} & \textsc{MNIST} & \textsc{CIFAR10} & \textsc{MolHIV} & \textsc{MolTox21}\\ \midrule
        \multirow{3}{*}{$\omega_{c}^{(\ell)}$} 
         & GCN      & VC  & SC & VC & VC & VC & VC \\
         & GINE     & SC  & VC & VC & VC & VC & SC  \\
         & GatedGCN & VC  & SC & VC & VC & VC & VC \\
        \midrule
        \multirow{3}{*}{$C$}
         & GCN      & 32 & 32 & 32 & 32 & 32 & 32\\
         & GINE     & 32 & 16 & 32 & 32 & 32 & 32\\
         & GatedGCN & 32 & 32 &  4 &  4 & 32 & 32\\
        \midrule
        \multirow{3}{*}{$\mathsf{HPF}$}
         & GCN      & True & True & True & True & True & True\\
         & GINE     & True & True & True & True & True & True \\
         & GatedGCN & True & True & False & True & True & True\\
        \midrule
        \multirow{3}{*}{$k$-hop}
         & GCN      & 1 & 1 & 1 & 1 & 2 & 1\\
         & GINE     & 1 & 1 & 1 & 1 & 2 & 1\\
         & GatedGCN & 1 & 1 & 1 & 1 & 2 & 1\\
        \midrule
        \multirow{3}{*}{$L$} 
         & GCN      & 4 & 4 & 4 & 7 & 2 & 5\\
         & GINE     & 4 & 4 & 4 & 7 & 2 & 4\\
         & GatedGCN & 4 & 4 & 4 & 8 & 2 & 5\\
        \midrule
        \multirow{3}{*}{$L_{\mathsf{M}}$} 
         & GCN      & 2 & 2 & 4 & 1 & 2 & 4\\
         & GINE     & 2 & 2 & 4 & 1 & 2 & 4\\
         & GatedGCN & 2 & 2 & 4 & 1 & 2 & 4\\
        \bottomrule
    \end{tabular}
\end{table*}

\section{Experimental Details on Large-scale Node Classification}\label{app:ogbn}

\subsection{Implementation for Node Classification}\label{app:imp_node}
While our main experiment focuses on graph-level tasks, our fractal node method can naturally extend to node classification tasks. The key distinction lies in using the processed fractal node representations from the MLP-Mixer layer to make node-level predictions rather than graph-level ones.

For graph-level tasks, as shown in \cref{eq:mix}, the fractal nodes are mixed through the MLP-Mixer to produce
\begin{equation}
    \tilde{F} = \mathsf{MLPMixer}(F^{(L)}), \quad F^{(L)} = [f_1^{(L)}, f_2^{(L)}, ..., f_{C}^{(L)}].
\end{equation}
These mixed representations are then used directly for graph-level prediction via global pooling.

For node classification, however, we need to propagate this mixed global information back to individual nodes. After the MLP-Mixer processes the $C$ fractal nodes according to \cref{eq:mixer1,eq:mixer2}, we obtain $\tilde{F}^{(L)} \in \mathbb{R}^{C\times d}$. These processed fractal node representations need to be aligned with all nodes in their respective subgraphs.

Let $\mathcal{V}_c$ be the set of nodes in subgraph $c$. For each node $v \in \mathcal{V}_c$, we update its final representation by combining its current features with the processed fractal node information from its corresponding subgraph:
\begin{equation}
    h_v^{(\text{final})} = h_v^{(L)} + \tilde{f}_c^{(L)}, \quad \forall v \in \mathcal{V}_c,
\end{equation}
where $\tilde{f}_c^{(L)}$ is the $c$-th row of $\tilde{F}^{(L)}$ corresponding to the fractal node of subgraph $c$. This operation ensures that each node receives the processed global context from its subgraph's fractal node and maintains consistency with our method while adapting it for node-level predictions.

In implementation, this process can be efficiently vectorized using a batch membership index that maps each node to its corresponding fractal node representation. This adaptation allows our fractal node framework to effectively handle both graph-level and node-level tasks while maintaining its computational efficiency and theoretical properties.

\subsection{Dataset Description}
% \paragraph{Large-scale graphs.}
We consider a collection of large-scale graphs released by the Open Graph Benchmark (OGB)~\citep{hu2021ogb}: ogbn-arxiv and ogbn-products with node numbers 0.16M and 2.4M, respectively. We maintain all the OGB standard evaluation settings.
\paragraph{ogbn-arxiv.} (\texttt{ODC-BY License})~\citep{hu2021ogb}: ogbn-arxiv is a citation network among all Computer Science (CS) papers on Arxiv, as noted by \citet{hu2021ogb}. In this network, nodes represent individual Arxiv papers, with edges denoting citation relationships between them. Each paper is linked to a 128-dimensional feature vector, derived by averaging the word embeddings from its title and abstract. These embeddings are created using the WORD2VEC The aim is to predict the subject areas of CS papers on Arxiv, focusing on 40 distinct subject areas. We employ the split method from \citet{hu2021ogb}, training on papers up to 2017, validating with 2018 publications, and testing on papers published from 2019.

\paragraph{ogbn-products.} (\texttt{Amazon License})~\citep{hu2021ogb}: ogbn-products dataset is a large-scale undirected graph representing Amazon’s product co-purchasing network, where nodes correspond to products and edges indicate co-purchase relationships. Each node is associated with a feature vector derived from bag-of-words representations of product descriptions, and the task is to classify products into 47 top-level categories. To better reflect real-world scenarios, the dataset uses a sales-based split: nodes are sorted by popularity, with the top 8\% used for training, the next 2\% for validation, and the rest for testing -- simulating a setting where labels are available only for the most popular products.

\subsection{Setup \& Hyperparameters}
\paragraph{Baselines.}
Our main focus lies on classic MPNNs: GCN~\citep{kipf2017GCN}, and GraphSAGE~\citep{hamilton2017graphSAGE}; the state-of-the-art scalable graph Transformers: GraphGPS~\citep{rampavsek2022gps}, NAGphormer~\citep{chen2023nagphormer}, Exphormer~\citep{shirzad2023exphormer}, NodeFormer~\citep{wu2022nodeformer}, DiffFormer~\citep{wu2023difformer}, PolyNormer~\citep{zakar2023towards}, and SGFormer~\citep{wu2023sgformer}; hierarchical methods: HC-GNN~\citep{zhong2023hcgnn}, ANS-GT~\citep{cai2021ansgt}, and HSGT~\citep{zhu2023hsgt}; MLP-based method: LINKX~\citep{lim2022LINKX}.

\begin{table}[h!]
    \footnotesize
    \setlength{\tabcolsep}{2pt}
    \centering
    \caption{Best hyperparameter of \textsf{FN} for ogbn-arxiv and ogbn-product}
    \label{tab:best-param-ogbn-fn}
    \begin{tabular}{ll cc}\toprule
        Hyperparameter & Method & ogbn-arxiv & ogbn-product\\ \midrule
        \multirow{2}{*}{$\omega_{c}^{(\ell)}$} 
         & GCN      & SC  & SC \\
         & GraphSAGE & SC  & VC  \\
        \midrule
        \multirow{2}{*}{$C$}
         & GCN      & 64 & 32\\
         & GraphSAGE     & 64 & 32\\
        \midrule
        \multirow{2}{*}{$\mathsf{HPF}$}
         & GCN      & True & True \\
         & GraphSAGE     & True & True \\
        \midrule
        \multirow{2}{*}{dim($h$)}
         & GCN      & 512 & 200\\
         & GraphSAGE & 256 & 200\\
        \midrule
        \multirow{2}{*}{$L$} 
         & GCN           & 5 & 4\\
         & GraphSAGE     & 4 & 4\\
        \bottomrule
    \end{tabular}
\end{table}

\paragraph{Setting.}
We conduct hyperparameter tuning on classic MPNNs, which is consistent with the hyperparameter search space of \citet{deng2024polynormer}. Specifically, we use the Adam optimizer with a learning rate from \{0.001, 0.005, 0.01\} and an epoch limit of 2500. We tune the hidden dimension from \{64, 256, 512\}. We consider whether to use batch or layer normalization, residual connections, and dropout rates from \{0.2, 0.3, 0.5, 0.7\}, the number of layers from \{1, 2, 3, 4, 5, 6, 7, 8, 9, 10\}, and $C$ from \{32, 64, 128\}.

\section{Scalability Analysis of of Fractal Node}\label{app:scale}
% \subsection{Profiling Results on Synthetic Graphs}
To evaluate the efficiency and scalability of our $\mathsf{FN}$ integrated GCN model, we conducted experiments on synthetic Erdos-Renyi~\citep{erdos1960evolution} graphs with node counts ranging from 1,000 to 100,000. The edge probability in the Erdos-Renyi network is set to achieve an average node degree of approximately 5, with the node feature dimension fixed at 100.

\begin{figure}[h!]
    \centering
    \subfigure[Memory Usage vs. Graph Size]{\includegraphics[width=0.48\linewidth]{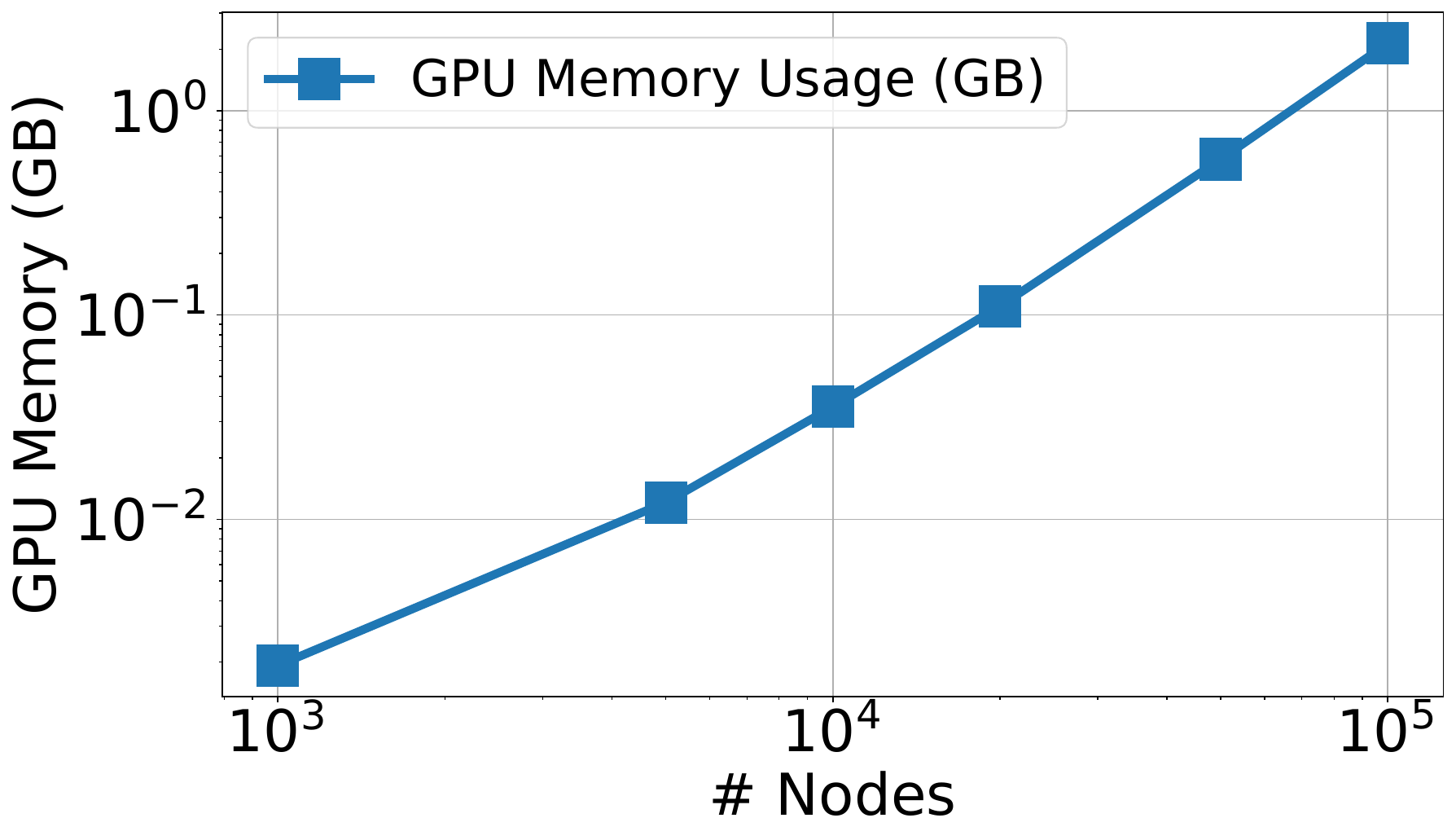}
    \label{fig:gpu-memory}}
    % \hfill
    \subfigure[Training Time vs. Graph Size]{\includegraphics[width=0.48\linewidth]{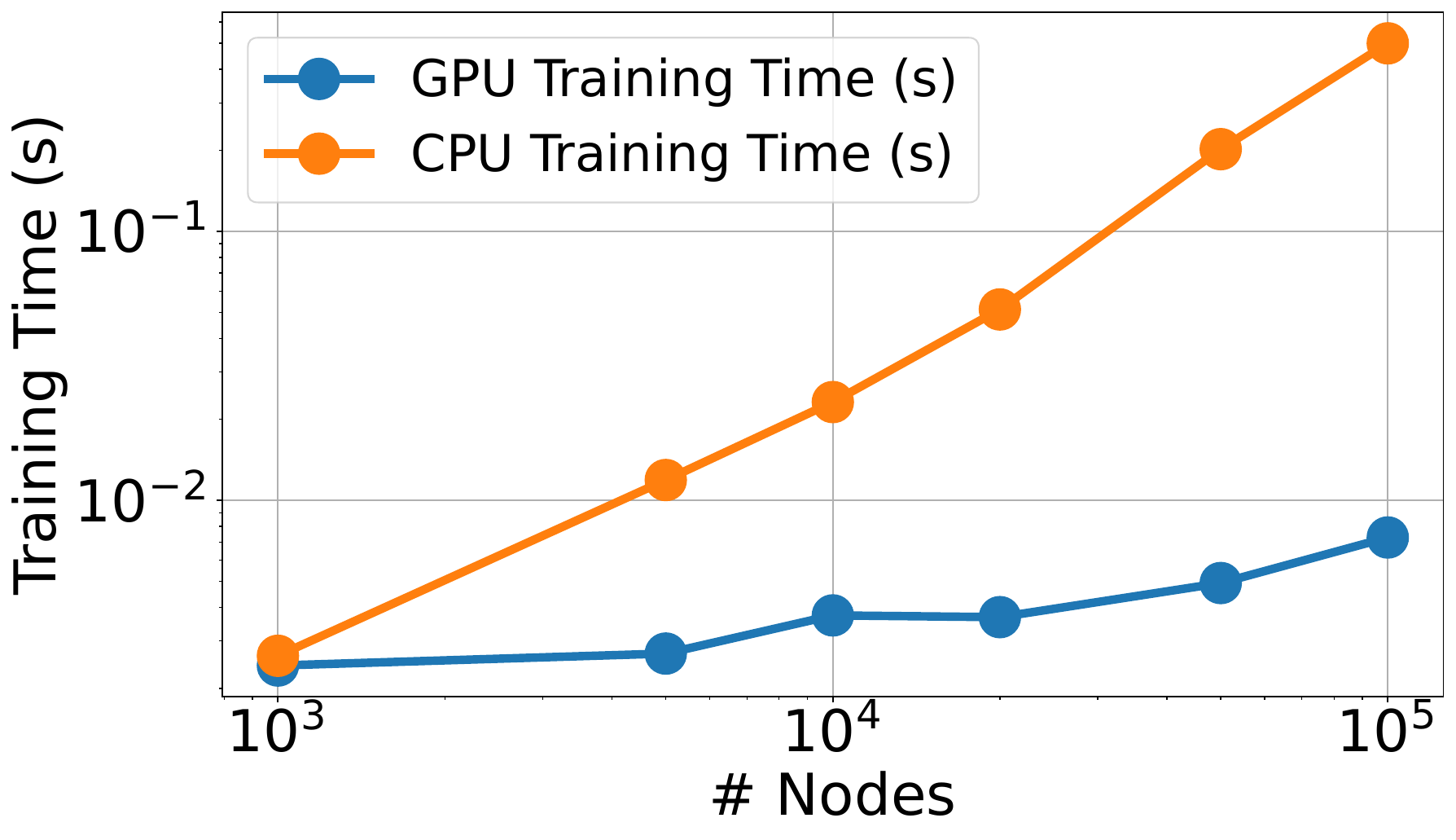}
    \label{fig:gpu-time}}
    \caption{GPU memory usage and training time of GCN+$\mathsf{FN}$ on synthetic graphs.}
    \label{fig:gpu}
\end{figure}

\Cref{fig:gpu-memory} represents that the GPU memory usage of GCN+$\mathsf{FN}$ increases linearly with the graph size and validates its linear space complexity. \Cref{fig:gpu-time} shows the training time for both GPU and CPU implementations. The GPU training time exhibits a sub-linear growth trend as the graph size increases. This means the ability of fractal nodes to effectively use GPU parallelism for large-scale graph computations. In contrast, the CPU training time grows linearly with the graph size and indicates the sequential nature of CPU computations and its limitations in handling large-scale parallel graph operations.

The results demonstrate that the GPU device (RTX A6000 used in our experiments) efficiently handles the computational workload on varying graph sizes. These observations validate the scalability and practicality of our proposed GCN+$\mathsf{FN}$ model, particularly for large-scale graph learning tasks where both memory efficiency and computational speed are critical.

\section{Ablation, Sensitivity and Additional Studies}\label{app:ablation}

\subsection{Impact of $\mathsf{HPF}$}\label{app:hpf}
We use both $\mathsf{LPF}$ and $\mathsf{HPF}$ to create fractal nodes, as shown in \cref{eq:f_c}. We analyze the cases when $\omega^{(\ell)}_c$ is 0, i.e., with and without $\mathsf{HPF}$. Our results are reported in \Cref{tab:hpf}, and we obtain the best performance when using $\mathsf{HPF}$ in almost all cases.

\begin{table*}[h]
    % \footnotesize
    \small
    \setlength{\tabcolsep}{2.5pt}
    \centering
    \caption{Ablation study on $\mathsf{HPF}$}
    \label{tab:hpf}
    \begin{tabular}{l c cccccc}\toprule
        \multirow{2}{*}{Method} & \multirow{2}{*}{$\mathsf{HPF}$} & \textsc{Peptides-func}   & \textsc{Peptides-struct} & \textsc{MNIST} & \textsc{CIFAR10} & \textsc{MolHIV} & \textsc{MolTox21}\\ \cmidrule(lr){3-3}\cmidrule(lr){4-4}\cmidrule(lr){5-5}\cmidrule(lr){6-6}\cmidrule(lr){7-7}\cmidrule(lr){8-8}
        & & AP $\uparrow$ & MAE $\downarrow$ & Accuracy $\uparrow$ & Accuracy $\uparrow$ & ROCAUC $\uparrow$ &  ROCAUC $\uparrow$\\ \midrule
        \multirow{2}{*}{GCN + $\mathsf{FN}$}
        &True& \textbf{0.6802\std{0.0043}} & \textbf{0.2530\std{0.0004}} &  \textbf{0.9393\std{0.0084}} & \textbf{0.6006\std{0.0070}} & \textbf{0.7564\std{0.0059}} & \textbf{0.7670\std{0.0073}} \\ 
        &False& 0.6768\std{0.0016} & 0.2547\std{0.0023} &  0.9383\std{0.0102} & 0.5993\std{0.0081} & 0.7551\std{0.0084} & 0.7608\std{0.0093} \\ 
        \cmidrule(lr){1-8}
        \multirow{2}{*}{GINE + $\mathsf{FN}$}
        &True& \textbf{0.6815\std{0.0059}} & \textbf{0.2515\std{0.0020}} & \textbf{0.9790\std{0.0012}} & \textbf{0.6584\std{0.0069}} & \textbf{0.7882\std{0.0050}} & \textbf{0.7751\std{0.0029}}\\ 
        &False& 0.6749\std{0.0111} & 0.2524\std{0.0021} & 0.9788\std{0.0008} & 0.6584\std{0.0069} & 0.7861\std{0.0054} & 0.7702\std{0.0045}\\ 
        \cmidrule(lr){1-8}
        \multirow{2}{*}{GatedGCN + $\mathsf{FN}$}
        &True& \textbf{0.6778\std{0.0071}} & \textbf{0.2536\std{0.0019}} & \textbf{0.9826\std{0.0012}} & \textbf{0.7125\std{0.0035}} & \textbf{0.7967\std{0.0098}} & \textbf{0.7759\std{0.0054}}\\ 
        &False& 0.6661\std{0.0103} & 0.2609\std{0.0016} & 0.9801\std{0.0015} & 0.7010\std{0.0031} &  0.7908\std{0.0084} & 0.7674\std{0.0024}\\ 
        \midrule
        \multirow{2}{*}{GCN + $\mathsf{FN}_M$}
        &True& \textbf{0.6787\std{0.0048}} & \textbf{0.2464\std{0.0014}} & \textbf{ 0.9455\std{0.0004}} & \textbf{0.6413\std{0.0070}} & \textbf{0.7866\std{0.0034}} & \textbf{0.7882\std{0.0041}}\\
        &False& 0.6778\std{0.0056} & 0.2461\std{0.0022} & 0.9448\std{0.0007} & 0.6130\std{0.0080} & 0.7689\std{0.0124} & 0.7874\std{0.0080}\\
        \cmidrule(lr){1-8}
        \multirow{2}{*}{GINE + $\mathsf{FN}_M$}
        &True& \textbf{0.7018\std{0.0074}} & \textbf{0.2446\std{0.0018}} & \textbf{0.9786\std{0.0004}} & \textbf{0.6672\std{0.0068}} & \textbf{0.8127\std{0.0076}} &  \textbf{0.7926\std{0.0021}}\\ 
        &False& 0.6647\std{0.0052} & 0.2484\std{0.0018} & 0.9744\std{0.0007} & 0.6670\std{0.0056} & 0.7959\std{0.0079} &  0.7895\std{0.0067}\\ 
        \cmidrule(lr){1-8}
        \multirow{2}{*}{GatedGCN + $\mathsf{FN}_M$}
        &True& \textbf{0.6950\std{0.0047}} & \textbf{0.2453\std{0.0014}} & 0.9836\std{0.0010} & \textbf{0.7526\std{0.0033}} & \textbf{0.8097\std{0.0047}} & \textbf{0.7922\std{0.0054}}\\
        &False& 0.6900\std{0.0055} & 0.2477\std{0.0005} & \textbf{0.9848\std{0.0005}} & 0.7501\std{0.0042} & 0.7930\std{0.0057} & 0.7883\std{0.0067}\\
    \bottomrule
    \end{tabular}
\end{table*}

\subsection{Impact of type of $\omega^{(\ell)}_c$}\label{app:omega}
When creating a fractal node, we can use a learnable scalar parameter (denoted as `SC') or a learnable vector parameter (denoted as `VC') to make the contribution of high frequency components. We report the results in \Cref{tab:omega}.

\begin{table*}[h]
    % \footnotesize
    \small
    \setlength{\tabcolsep}{2.5pt}
    \centering
    \caption{Sensitivity study on $\omega_c^{(\ell)}$}
    \label{tab:omega}
    \begin{tabular}{l c cccccc}\toprule
        \multirow{2}{*}{Method} & \multirow{2}{*}{$\omega_c^{(\ell)}$} & \textsc{Peptides-func}   & \textsc{Peptides-struct} & \textsc{MNIST} & \textsc{CIFAR10} & \textsc{MolHIV} & \textsc{MolTox21}\\ \cmidrule(lr){3-3}\cmidrule(lr){4-4}\cmidrule(lr){5-5}\cmidrule(lr){6-6}\cmidrule(lr){7-7}\cmidrule(lr){8-8}
        & & AP $\uparrow$ & MAE $\downarrow$ & Accuracy $\uparrow$ & Accuracy $\uparrow$ & ROCAUC $\uparrow$ &  ROCAUC $\uparrow$\\ \midrule
        \multirow{2}{*}{GCN + $\mathsf{FN}$}
        &SC& 0.6797\std{0.0056} & \textbf{0.2530\std{0.0004}} &  0.9377\std{0.0080} & 0.6003\std{0.0075} & 0.7553\std{0.0061} & \textbf{0.7670\std{0.0073}} \\ 
        &VC& \textbf{0.6802\std{0.0043}} & 0.2535\std{0.0033} &  \textbf{0.9393\std{0.0084}} & \textbf{0.6006\std{0.0070}} & \textbf{0.7564\std{0.0059}} & 0.7667\std{0.0045} \\ 
        \cmidrule(lr){1-8}
        \multirow{2}{*}{GINE + $\mathsf{FN}$}
        &SC& \textbf{0.6815\std{0.0059}} & 0.2534\std{0.0016} & 0.9784\std{0.0010} & 0.6548\std{0.0088} & \textbf{0.7882\std{0.0050}} & \textbf{0.7751\std{0.0029}}\\ 
        &VC& 0.6796\std{0.0024} & \textbf{0.2515\std{0.0020}} & \textbf{0.9790\std{0.0012}} & \textbf{0.6584\std{0.0069}} & 0.7849\std{0.0047} & 0.7672\std{0.0009}\\ 
        \cmidrule(lr){1-8}
        \multirow{2}{*}{GatedGCN + $\mathsf{FN}$}
        &SC& \textbf{0.6778\std{0.0071}} & 0.2546\std{0.0020} & 0.9813\std{0.0018} & 0.7083\std{0.0032} & 0.7910\std{0.0090} & \textbf{0.7759\std{0.0054}}\\ 
        &VC& 0.6647\std{0.0052} & \textbf{0.2536\std{0.0019}} & \textbf{0.9826\std{0.0012}} & \textbf{0.7125\std{0.0035}} & \textbf{0.7967\std{0.0098}} & 0.7662\std{0.0090}\\ 
        \midrule
        \multirow{2}{*}{GCN + $\mathsf{FN}_M$}
        &SC& 0.6773\std{0.0039} & \textbf{0.2464\std{0.0014}} & 0.9444\std{0.0008} & 0.6405\std{0.0065} & 0.7762\std{0.0089} & \textbf{0.7882\std{0.0041}}\\
        &VC& \textbf{0.6787\std{0.0048}} & 0.2485\std{0.0016} &\textbf{ 0.9455\std{0.0004}} & \textbf{0.6413\std{0.0070}} & \textbf{0.7866\std{0.0034}} & 0.7862\std{0.0037}\\
        \cmidrule(lr){1-8}
        \multirow{2}{*}{GINE + $\mathsf{FN}_M$}
        &SC& \textbf{0.7018\std{0.0074}} & 0.2451\std{0.0011} & 0.9735\std{0.0009} & 0.6655\std{0.0066} & 0.8070\std{0.0084} &  0.7924\std{0.0019}\\ 
        &VC& 0.6926\std{0.0105} & \textbf{0.2446\std{0.0018}} & \textbf{0.9786\std{0.0004}} & \textbf{0.6672\std{0.0068}} & \textbf{0.8127\std{0.0076}} &  \textbf{0.7926\std{0.0021}}\\ 
        \cmidrule(lr){1-8}
        \multirow{2}{*}{GatedGCN + $\mathsf{FN}_M$}
        &SC& 0.6932\std{0.0056} & \textbf{0.2453\std{0.0014}} & 0.9836\std{0.0010} & 0.7495\std{0.0051} & \textbf{0.8097\std{0.0047}}  & \textbf{0.7922\std{0.0054}}\\
        &VC& \textbf{0.6950\std{0.0047}} & 0.2461\std{0.0009} & \textbf{0.9836\std{0.0009}} & \textbf{0.7526\std{0.0033}} & 0.8025\std{0.0087} & 0.7885\std{0.0043}\\
    \bottomrule
    \end{tabular}
\end{table*}

\subsection{Sensitivity to $C$}\label{app:sens-c}
\begin{figure}[h!]
    % \vspace{-1em}
    \centering
    \subfigure[\textsc{Peptides-func}]{\includegraphics[width=0.49\columnwidth]{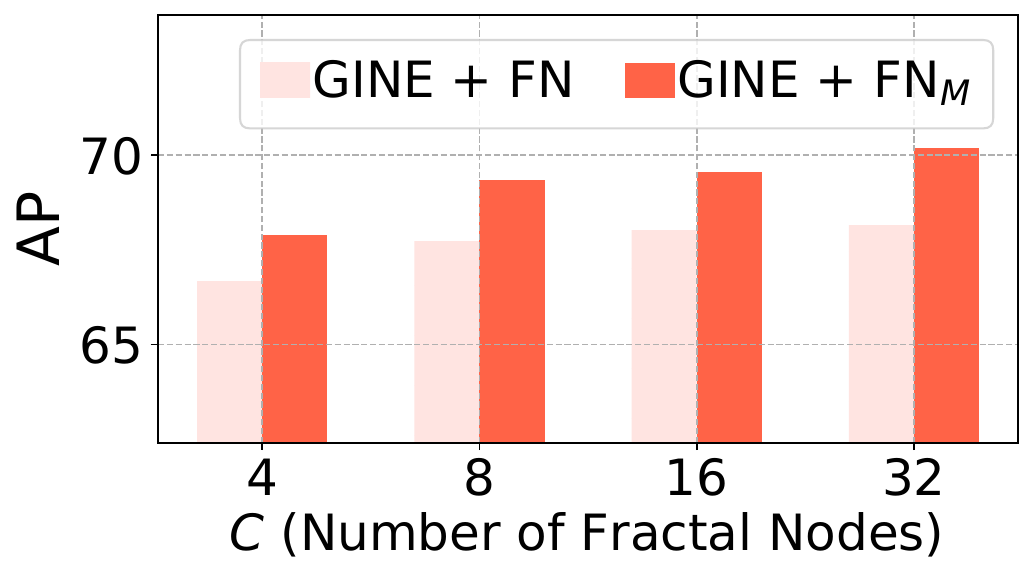}}
    \subfigure[\textsc{Peptides-struct}]{\includegraphics[width=0.49\columnwidth]{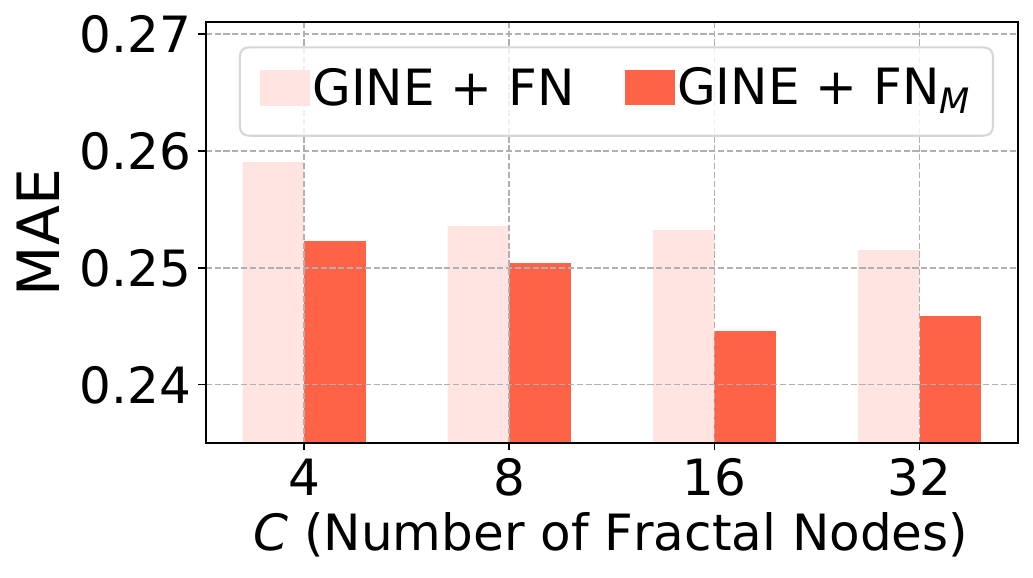}}
    \subfigure[\textsc{MNIST}]{\includegraphics[width=0.49\columnwidth]{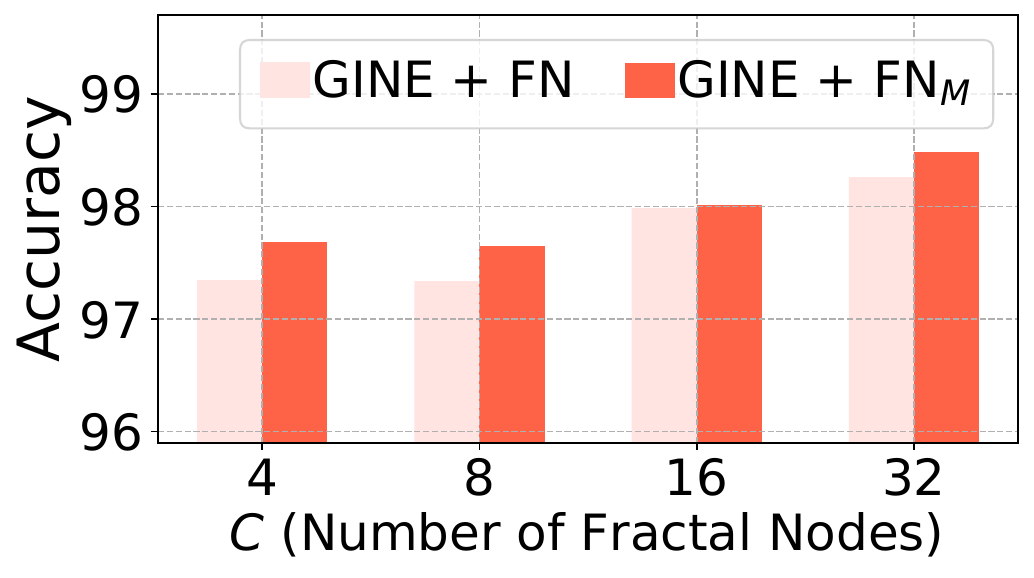}}
    \subfigure[\textsc{CIFAR10}]{\includegraphics[width=0.49\columnwidth]{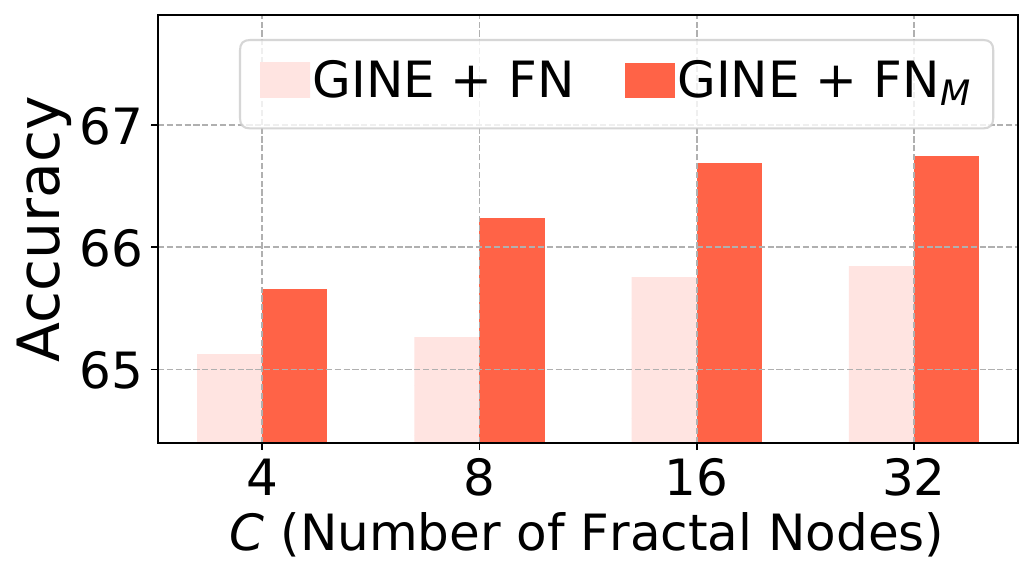}}
    \subfigure[\textsc{MolHiv}]{\includegraphics[width=0.49\columnwidth]{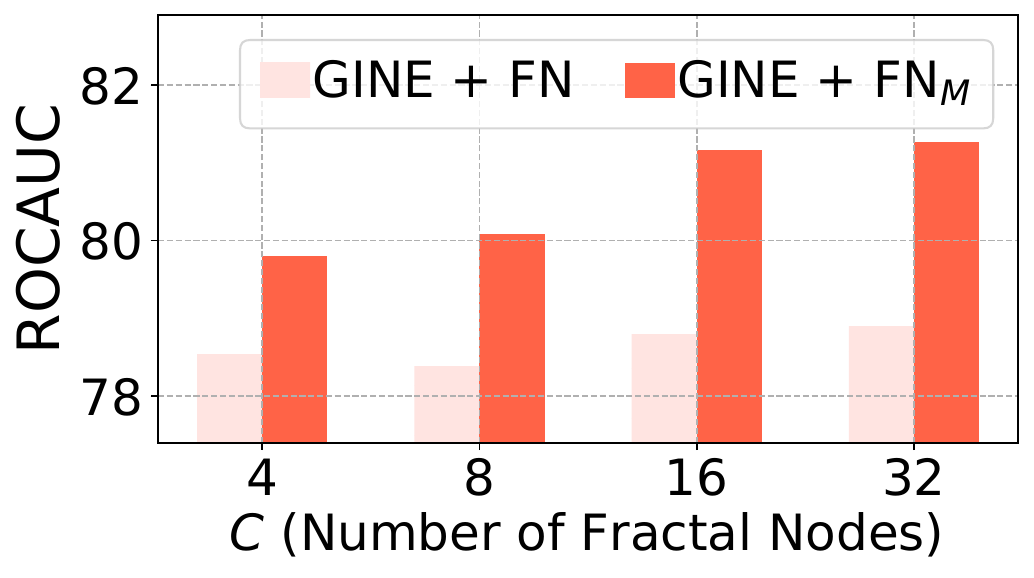}}
    \subfigure[\textsc{MolTox21}]{\includegraphics[width=0.49\columnwidth]{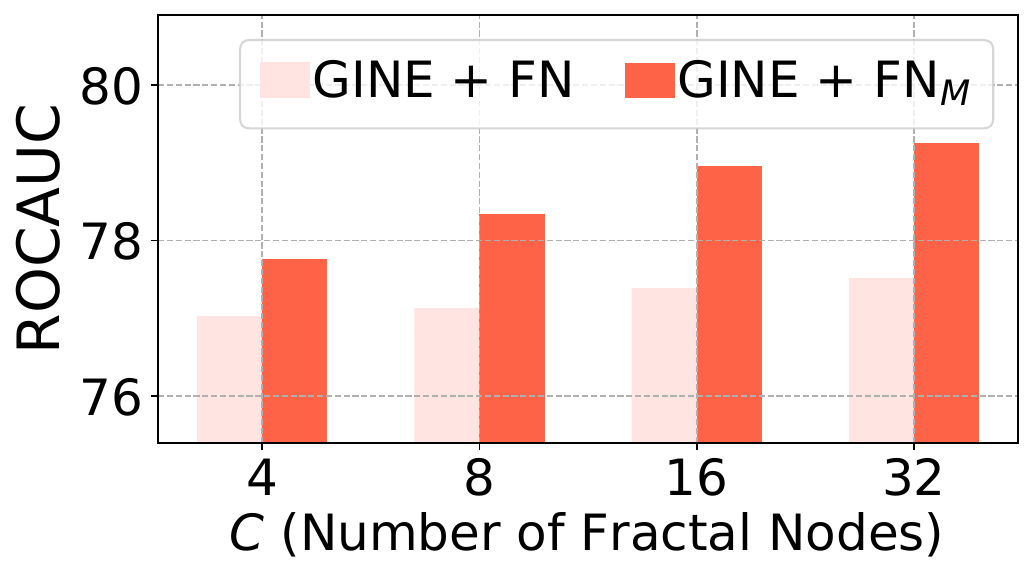}}
    \caption{Sensitivity to $C$ with GINE.}
    \label{fig:sens-c-gine}
\end{figure}

The analysis of sensitivity to the number of fractal nodes ($C$) reveals distinct performance patterns in various datasets. 
As shown in \Cref{fig:sens-c-gine}, for \textsc{Peptides-func} and \textsc{Peptides-struct}, there is relatively stable performance across different $C$ values, with GINE+$\mathsf{FN}_M$ consistently outperforming the baseline GINE+$\mathsf{FN}$. In MNIST, both GINE variants show an upward trend as $C$ increases, with GINE+$\mathsf{FN}_M$ achieving peak accuracy at $C=32$. 

The optimal results are typically achieved at $C=32$, which indicates that graph tasks benefit from finer-grained subgraph partitioning and additional mixing operations in $\mathsf{FN}_M$. Overall, the results indicate that larger $C$ values (16 or 32) generally yield better performance for most datasets.

\subsection{Additional Results on All-layer Fratal Node Message Passing}\label{app:all-fn}

While our main $\mathsf{FN}_M$ design uses an MLP-Mixer in the final layer for fractal node interactions, we also explored an alternative approach with message passing between fractal nodes across all layers (denoted as $\mathsf{FN}_A$). This analysis aims to empirically validate our architectural choice.

\Cref{tab:fn-all} compares 3 variants: i) $\mathsf{FN}$: is a base MPNN with no explicit fractal node interactions; ii) $\mathsf{FN}_A$ is an all-layer message passing between fractal nodes; and iii) $\mathsf{FN}_M$ is an MLP-Mixer in the final layer only (our proposed approach).
The results show that while $\mathsf{FN}_A$  shows some improvements over the base FN model in certain cases (e.g., MOLHIV accuracy improves from 0.7564 to 0.7783 for GCN), it consistently underperforms compared to our proposed $\mathsf{FN}_M$ design. This pattern holds across different base architectures (GCN, GINE, GatedGCN) and datasets.

These empirical results validate our design choice of using MLP-Mixer in the final layer rather than implementing message passing between fractal nodes throughout all layers. This result indicates that the flexible mixing capabilities of the MLP-Mixer provide more effective fractal node interactions than explicit message passing approaches.

\begin{table*}[h]
    % \footnotesize
    % \scriptsize
    % \setlength{\tabcolsep}{2.5pt}
    \small
    \centering
    \caption{Comparison on $\mathsf{FN}$, $\mathsf{FN}_A$ and $\mathsf{FN}_M$}
    \label{tab:fn-all}
    \begin{tabular}{l cccc}\toprule
        \multirow{2}{*}{Method} & \textsc{Peptides-func}   & \textsc{Peptides-struct} & \textsc{MolHIV} & \textsc{MolTox21}\\ \cmidrule(lr){2-2}\cmidrule(lr){3-3}\cmidrule(lr){4-4}\cmidrule(lr){5-5}
        & AP $\uparrow$ & MAE $\downarrow$ & ROCAUC $\uparrow$ &  ROCAUC $\uparrow$\\ \midrule
        GCN + $\mathsf{FN}$ & 0.6802\std{0.0043} & 0.2530\std{0.0004} & 0.7564\std{0.0059} & 0.7670\std{0.0073} \\ 
        GCN + $\mathsf{FN}_A$ & 0.6582\std{0.0032} & 0.2531\std{0.0008} & 0.7783\std{0.0164} &  0.7600\std{0.0037} \\
        GCN + $\mathsf{FN}_M$ & \textbf{0.6787\std{0.0048}} & \textbf{0.2464\std{0.0014}} & \textbf{0.7866\std{0.0034}} & \textbf{0.7882\std{0.0041}}\\
        \cmidrule(lr){1-5}
        GINE + $\mathsf{FN}$ & 0.6815\std{0.0059} & 0.2515\std{0.0020} & 0.7890\std{0.0104} & 0.7751\std{0.0029}\\ 
        GINE + $\mathsf{FN}_A$ & 0.6660\std{0.0067} & 0.2530\std{0.0011} & 0.8025\std{0.0100} & 0.7680\std{0.0056}  \\ 
        GINE + $\mathsf{FN}_M$ & \textbf{0.7018\std{0.0074}} & \textbf{0.2446\std{0.0018}} & \textbf{0.8127\std{0.0076}} &  \textbf{0.7926\std{0.0021}}\\ 
        \cmidrule(lr){1-5}
        GatedGCN + $\mathsf{FN}$ & 0.6778\std{0.0056} & 0.2536\std{0.0019} & 0.7967\std{0.0098} & 0.7759\std{0.0054}\\ 
        GatedGCN + $\mathsf{FN}_A$ & 0.6658\std{0.0048} & 0.2531\std{0.0009} & 0.7898\std{0.0065} & 0.7642\std{0.0050} \\
        GatedGCN + $\mathsf{FN}_M$ & \textbf{0.6950\std{0.0047}} & \textbf{0.2453\std{0.0014}} & \textbf{0.8097\std{0.0047}} & \textbf{0.7922\std{0.0054}}\\
    \bottomrule
    \end{tabular}
\end{table*}

\section{Different Partitioning Algorithms}\label{app:partition}

To verify the effectiveness of partitioning other than METIS partitioning, we conduct experiments applying FN and FN$_M$ to GINE on \textsc{Peptides-func}, \textsc{Peptides-struct}, \textsc{MolHIV}, and \textsc{MolTox21} datasets using random partitioning and Louvain~\citep{blondel2008louvain} and Girvan-Newman~\citep{girvan2002community} partitioning.

In \Cref{tab:partition}, our results provide a comprehensive comparison of different graph partitioning methods for GINE with $\mathsf{FN}$ and $\mathsf{FN}_M$ architectures on multiple molecular and peptide datasets. METIS consistently shows superior or competitive performance on all datasets. It achieves the best results in most cases, such as 0.6815 AP on \textsc{Peptides-func} with GINE+$\mathsf{FN}$ and 0.7018 AP with GINE+$\mathsf{FN}_M$. While random partitioning shows surprisingly competitive performance, particularly on \textsc{MolHiv} where it achieves 0.8039 ROCAUC with GINE+$\mathsf{FN}$, community detection algorithms such as Louvain and Girvan-Newman generally underperform compared to METIS and random partitioning. The performance gap between different partitioning methods becomes more pronounced when using $\mathsf{FN}_M$ compared to $\mathsf{FN}$. 
METIS shows more stable performance with lower standard deviations across all metrics. For molecular property prediction tasks, the choice of partitioning method appears less critical.
However, on \textsc{Peptides-func} and \textsc{Peptides-struct}, METIS shows clear advantages with consistently lower MAE scores. 
These findings validate our choice of METIS as the default partitioning algorithm while suggesting that the optimal partitioning strategy may depend on the specific graph structure and task requirements.

In \Cref{tab:partition-ogbn}, the analysis of results on ogbn-arxiv provides additional insights into partitioning methods on large scale graph datasets. 
The performance differences between partitioning methods are relatively small, with scores ranging between 72.46\% and 73.03\% accuracy. For GCN+$\mathsf{FN}$, METIS achieves the best performance at 73.03\%, while random partitioning performs best for GCN+$\mathsf{FN}_M$ at 73.01\%. GraphSAGE shows slightly lower performance compared to GCN across all partitioning strategies, with Louvain partitioning achieving the best results at 72.76\% for GraphSAGE+$\mathsf{FN}$. Interestingly, the Girvan-Newman algorithm consistently times out on this dataset, indicating scalability issues with larger graphs such as ogbn-arxiv. 
The standard deviations are generally smaller for GraphSAGE compared to GCN, suggesting more stable performance across different random seeds. These results further support that METIS remains competitive.

\begin{table*}[t]
    \small
    % \scriptsize
    \setlength{\tabcolsep}{5pt}
    \centering
    \caption{Comparison of different graph partitioning methods for GINE with $\mathsf{FN}$ and $\mathsf{FN}_M$ architectures on \textsc{Peptides-func/struct} and molecular property prediction tasks. Best results for each metric are shown in \textbf{bold}.}
    \label{tab:partition}
    \begin{tabular}{ll cccc}\toprule
        \multirow{2}{*}{Method} & \multirow{2}{*}{Partitioning} & \textsc{Peptides-func}   & \textsc{Peptides-struct} & \textsc{MolHIV} & \textsc{MolTox21}\\ \cmidrule(lr){3-3}\cmidrule(lr){4-4}\cmidrule(lr){5-5}\cmidrule(lr){6-6}
        & & AP $\uparrow$ & MAE $\downarrow$ & ROCAUC $\uparrow$ &  ROCAUC $\uparrow$\\ \midrule
        \multirow{4}{*}{GINE + $\mathsf{FN}$}
        & METIS & \textbf{0.6815\std{0.0059}} & \textbf{0.2515\std{0.0020}} & 0.7882\std{0.0050} & \textbf{0.7751\std{0.0029}} \\ 
        & Random & 0.6533\std{0.0103} & 0.2688\std{0.0014} & \textbf{0.8039\std{0.0078}} & 0.7653\std{0.0065} \\ 
        & Louvain & 0.6044\std{0.0068} & 0.2799\std{0.0015} & 0.7844\std{0.0050} & 0.7701\std{0.0026} \\
        & Girvan-Newman & 0.6528\std{0.0051} & 0.2628\std{0.0045} & 0.7837\std{0.0078} & 0.7630\std{0.0060}\\
        \cmidrule(lr){1-6}
        \multirow{4}{*}{GINE + $\mathsf{FN}_M$}
        & METIS & \textbf{0.7018\std{0.0074}} & \textbf{0.2446\std{0.0018}} & \textbf{0.8127\std{0.0076}} & \textbf{0.7926\std{0.0021}} \\ 
        & Random & 0.6680\std{0.0066} & 0.2538\std{0.0013} & 0.8090\std{0.0061} & 0.7867\std{0.0045} \\
        & Louvain & 0.6164\std{0.0120} & 0.2789\std{0.0022} & 0.7629\std{0.0164} & 0.7510\std{0.0118} \\
        & Girvan-Newman & 0.6514\std{0.0064} & 0.2655\std{0.0037} & 0.7763\std{0.0174} & 0.7579\std{0.0097}\\
    \bottomrule
    \end{tabular}
\end{table*}

\begin{table*}[t]
    \small
    \centering
    \caption{Comparison of different graph partitioning methods for GCN/GraphSAGE with $\mathsf{FN}$ and $\mathsf{FN}_M$ on ogbn-arxiv dataset. Results show accuracy (\%) and best results for each metric are shown in \textbf{bold}.}
    \label{tab:partition-ogbn}
    \begin{tabular}{l cccc}
        \toprule
        ogbn-arxiv  & GCN + $\mathsf{FN}$  & GCN + $\mathsf{FN}_{M}$ & GraphSAGE + $\mathsf{FN}$ & GraphSAGE + $\mathsf{FN}_M$\\
        \cmidrule(lr){1-5}
        % \midrule
        METIS & \textbf{73.03\std{0.37}} & 72.93\std{0.35} & 72.70\std{0.11} & 72.54\std{0.30} \\
        Random & 72.79\std{0.37} & \textbf{73.01\std{0.41}} & 72.46\std{0.20} & 72.46\std{0.27} \\
        Louvain & 72.73\std{0.57} & 72.95\std{0.26} & \textbf{72.76\std{0.15}} & \textbf{72.56\std{0.58}} \\
        Girvan-Newman & Time-out & Time-out & Time-out & Time-out \\
        \bottomrule
    \end{tabular}
\end{table*}

\Cref{tab:runtime-partition} demonstrates the empirical runtime performance of different graph partitioning algorithms across various graph-level tasks, providing evidence for the practicality of our approach. While all algorithms show comparable performance on smaller datasets like Peptides (with runtimes in microseconds), noticeable differences emerge starting with medium-sized datasets like MNIST.

The distinction becomes particularly pronounced on large-scale datasets like ogbn-arxiv. We opt for METIS as our default partitioning algorithm due to its theoretical time complexity of $O(|E|)$ and superior empirical performance. METIS efficiently partitions large graphs such as ogbn-arxiv in under 9 seconds, and even handles massive graphs like ogbn-products around 15 minutes.

In contrast, the Louvain algorithm requires over 50 seconds for ogbn-arxiv, while the Girvan-Newman algorithm encounters runtime limitations, making it impractical for large-scale graphs like ogbn-arxiv and ogbn-products. These results validate our choice of METIS as the primary partitioning algorithm, as it provides an effective balance between computational efficiency and partition quality across different graph scales.

\begin{table*}[t]
    \small
    \centering
    \setlength{\tabcolsep}{1.2pt}
    \caption{Empirical runtime of partitioning algorithms.}
    \label{tab:runtime-partition}
    \begin{tabular}{l ccccc}
        \toprule
        Algorithm  & \textsc{Peptides-func/struct}   & \textsc{MNIST} & \textsc{MolHIV} & ogbn-arxiv & ogbn-product\\
        \cmidrule(lr){1-6}
        % \midrule
        METIS & 0.71 $\mu$s & 0.36 s & 0.71$\mu$s & 8.57 s & 923.27 s\\
        Louvain & 1.19 $\mu$s & 0.36 s & 1.19$\mu$s & 52.12s & 119 m\\
        Girvan-Newman & 1.19 $\mu$s & 0.36 s & 0.72$\mu$s & Time-out & Time-out\\
        \bottomrule
    \end{tabular}
\end{table*}

\section{Distribution Analysis of Subgraph Size Ratio}\label{app:ratio}
We analyze the distribution of subgraph size ratios produced by METIS partitioning across different numbers of partitions ($C$) and datasets.

In general, as $C$ increases from 2 to 32, the average subgraph size ratio naturally decreases since each partition contains a smaller portion of the original graph. The width of the distributions generally increases with $C$, indicating more variance in partition sizes with finer granularity. Most datasets show roughly normal or slightly skewed distributions around the expected mean ratio of 1/$C$.

\begin{figure*}[h]
    \centering
    \subfigure[$C=2$]{\includegraphics[width=0.25\linewidth]{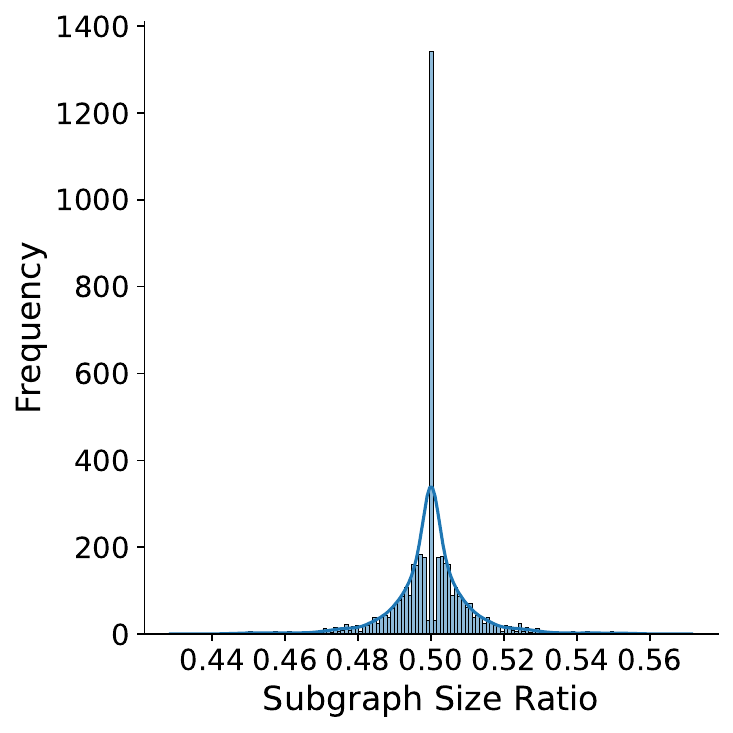}}
    \subfigure[$C=4$]{\includegraphics[width=0.25\linewidth]{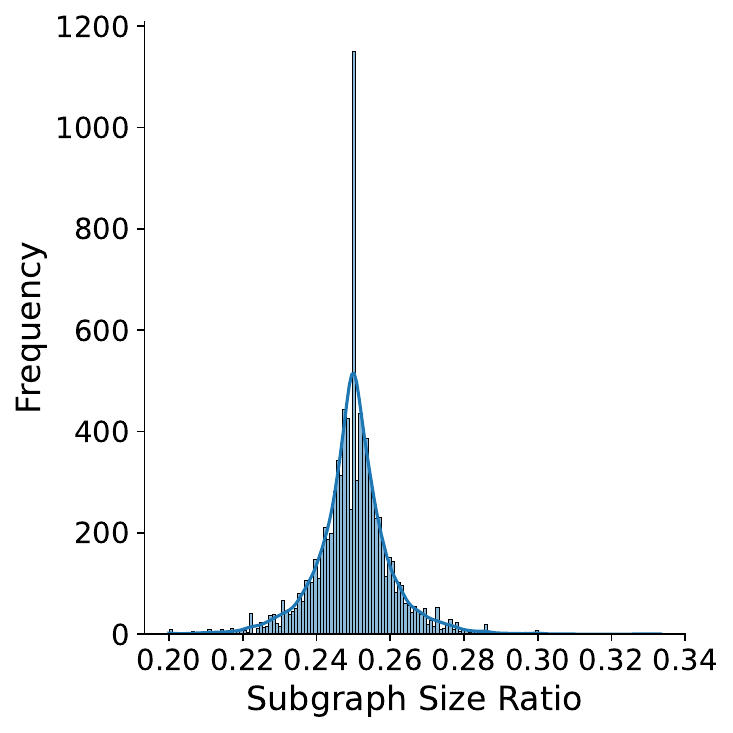}}
    \subfigure[$C=8$]{\includegraphics[width=0.25\linewidth]{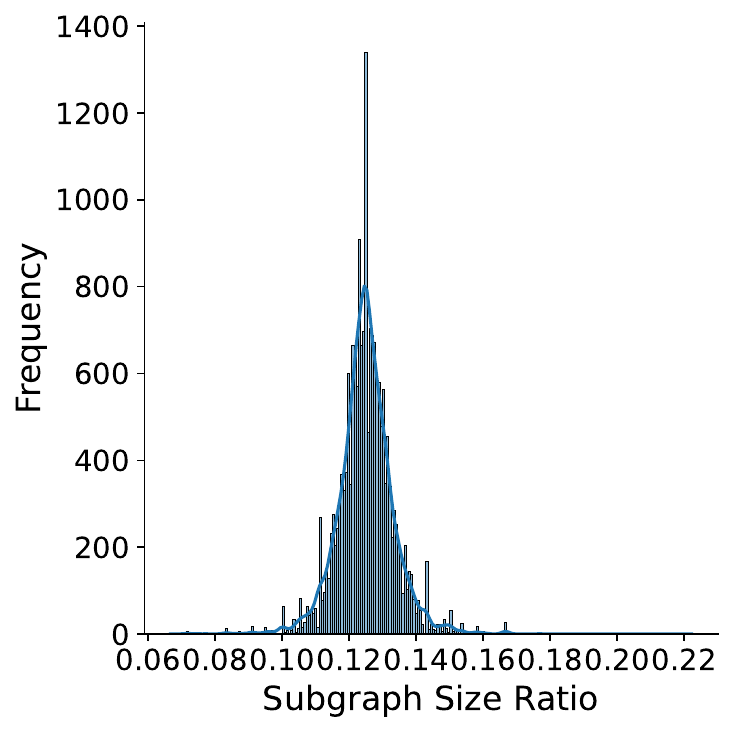}}
    \subfigure[$C=16$]{\includegraphics[width=0.25\linewidth]{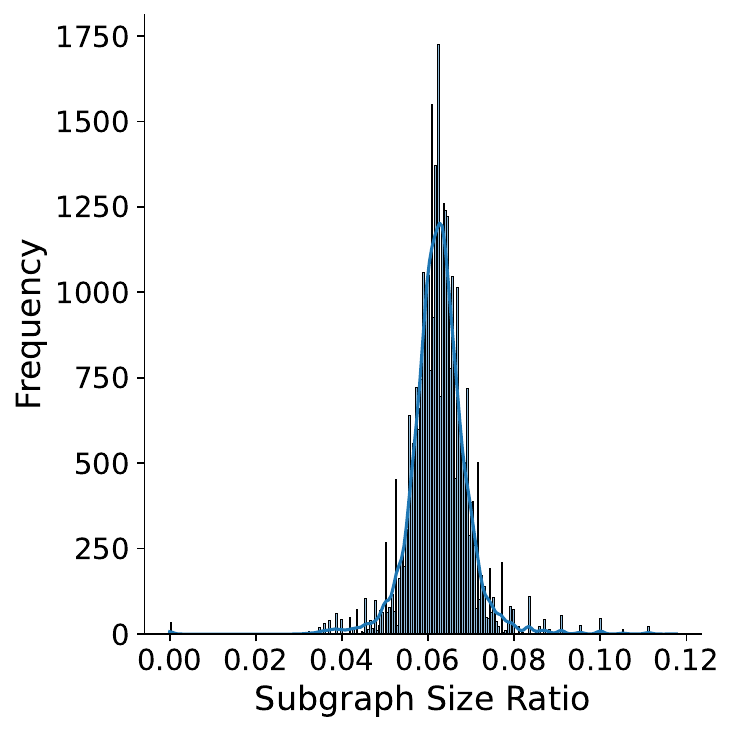}}
    \subfigure[$C=32$]{\includegraphics[width=0.25\linewidth]{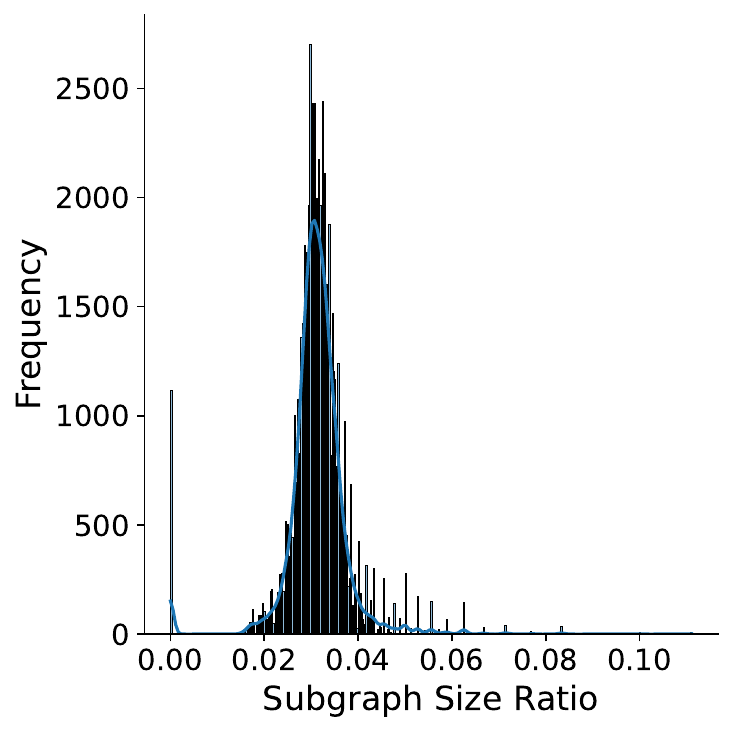}}
    \caption{Similarity of node centrality distribution in \textsc{Peptides-func/struct}.}
    \label{fig:ratio-peptide}
    % \vspace{-0.5cm}
\end{figure*}

\begin{figure*}[h]
    \centering
    \subfigure[$C=2$]{\includegraphics[width=0.25\linewidth]{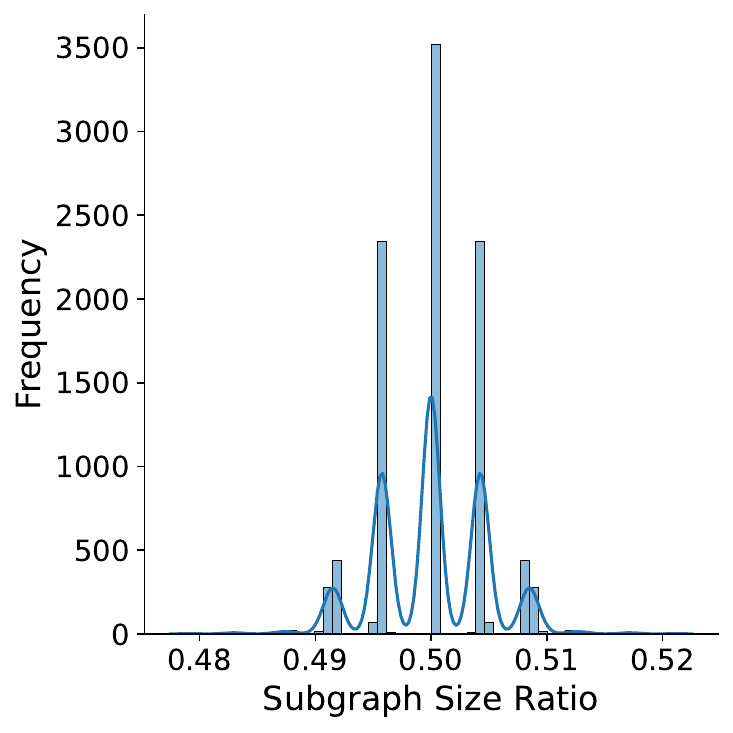}}
    \subfigure[$C=4$]{\includegraphics[width=0.25\linewidth]{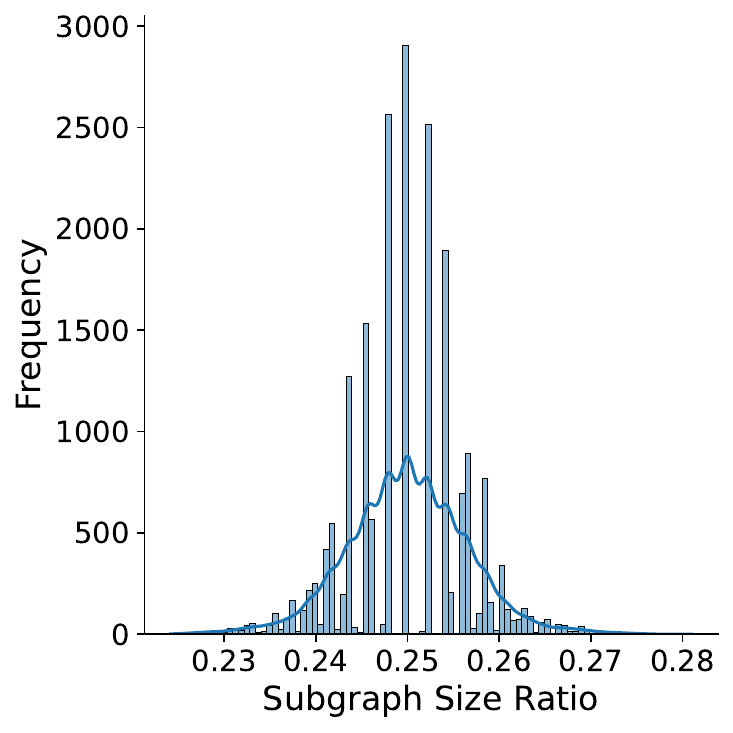}}
    \subfigure[$C=8$]{\includegraphics[width=0.25\linewidth]{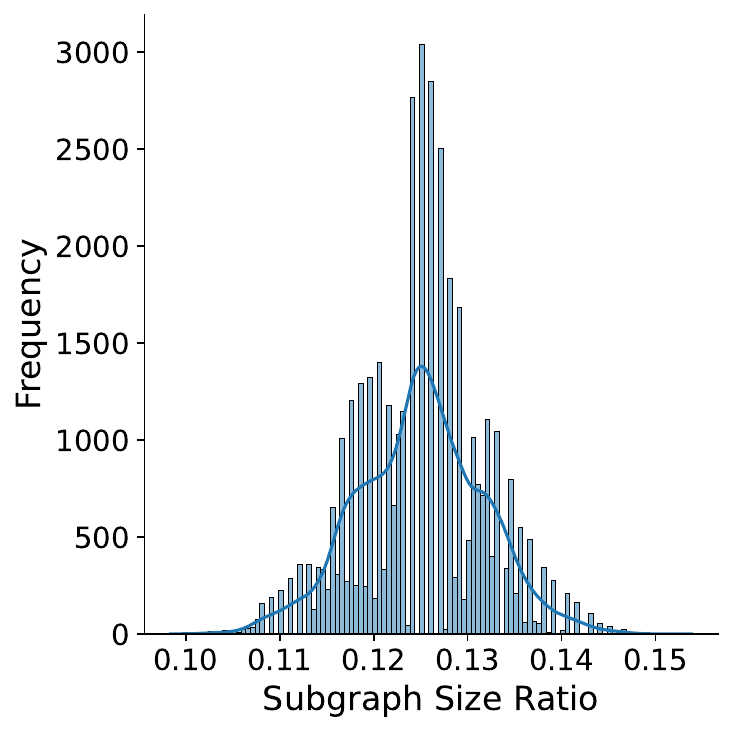}}
    \subfigure[$C=16$]{\includegraphics[width=0.25\linewidth]{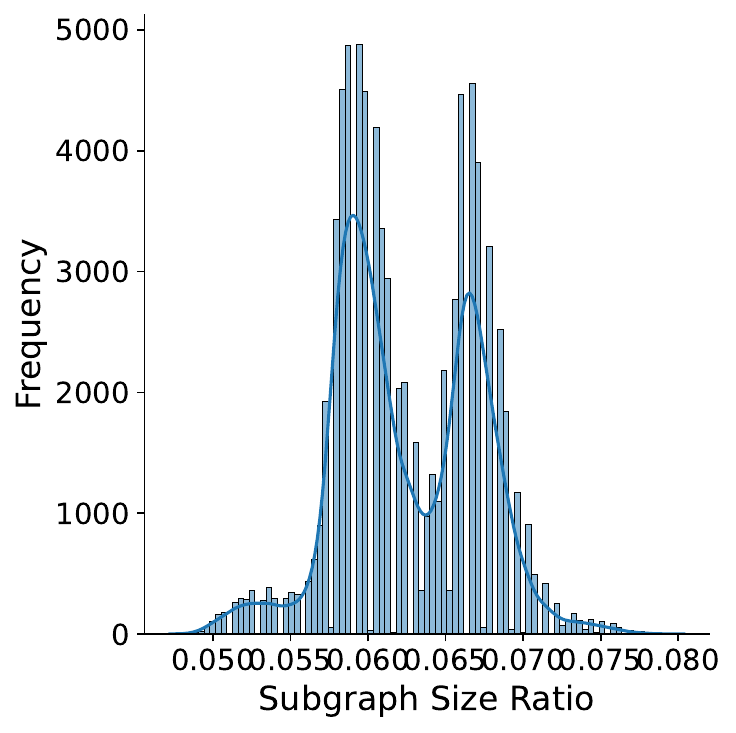}}
    \subfigure[$C=32$]{\includegraphics[width=0.25\linewidth]{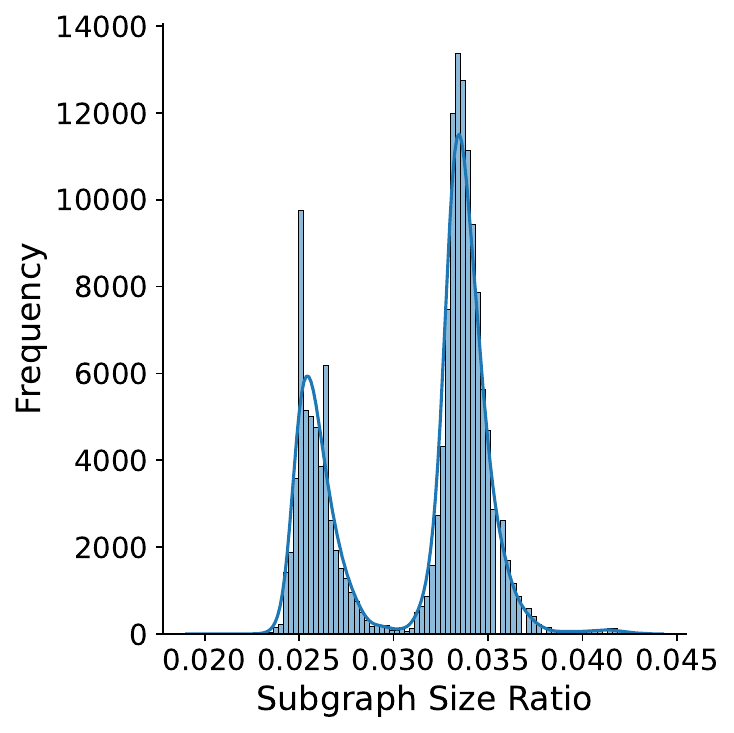}}
    \caption{Similarity of node centrality distribution in \textsc{CIFAR10}.}
    \label{fig:ratio-cifar10}
\end{figure*}

\begin{figure*}[h]
    \centering
    \subfigure[$C=2$]{\includegraphics[width=0.25\linewidth]{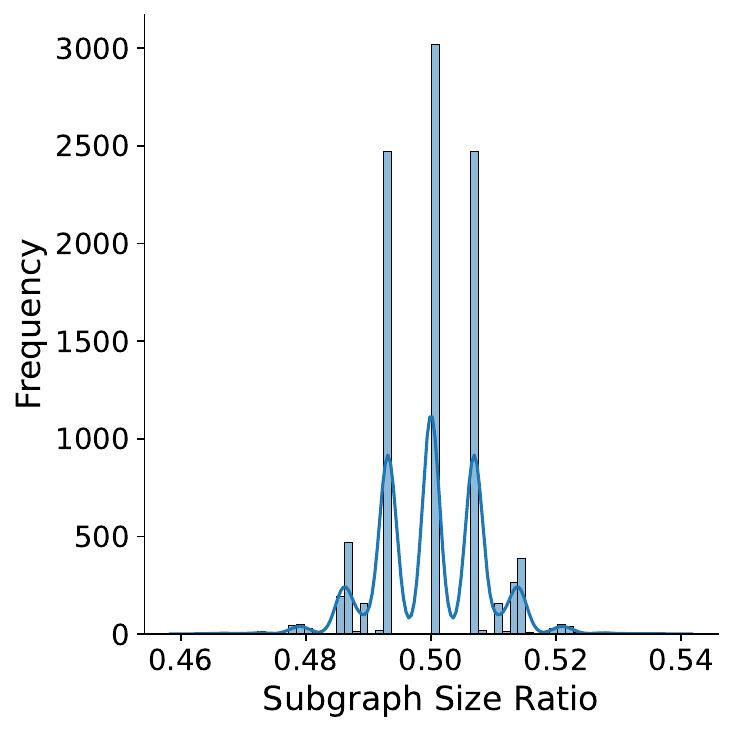}}
    \subfigure[$C=4$]{\includegraphics[width=0.25\linewidth]{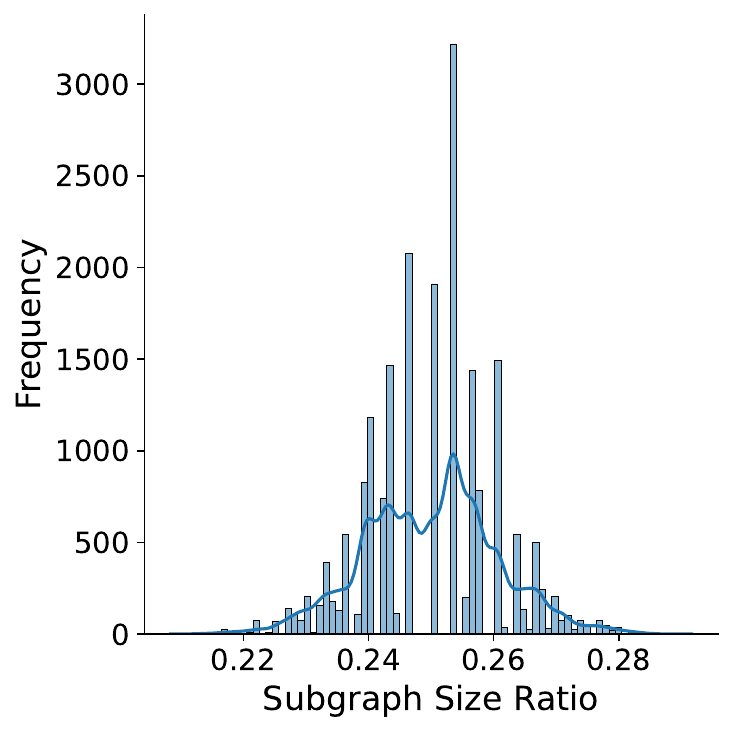}}
    \subfigure[$C=8$]{\includegraphics[width=0.25\linewidth]{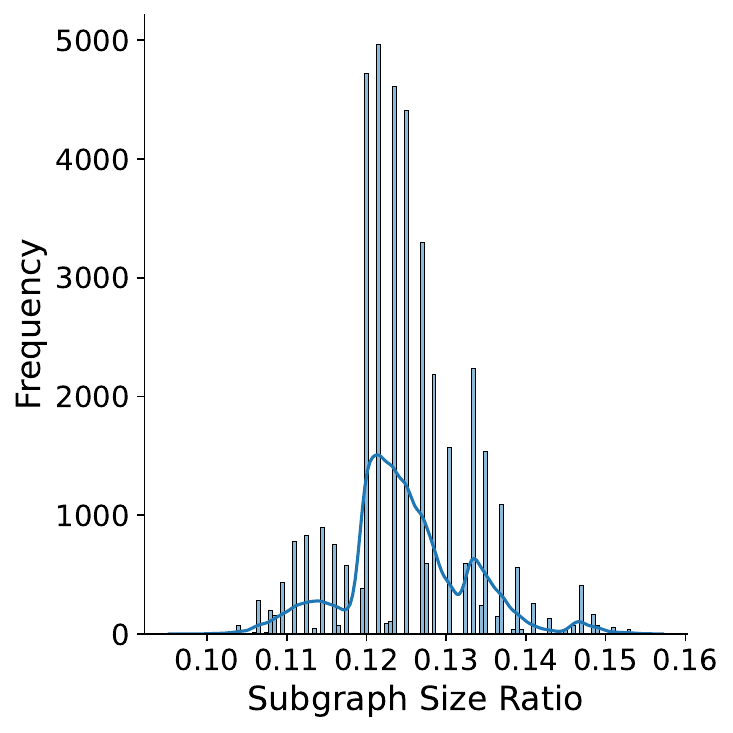}}
    \subfigure[$C=16$]{\includegraphics[width=0.25\linewidth]{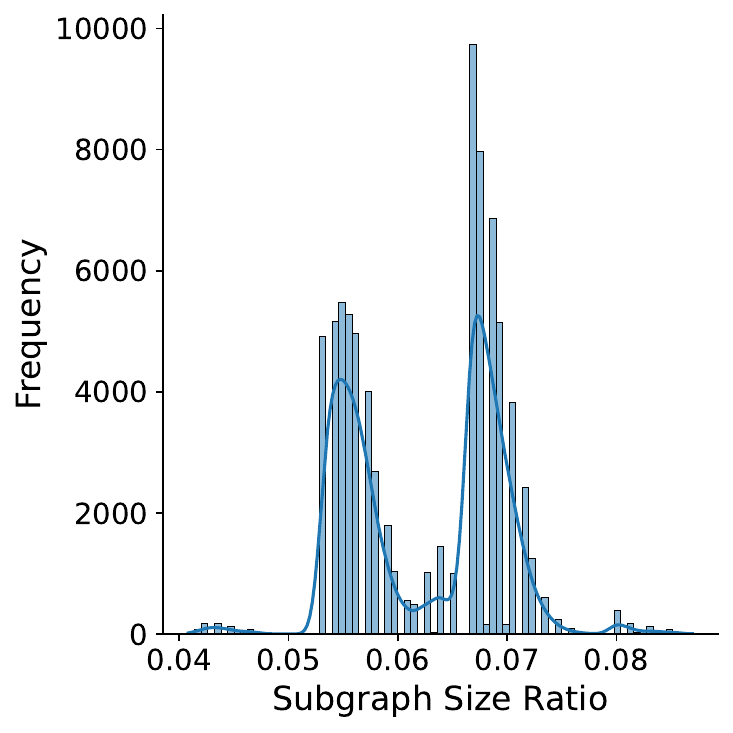}}
    \subfigure[$C=32$]{\includegraphics[width=0.25\linewidth]{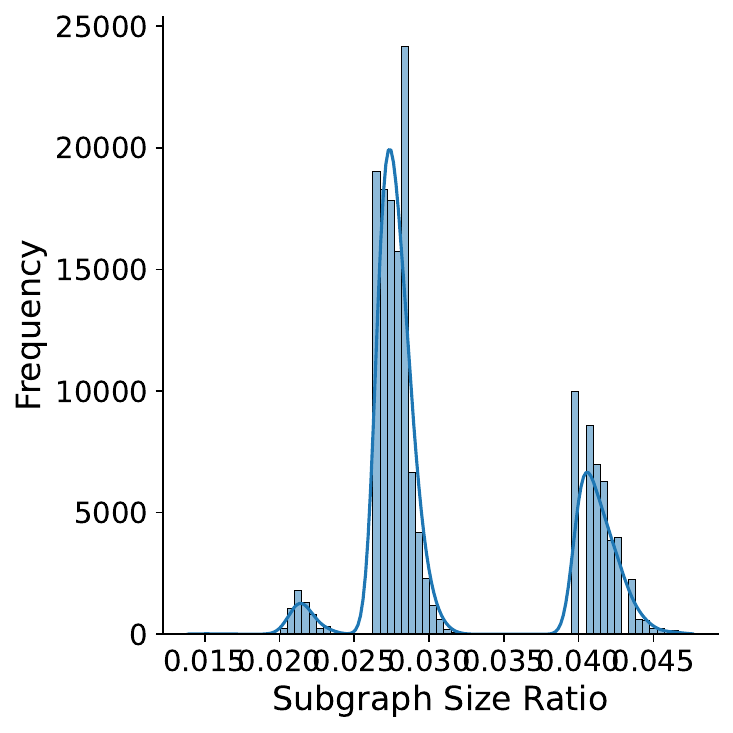}}
    \caption{Similarity of node centrality distribution in \textsc{MNIST}.}
    \label{fig:ratio-mnist}
\end{figure*}

\begin{figure*}[h]
    \centering
    \subfigure[$C=2$]{\includegraphics[width=0.25\linewidth]{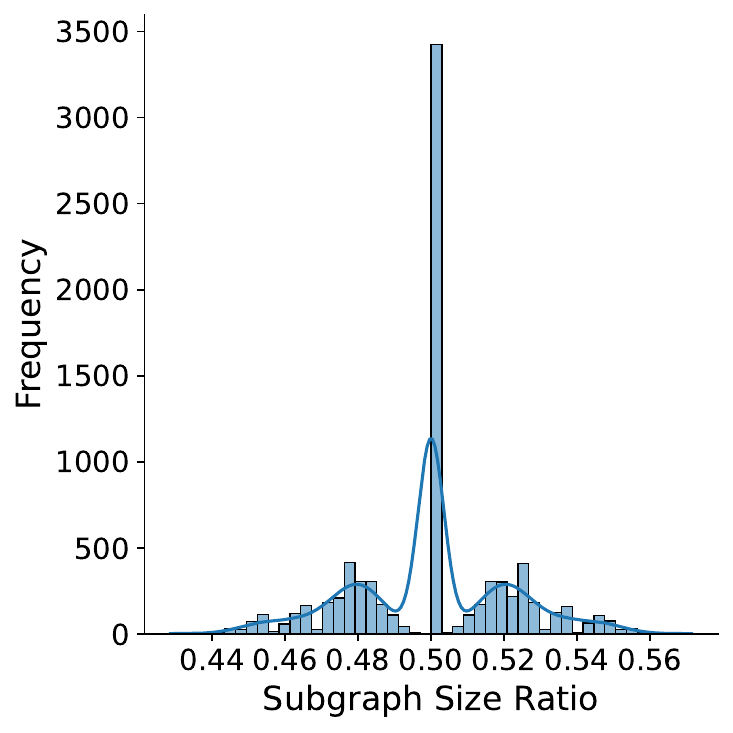}}
    \subfigure[$C=4$]{\includegraphics[width=0.25\linewidth]{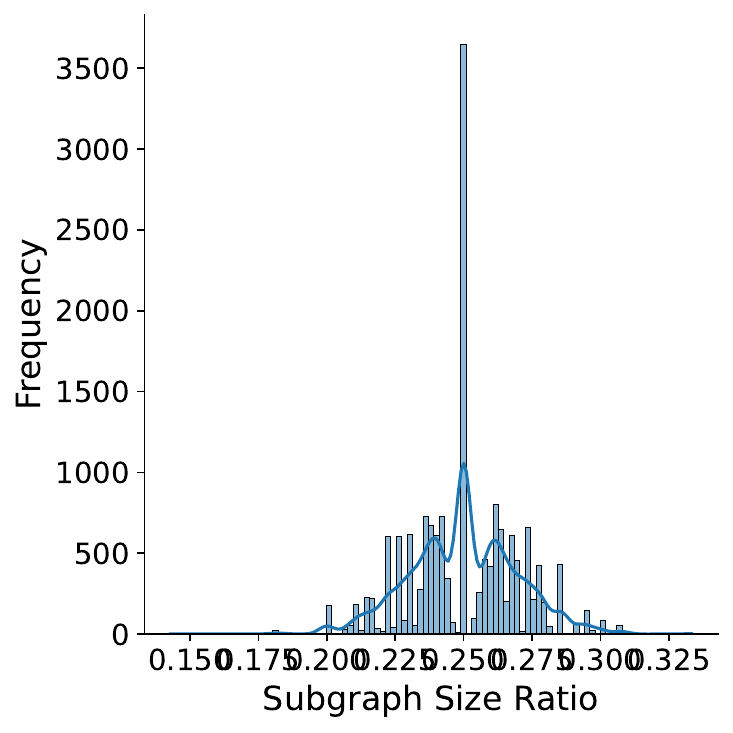}}
    \subfigure[$C=8$]{\includegraphics[width=0.25\linewidth]{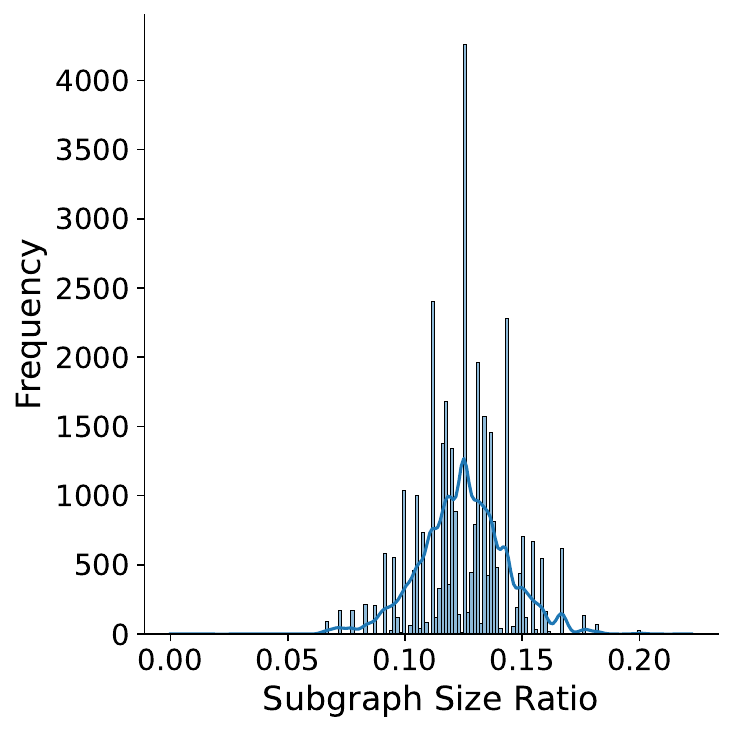}}
    \subfigure[$C=16$]{\includegraphics[width=0.25\linewidth]{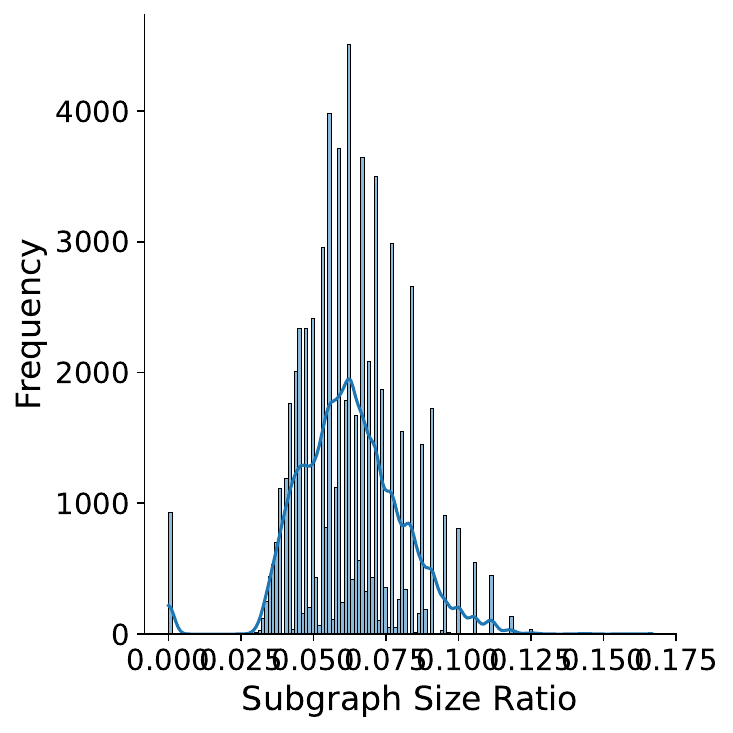}}
    \subfigure[$C=32$]{\includegraphics[width=0.25\linewidth]{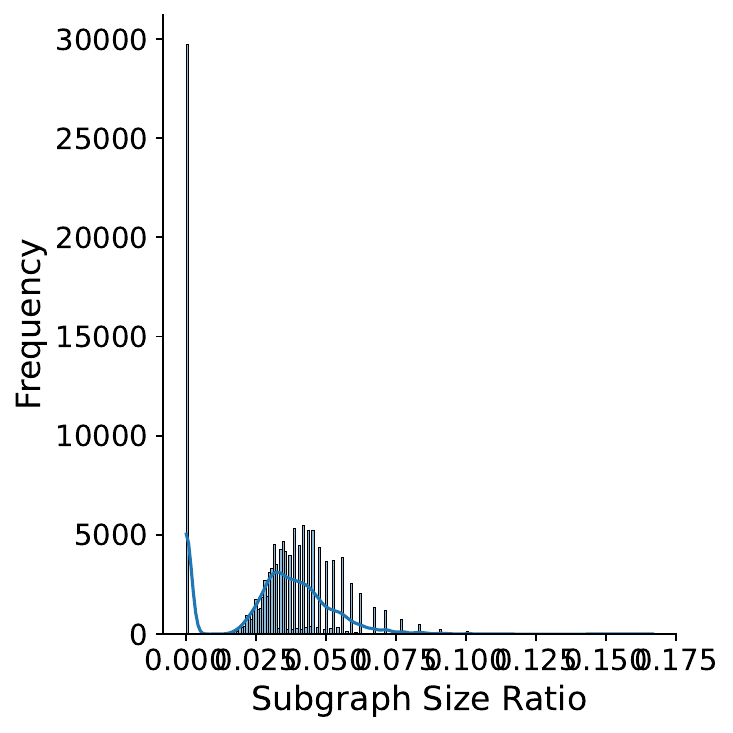}}
    \caption{Similarity of node centrality distribution in \textsc{MolHIV}.}
    \label{fig:ratio-molhiv}
\end{figure*}

\begin{figure*}[h]
    \centering
    \subfigure[$C=2$]{\includegraphics[width=0.25\linewidth]{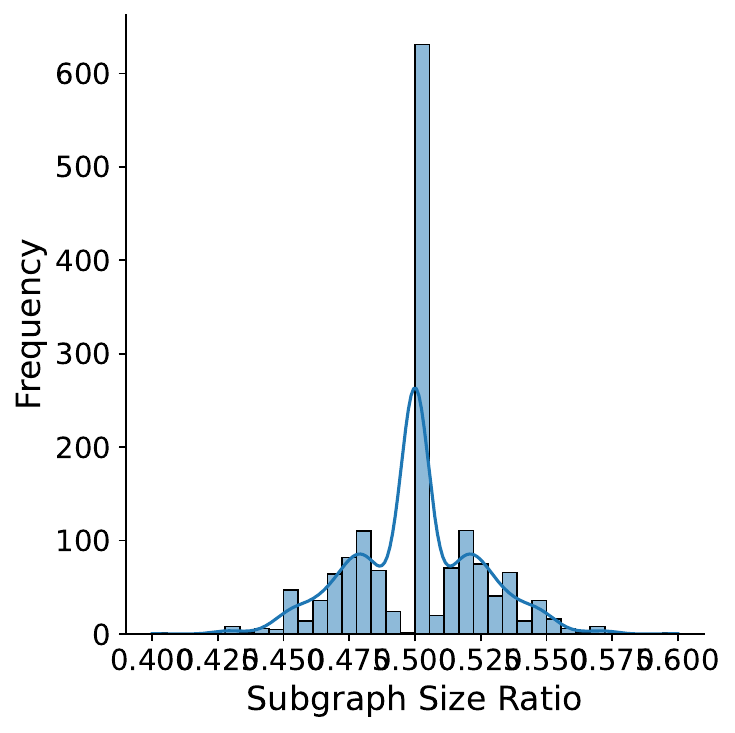}}
    \subfigure[$C=4$]{\includegraphics[width=0.25\linewidth]{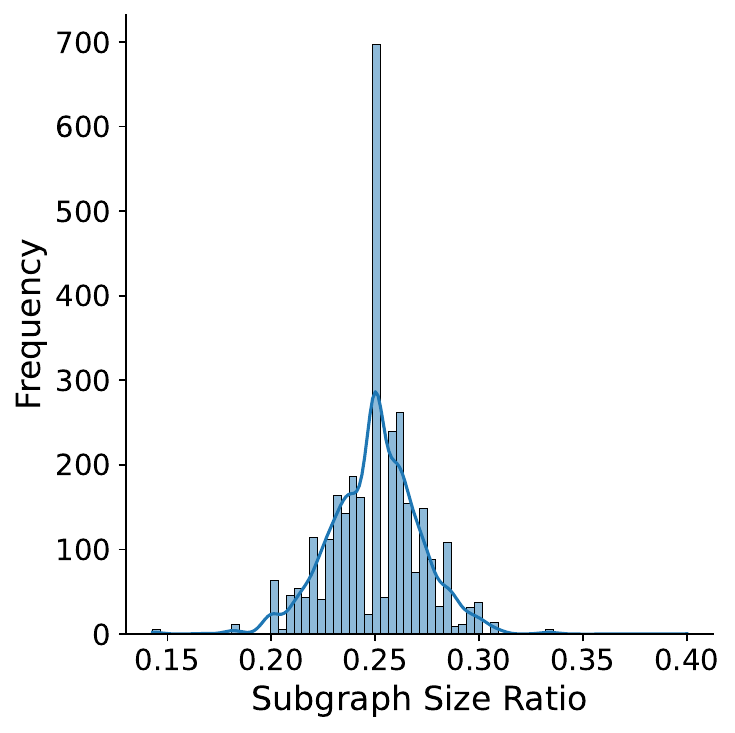}}
    \subfigure[$C=8$]{\includegraphics[width=0.25\linewidth]{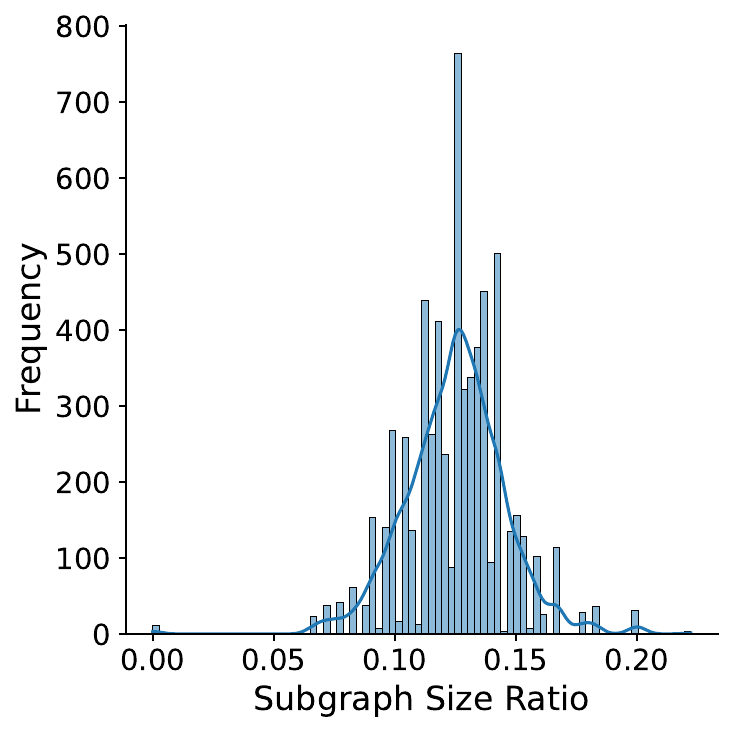}}
    \subfigure[$C=16$]{\includegraphics[width=0.25\linewidth]{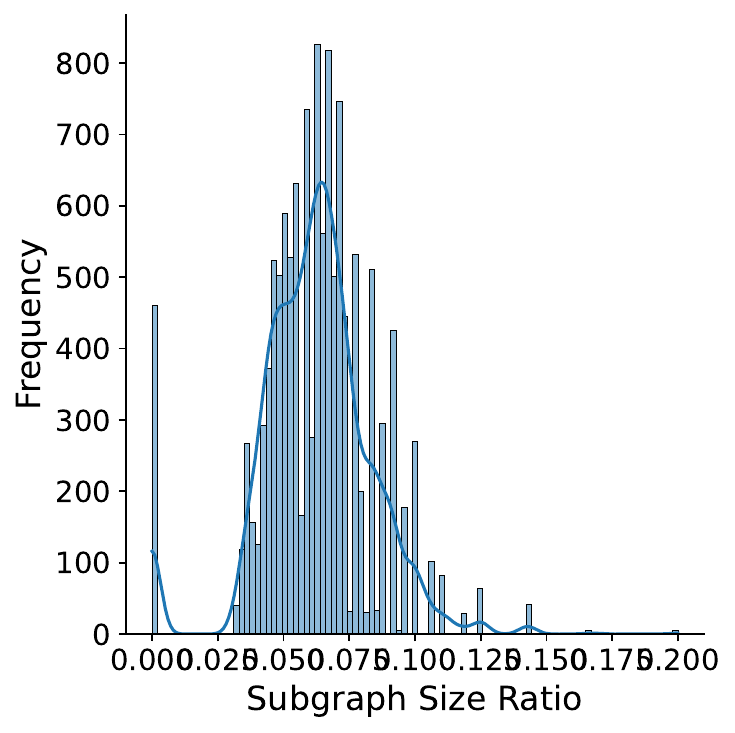}}
    \subfigure[$C=32$]{\includegraphics[width=0.25\linewidth]{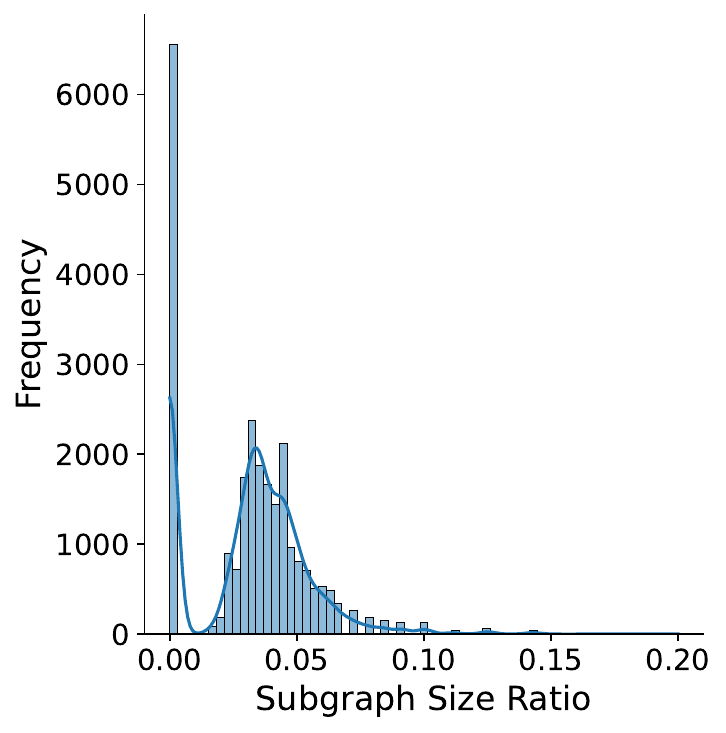}}
    \caption{Similarity of node centrality distribution in \textsc{MolTox21}.}
    \label{fig:ratio-moltox}
\end{figure*}

As shown in \Cref{fig:ratio-peptide}, \textsc{Peptides-func/struct} show relatively tight, symmetric distributions. In indicates that METIS creates balanced partitions for molecular graphs. \textsc{CIFAR10} and \textsc{MNIST} show distinct bimodal patterns, especially at $C=16$ and $C=32$, likely due to the regular grid-like structure of superpixel graphs (See \Cref{fig:ratio-cifar10} and \Cref{fig:ratio-mnist}).
As shown in \Cref{fig:ratio-molhiv} and \Cref{fig:ratio-moltox}, \textsc{MolHiv} and \textsc{MolTox21} show broader distributions, particularly at higher $C$ values, reflecting the more heterogeneous nature of these molecular graphs.

The consistent distributions for molecular datasets indicate METIS partitioning is well-suited for these graph types. The bimodal distributions in image-based graphs indicate the natural clustering of superpixels into regions of different sizes. Higher $C$ values (i.e., 16, 32) generally maintain reasonable balance while allowing for more fine-grained capture of graph structure.

\end{document}